\documentclass[journal]{IEEEtran}
\PassOptionsToPackage{dvipsnames}{xcolor}

\usepackage{booktabs}

\usepackage{graphicx} %
\usepackage{amsmath}
\usepackage{amssymb}
\usepackage{amsfonts}
\usepackage{bbm}
\usepackage{commath}  %
\usepackage{cuted}  %
\usepackage[hidelinks]{hyperref}
\usepackage[inline]{enumitem}
\usepackage{todonotes}
\usepackage{xcolor}
\usepackage[hang,flushmargin]{footmisc}
\usepackage[export]{adjustbox}  %
\usepackage{tcolorbox} %
\usepackage{pmboxdraw} %
\usepackage{xspace}
\usepackage{flushend}

\usepackage{subcaption} %
\usepackage[absolute,overlay]{textpos} %

\definecolor{CrimsonRed}{HTML}{DC143C}

\newcommand{\para}[1]{\parskip=5pt\noindent\textit{#1}}

\renewcommand{\secref}[1]{Sec.~\ref{#1}}
\renewcommand{\eqref}[1]{Eq.~(\ref{#1})}
\renewcommand{\figref}[1]{Fig.~\ref{#1}}
\newcommand{\tabref}[1]{Tab.~\ref{#1}}

\newcommand\subfiguresubref[1]{Subfig.~(\subref{#1})}

\newcommand{\etal}{\emph{et al.}}

\newcommand{\mani}{\mathcal{M}}

\newcommand{\GP}{\mathcal{G\mkern-2muP}}
\newcommand{\GPM}{\mathcal{G\mkern-2muP\mkern-2muM}}
\newcommand{\DM}{\mathcal{D\mkern-2muM}}

\newcommand{\logmap}{\text{Log}}

\newcommand{\ourmethod}{MiDiGaP\xspace}
\newcommand{\dtgp}{DiGaP}
\newcommand{\cgp}{CoGap}
\newcommand{\vapor}{VAPOR}
\newcommand{\vaporfull}{Variance-Aware Path Optimization}

\usepackage{siunitx}

\usepackage{threeparttable}
\usepackage{tabularx}
\newcolumntype{Y}{>{\centering\arraybackslash}X}
\newcolumntype{Z}{>{\raggedleft\arraybackslash}X}
\newcolumntype{R}{>{\raggedright\arraybackslash}X}

\usepackage{multirow}
\usepackage{makecell}
\usepackage{microtype}

\usepackage{utfsym}
\usepackage{cite}

\usepackage{pifont}%
\newcommand{\cmark}{\ding{51}}%
\newcommand{\xmark}{}

\usepackage{layouts}  %

\newcommand{\rebuttal}[1]{#1}
\newcommand{\finalchanges}[1]{#1}

\author{Jan Ole von Hartz, Joschka Boedecker, and Abhinav Valada}
\date{May 2025}

\begin{document}

\title{\LARGE \bf
The Unreasonable Effectiveness of Discrete-Time Gaussian Process Mixtures for Robot Policy Learning}
\author{
Jan Ole von Hartz, Adrian Röfer, Joschka Boedecker, and Abhinav Valada%
\thanks{Department of Computer Science, University of Freiburg, Germany.}%
\thanks{This work was supported by Carl Zeiss Foundation with the ReScaLe project, the German Research Foundation (DFG): 417962828, and by the BrainLinks-BrainTools center of the University of Freiburg. {Contact: \tt\footnotesize hartzj@cs.uni-freiburg.de}}%
}

\maketitle

\begin{abstract}
    We present Mixture of Discrete-time Gaussian Processes (\ourmethod), a novel approach for flexible policy representation and imitation learning in robot manipulation.
    \ourmethod{} enables learning from as few as five demonstrations using only camera observations and generalizes across a wide range of challenging tasks. It excels at long-horizon behaviors such as making coffee, highly constrained motions such as opening doors, dynamic actions such as scooping with a spatula, and multimodal tasks such as hanging a mug.
    \ourmethod{} learns these tasks on a CPU in less than a minute and scales linearly to large datasets. We also develop a rich suite of tools for inference-time steering using evidence such as collision signals and robot kinematic constraints. This steering enables novel generalization capabilities, including obstacle avoidance and cross-embodiment policy transfer. \ourmethod{} achieves state-of-the-art performance on diverse few-shot manipulation benchmarks. On constrained RLBench tasks, it improves policy success by 76 percentage points and reduces trajectory cost by 67\%.
    On multimodal tasks, it improves policy success by 48 percentage points \rebuttal{while improving sample efficiency 7-fold}.
    In cross-embodiment transfer, it more than doubles policy success.
    We make the code publicly available.
\end{abstract}

\section{Introduction}
Learning to imitate a motion is easy, given a suitable representation.
For example, if we know the motion to follow a sine function, estimating its parameters from a demonstration is straightforward.
However, robot manipulation policies are more intricate in practice than simple sine functions. Therefore, policy representations must trade off challenging requirements, such as expressivity, multimodality, sample efficiency, and computational cost.
Additionally, to move beyond the confines of controlled lab experiments and into real-world environments like homes, a policy should react to new information, \rebuttal{such as clutter}, without necessitating retraining.
\rebuttal{Prior approaches often satisfy some of these  competing requirements only by compromising others, whereas we propose a representation that seeks to reconcile them within a single framework.}

\begin{figure}
    \centering
    \includegraphics[width=\columnwidth, trim={0 1.5cm 0 1.8cm},clip]{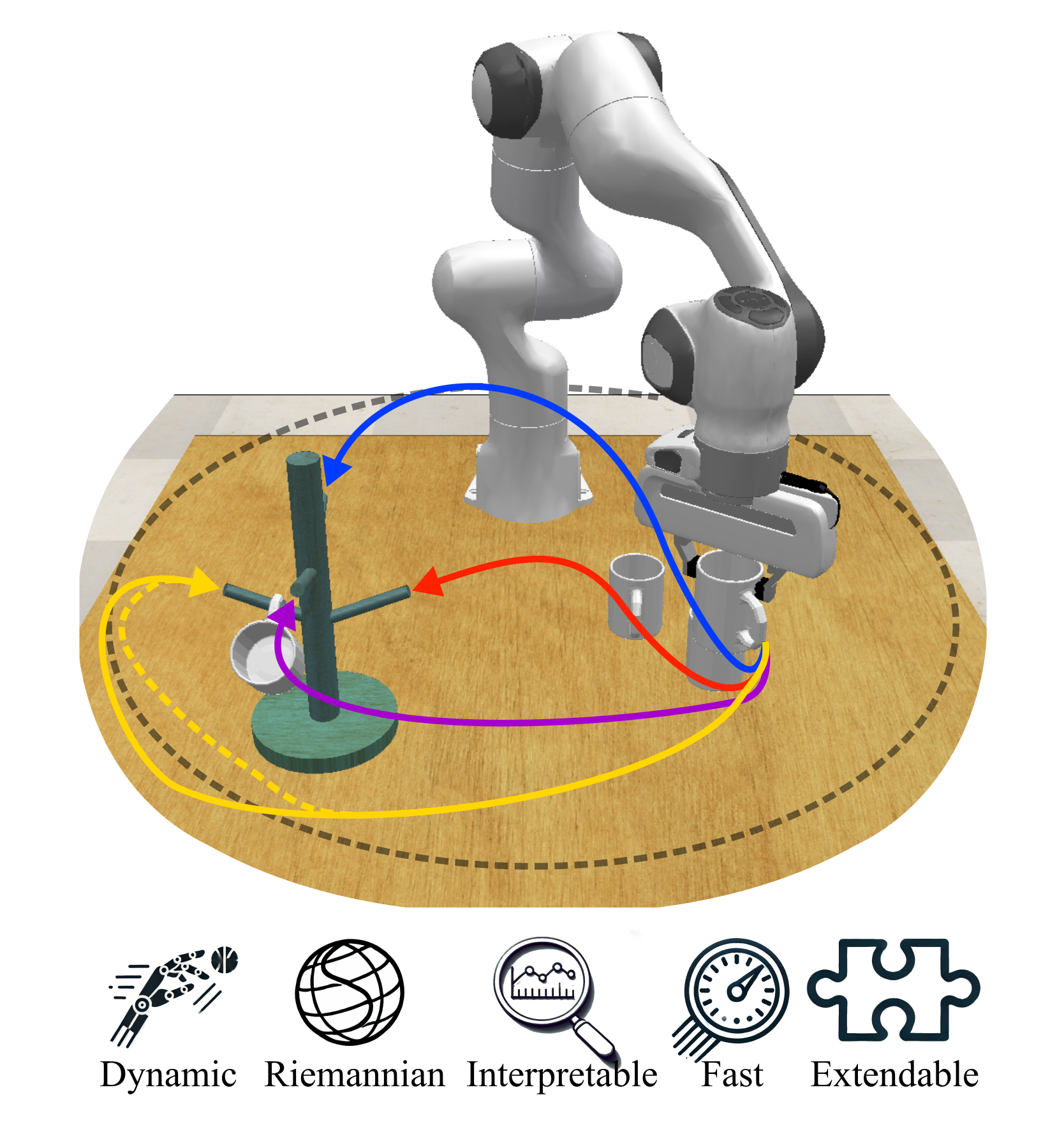}
    \caption{\textit{Mixtures of Discrete-Time Gaussian Processes} (\ourmethod{}) effectively model multimodal trajectory distributions.
    It makes minimal assumptions, allowing it to model even highly-constrained movements. \ourmethod{} is computationally inexpensive and scales linearly with large datasets and long horizons.
    Its interpretability makes it safe to execute dynamic tasks such as scooping.
    We develop a rich tool set to update \ourmethod{} using additional evidence such as reachability and collision information.
    \rebuttal{In this example, four modes of solving the task are available initially (blue, red, purple, and yellow).
    The yellow mode violates the robots workspace limits (gray dotted line) in this instance.
    Updating it (yellow dotted line) enables execution of the mode.}
    }
    \label{fig:overview}
  \vspace{-0.6cm}
\end{figure}

Among deep learning approaches, Diffusion Policy~\cite{janner2022planning, chi2023diffusionpolicy} and Conditional Flow Matching~\cite{chisari2024learningroboticmanipulationpolicies, lipman2022flow, braun2024riemannian} stand out as they can model vast families of motions. Notably, they can model \emph{multimodal} trajectory distributions~\cite{chi2023diffusionpolicy}.
However, they are computationally expensive and usually require in the order of one hundred demonstrations~\cite{vonhartz2023treachery, chi2023diffusionpolicy, zhang2024arp}.
Moreover, deep learning policies typically predict a point estimate, instead of a full action distribution.
As we will demonstrate, this makes it challenging to steer the policies at inference time.
In contrast, Gaussian Mixture Models (GMM)~\cite{calinon2006learning, zeestraten2018programming, vonhartz2024art} and Continuous Gaussian Processes (\cgp)~\cite{deisenroth2013gaussian, titsias2009variational, franzese2024generalization}  predict a Gaussian action distribution, hence allowing for Bayesian updating.
Moreover, by leveraging multi-stream learning, GMMs only require five demonstrations to generalize well~\cite{calinon2006learning}.
This holds even when learning complex long-horizon tasks from camera observations~\cite{vonhartz2024art}.
However, they exhibit critical limitations of their own.
First, \finalchanges{GMMs typically use Gaussian Mixture Regression for inference, which assumes locally linear dynamics}~\cite{sun2023damm}.
This limits the kind of trajectories it can effectively model.
As shown in \figref{fig:tp_models}, GMMs fail at tasks such as door opening.
Furthermore, they are sensitive to their initialization~\cite{zeestraten2018programming} and suffer from the curse of dimensionality~\cite{bouveyron2007high}.
Consequently, they are not well suited to model multimodal trajectory distributions.
Additionally, they are prone to learning spurious correlations, leading to faulty predictions~\cite{vonhartz2024art}.
Continuous Gaussian Processes~\cite{deisenroth2013gaussian, titsias2009variational, franzese2024generalization} offer greater expressivity but are constrained by the choice of kernel function~\cite{deisenroth2013gaussian}.
They fail to model piecewise linear, oscillatory, and non-stationary trajectory distributions.
Illustrative examples and further discussion can be found in \secref{sec:background}.
Although probabilistic, \cgp{}s provide only certainty estimates, not variability, which limits their use in multi-stream learning~\cite{zeestraten2018programming}.
Additionally, their cubic complexity impedes scaling to large datasets.
Moreover, they are restricted to unimodal trajectory distributions.
Finally, policies are usually learned for a specific embodiment, with significantly fewer works studying cross-embodiment transfer~\cite{chen2024mirage}.

To address these problems, we propose Mixtures of Discrete-time Gaussian Processes (\ourmethod{}) as a powerful and flexible policy representation for imitation learning.
\finalchanges{While \cgp{}s define distributions over functions via kernels, discrete-time Gaussian Processes (\dtgp{}s) represent trajectories as finite sequences of multivariate Gaussians indexed by time, where each time step may encode structured variables such as robot poses. 
This discrete formulation avoids restrictive function-class assumptions and enables learning highly constrained behaviors from few demonstrations.}
\dtgp{} scale linearly with dataset size and have constant inference cost, enabling learning of complex manipulation tasks in under a minute using only a CPU.
In addition, \emph{Mixtures} of \dtgp{}s (\ourmethod{}) can represent multimodal trajectory distributions (see \figref{fig:overview}).

Due to their similarity to GMMs and \cgp{}s, many tools and extensions developed for these two methods can be applied to \ourmethod{} in a straightforward manner.
These include task parameterization~\cite{calinon2007learning}, automatic parameter selection~\cite{vonhartz2024art}, and the modeling of Riemannian data~\cite{zeestraten2018programming}.
We leverage all three in our extensive experiments.
Moreover, \ourmethod{} is probabilistic, enabling effective \emph{inference-time} updating based on additional evidence such as collision information.
For example, given the current scene configuration, \ourmethod{} can select and refine the appropriate mode from a multimodal behavior distribution. 
This process is illustrated in \figref{fig:evidence}.
However, not all constraints, particularly those arising from robot kinematics, can be readily expressed in end-effector space.
Some require path optimization to ensure kinematic feasibility.
To this end, we leverage \ourmethod{}’s probabilistic predictions for \vaporfull{}, which enforces kinematic validity.
We show that this enables effective cross-embodiment transfer of policies.

In summary, our main contributions are the following:
\begin{enumerate}
    \item We propose discrete-time Gaussian Processes\rebuttal{:} a powerful representation for unimodal robot policy learning.
    \item We introduce Mixtures of discrete-time Gaussian Processes (\ourmethod{)}\rebuttal{: a capable and efficient method for multimodal robot policy learning.}
    \item \rebuttal{We develop a powerful tool set to update the \ourmethod{} policy at inference time. We include convex evidence, such as workspace limits, and non-convex evidence, such as collision information.}
    \item \rebuttal{We show how to perform variance-aware path optimization (VAPOR) on \ourmethod{}.
    VAPOR ensures kinematic feasibility and enables effective cross-embodiment transfer.}
    \item We perform extensive experiments both in simulation and on a real robot, including highly constrained tasks, dynamic tasks, long-horizon tasks, and multimodal tasks.
    \item We make our code and models publicly available at \url{midigap.cs.uni-freiburg.de}.
\end{enumerate}
\section{Related Work}
\begin{table*}[ht]
    \centering
    \caption{Comparison of policy learning approaches.} 
    \vspace{-0.5em}
    \setlength{\tabcolsep}{5pt}
    \begin{threeparttable}
    \begin{tabular}{l c c c c c c c}
    \toprule
    \textbf{Method} & \makecell{\textbf{Sample} \\ \textbf{Efficient}} & \textbf{Interpretable} & \textbf{Scalable} &  \textbf{Probabilistic} & \makecell{\textbf{Multi} \\ \textbf{Stream}} & \makecell{\textbf{Highly} \\ \textbf{Expressive}} & \textbf{Multimodal}\\
    \midrule
    Diffusion Policy~\cite{chi2023diffusionpolicy} & \xmark & \xmark & \cmark & \xmark & \xmark & \cmark & \cmark \\
    RVT~\cite{goyal2023rvt}, ARP~\cite{zhang2024arp} & \cmark & \xmark & \cmark & \xmark & \xmark & \xmark & P \\
    Gaussian Mixture Models~\cite{calinon2006learning} & \cmark & \cmark & \xmark & \cmark & \cmark & \xmark & \xmark \\
    Continuous Gaussian Process~\cite{deisenroth2013gaussian} & \cmark & \cmark & \xmark & C & \xmark & \xmark & \xmark \\
    \textbf{Discrete-Time Gaussian Processes (Ours)} & \cmark & \cmark & \cmark & \cmark & \cmark & \cmark & \xmark \\
    \textbf{Mixture of Discrete-Time Gaussian Process (Ours)} & \cmark & \cmark & \cmark & \cmark & \cmark & \cmark & \cmark \\
    \bottomrule
    \end{tabular}
      \begin{tablenotes}[para,flushleft]
       \footnotesize      
       C: CGPs provide a certainty estimate, not a variability estimate.
       P: RVT and ARP use an additional language prompt to disambiguate multimodal tasks.
     \end{tablenotes}
   \end{threeparttable}
    \label{tab:related_work}
\vspace{-0.4cm}
\end{table*}
Imitation learning estimates the trajectory distribution suitable for solving a given task from a set of demonstrations.
We classify current imitation learning approaches by the following criteria:
\begin{enumerate*}
    \item their underlying representation,
    \item how they condition the model on environment observations,
    \item whether they are probabilistic,
    \item their prediction density, and
    \item whether they can represent multimodal trajectory distributions.
\end{enumerate*}

\para{Policy Representations:} Policy learning is a sequential decision problem, making sequence models such as LSTM~\cite{hochreiter1997long} a common choice~\cite{vonhartz2023treachery, chisari2022correct, florence2019self}.
Recently, Diffusion~\cite{janner2022planning, chi2023diffusionpolicy, rana2024affordancecentricpolicylearningsample} and Flow Matching~\cite{chisari2024learningroboticmanipulationpolicies, lipman2022flow, braun2024riemannian} models have become popular due to their favorable scaling properties.
Transformers~\cite{zhang2024arp, goyal2023rvt} have also been used.
However, deep learning methods typically require in the order of one hundred demonstrations to learn a task~\cite{vonhartz2023treachery, chi2023diffusionpolicy, zhang2024arp} and are challenging to interpret.
In contrast, Gaussian Mixture Models (GMM)~\cite{calinon2006learning, zeestraten2018programming, vonhartz2024art}, and continuous Gaussian Processes (\cgp)~\finalchanges{\cite{deisenroth2013gaussian, titsias2009variational, franzese2024generalization, joukov2017gaussian,  lang2017computationally, jaquier2020learning}} offer better interpretability and require only around five demonstrations to achieve good generalization~\cite{vonhartz2024art, deisenroth2013gaussian}.
However, in \secref{sec:background} we show they are less expressive.
Probabilistic Movement Primitives (ProMP)~\cite{paraschos2013probabilistic, paraschos2018using, rozo2022orientation} model a parameterized trajectory distribution.
They require in the order of 50 demonstrations~\cite{rueckert2015lowdim, brandi2014generalizing, zeestraten2018programming}, but this can be alleviated using factorization and variational inference~\cite{rueckert2015lowdim}.
Kernelized Movement Primitives (KMP)~\cite{huang2019kernelized} are a related few-shot method, but so far they are confined to Euclidean data~\cite{rozo2022orientation}.
Locally weighted regression (LWR)~\cite{kramberger2016generalization, calinon2016tutorial} is a non-parametric approach.
However, non-parametric approaches (including \cgp) tend to scale poorly with the dataset size~\cite{calinon2016tutorial}.
Finally, there are one-shot imitation methods, based on Dynamical Systems (Elastic-DS)~\cite{li2023task} and object-centric trajectories (DITTO)~\cite{heppert2024ditto}.
Yet, because they only leverage one demonstration, they need to make stronger assumptions regarding the trajectory distribution.
We propose using a discrete-time Gaussian Process (\dtgp), which retains the sample efficiency of GMMs and \cgp{}s, while offering better scalability and lower computational cost.
They further make milder assumptions regarding the trajectory distribution.

\para{Observation Conditioning:} Deep learning methods tend to be flexible, yet sample-inefficient, for they model the action distribution conditioned on some environment observations.
Different representations of environment observations have been used, such as images~\cite{chisari2022correct, chi2023diffusionpolicy}, keypoints~\cite{florence2019self, vonhartz2023treachery}, point clouds~\cite{chisari2024learningroboticmanipulationpolicies, Ze2024DP3, 3d_diffuser_actor}, 3D voxel grids~\cite{james2022coarse}, and object poses~\cite{vonhartz2024art, chi2023diffusionpolicy}.
Diffusion policy can effectively leverage image observations~\cite{chi2023diffusionpolicy}, and 3D point clouds~\cite{Ze2024DP3, 3d_diffuser_actor}.
However, it struggles to effectively leverage object pose observations~\cite{vonhartz2024art, chi2023diffusionpolicy}.
In contrast, multi-stream methods, such as Task-Parameterized GMM~\cite{calinon2006learning, zeestraten2018programming, vonhartz2024art} make effective use of object poses.
They do not directly condition the policy model on environment observations but instead learn multiple local object-centric models.
For this, they \emph{require} the environment state to be expressed as a set of coordinate frame poses.
However, in contrast to via-points, these can automatically be selected and extracted from visual observations~\cite{vonhartz2024art, chisari2023centergrasp, alizadeh2016identifying}.
Affordance-Centric Policy Decomposition (ACPD)~\cite{rana2024affordancecentricpolicylearningsample} similarly achieves sample-efficient generalization by modeling the trajectory distribution in a local coordinate frame. %
In contrast, DITTO blends two local trajectories using an exponential function~\cite{heppert2024ditto}, and Elastic-DS warps the trajectory to new endpoints via Laplacian editing~\cite{li2023task}.
ProMPs generalize to new task instances through conditioning of the joint density, which mostly limits them to interpolation~\cite{zeestraten2018programming}.
ProMP and KMP further support trajectory adaptation to via-points~\cite{paraschos2018using, rozo2020learning, huang2019kernelized}.
While there is an approach to frame transformation of ProMP~\cite{brandi2014generalizing}, it is limited to transforming the mean, not the full distribution.
This renders it inadequate for multi-stream learning.%

\para{Probabilistic Policies:} GMMs explicitly represent the learned trajectory distribution, including its variability~\cite{vonhartz2024art, deisenroth2013gaussian}.
In task-parameterized (multi-stream) learning, the variability at each point expresses the local relevance of the task parameters~\cite{calinon2006learning}.
In contrast, Gaussian Process Regression for \cgp{}s provides a certainty estimate, not a variability estimate~\cite{zeestraten2018programming}, making \cgp{}s unsuitable for multi-stream learning.
The absence of a variability estimate can be rectified by leveraging a heteroscedastic GP~\cite{arduengo2023gaussian}, \finalchanges{or a GMR-based GP~\cite{jaquier2020learning}}, or by modeling the variability using a second GP~\cite{monterolearning}. Only the latter has been adapted for multi-stream learning, and it requires a second training dataset to estimate parameter relevance~\cite{monterolearning}, which reduces sample efficiency.
Most deep learning methods provide only point estimates of the predicted actions~\cite{vonhartz2023treachery, chi2023diffusionpolicy}, making incorporating additional evidence into a learned policy challenging.
Diffusion Policy can be updated with additional evidence using guidance~\cite{ho2022classifier, ajay2022conditional, wang2024inference} and policy composition~\cite{wang2024poco}.
But so far, these methods come with high computational cost and limited success~\cite{wang2024poco, ajay2022conditional, wang2024inference}.
We find in \secref{sec:exp_constraints} that guidance can push Diffusion out of distribution, explaining at least part of the problem.
In contrast, as we discuss in \secref{sec:updating}, \ourmethod{}'s explicit representation of the trajectory distribution enables effective and efficient updating of the policy using additional evidence such as reachability and collision information.
As we discuss in \secref{sec:traj_opt}, it allows us to perform downstream trajectory optimization to ensure kinematic feasibility and even embodiment transfer.

\para{Prediction Density:} Deep learning methods are usually employed for dense, closed-loop control~\cite{vonhartz2023treachery, chisari2022correct}, or receding horizon control~\cite{chi2023diffusionpolicy, rana2024affordancecentricpolicylearningsample}.
In contrast, RVT achieves above-average sample efficiency by only sparsely predicting the next key pose combined with motion planning~\cite{goyal2023rvt}.
Yet, sparse predictions are limited with respect to constrained motions like opening a door.
Hierarchical approaches~\cite{xian2023chaineddiffuser} combine separate models for keypose prediction and dense trajectory prediction.
While GMMs can predict dense trajectories, they model the underlying trajectory distribution using a small number of Gaussian components~\cite{calinon2006learning}, usually fitted to locally near-linear segments of the trajectories~\cite{sun2023damm}.
When fitting Riemannian data, such as quaternions, in a task-parameterized manner, they can learn spurious correlations leading to faulty predictions~\cite{vonhartz2024art}.
Solving this issue using regularization severely limits the model's expressivity~\cite{vonhartz2024art}.
Gaussian Processes typically model a continuous trajectory distribution using kernel functions~\cite{deisenroth2013gaussian}.
This has two main drawbacks.
First, the training and inference complexities are cubic and quadratic in the number of training samples, respectively.
Second, the choice of the kernel functions severely limits the Gaussian Process's expressivity, see \secref{sec:background}.
While variational inference~\cite{titsias2009variational} alleviates the first issue, it aggravates the second one.
In contrast, we propose using discrete-time Gaussian Processes whose time density matches that of the demonstration trajectories.
In this way, the complexity of our approach scales linearly with the number of samples, and we model the trajectory distribution with sufficient time density to solve constrained motions.
At the same time, we make milder assumptions regarding the trajectory distribution, which we discuss in \secref{sec:background}.

\para{Multimodal Trajectory Distributions:} Neural networks typically struggle to learn multimodal target distributions~\cite{bishop1994mixture}, hence posing a grand challenge for deep imitation learning.
One of the key advantages of Diffusion Policy is its capability to overcome this hurdle~\cite{chi2023diffusionpolicy}.
However, this capability comes at the expense of a significant computational cost and requires a considerable number of demonstrations~\cite{chi2023diffusionpolicy}.
ProMPs cannot present multimodal distributions~\cite{rueckert2015lowdim} but can be extended using a mixture component~\cite{rueckert2015lowdim}.
Although the presented approach assumes a fixed and a priori known number of modes.
Similarly, \cgp{}s can only express unimodal distributions.
Multimodalities also pose a challenge to GMMs.
They only converge to the nearest \emph{local} optimum, which depends on the initialization and is hence not guaranteed to capture the multimodality~\cite{vonhartz2024art}.
We therefore propose to use Gaussian Process \emph{Mixtures}. 
In contrast to prior work~\cite{rueckert2015lowdim}, we propose an approach to automatically estimate the modes of the trajectory distribution from the set of demonstrations.

To the best of our knowledge, we are the first to leverage discrete-time Gaussian Processes and Mixtures of discrete-time Gaussian Processes for policy learning in robotics.
\tabref{tab:related_work} summarizes the differences from existing methods.
\section{Background}\label{sec:background}
\para{A Gaussian Mixture Model} (GMM) with \(K\) Gaussian components is defined as
\begin{equation}
    \mathcal{G}=\{\pi^k, \boldsymbol\mu_k, \boldsymbol\Sigma_k\}_{k=1}^K,
\end{equation}
where \(\boldsymbol\mu_k\) and \(\boldsymbol\Sigma_k\) denote the \(k\)-th component's mean and covariance matrix and \(\pi^k\) denotes its prior probability.
In imitation learning, a GMM \(\mathcal{G}\) models the joint distribution of a disjoint set of input variables \(\mathcal{I}\) and output variables \(\mathcal{O}\) as
\begin{equation}\label{eq:gmm}
    p(\mathcal{I}, \mathcal{O} \mid \mathcal{G}) = \sum_{k=1}^K \pi^k \cdot \mathcal{N}(\mathcal{I}, \mathcal{O}\mid \boldsymbol\mu_k, \boldsymbol\Sigma_k).
\end{equation}
For inference, Gaussian Mixture Regression~\cite{cohn1996active} then estimates the conditional density \(p_\mathcal{G} (\mathcal O\mid\mathcal I)\).

\begin{figure}
    \centering
    \begin{subfigure}[t]{\columnwidth}
        \includegraphics[width=\columnwidth]{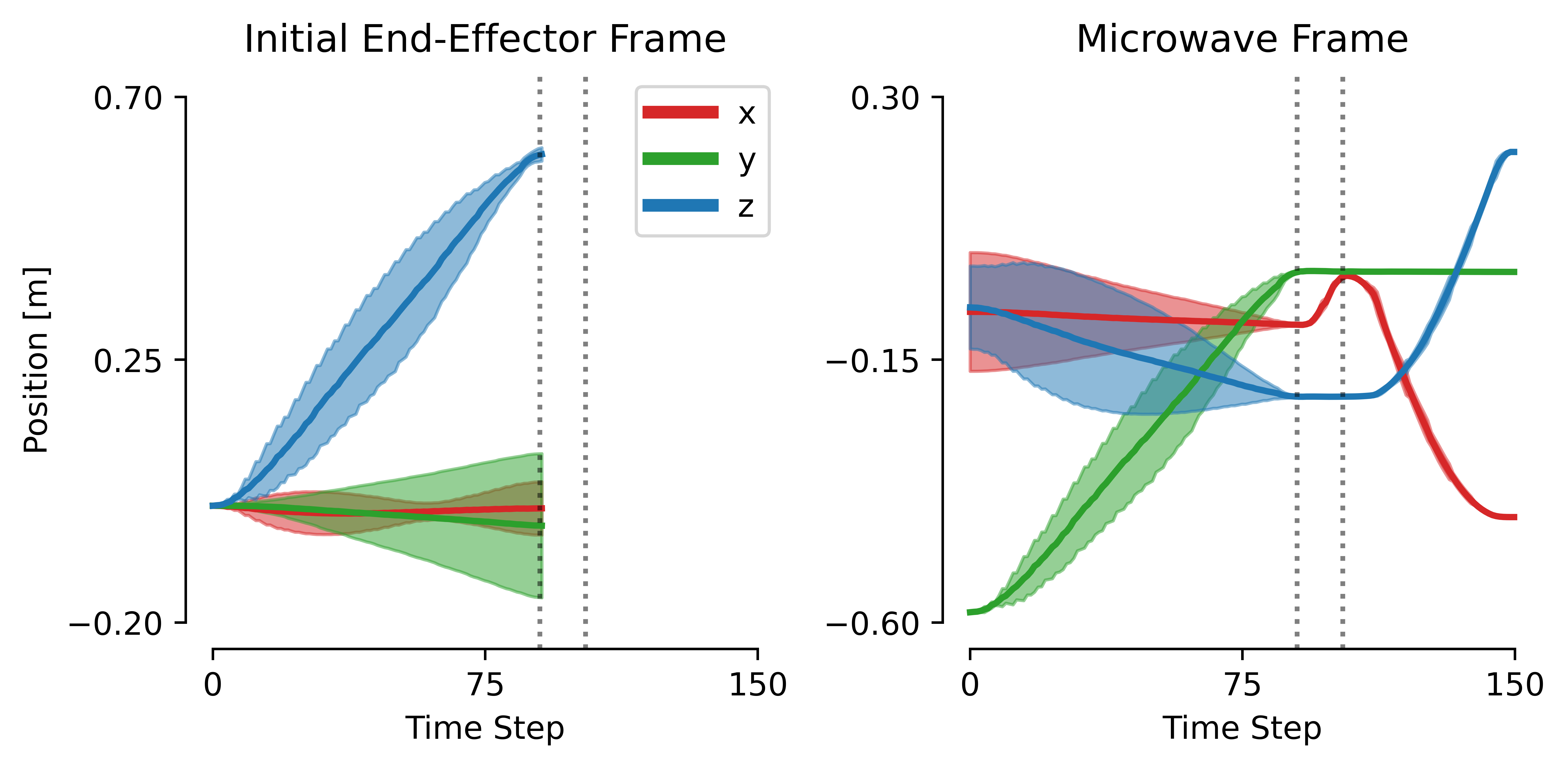}
        \caption{Per-frame discrete-time Gaussian Process in the task-parameterized setting. }\label{fig:tp_gp}
    \end{subfigure}\\ %
    \begin{subfigure}[t]{\columnwidth}
        \includegraphics[width=0.89\linewidth]{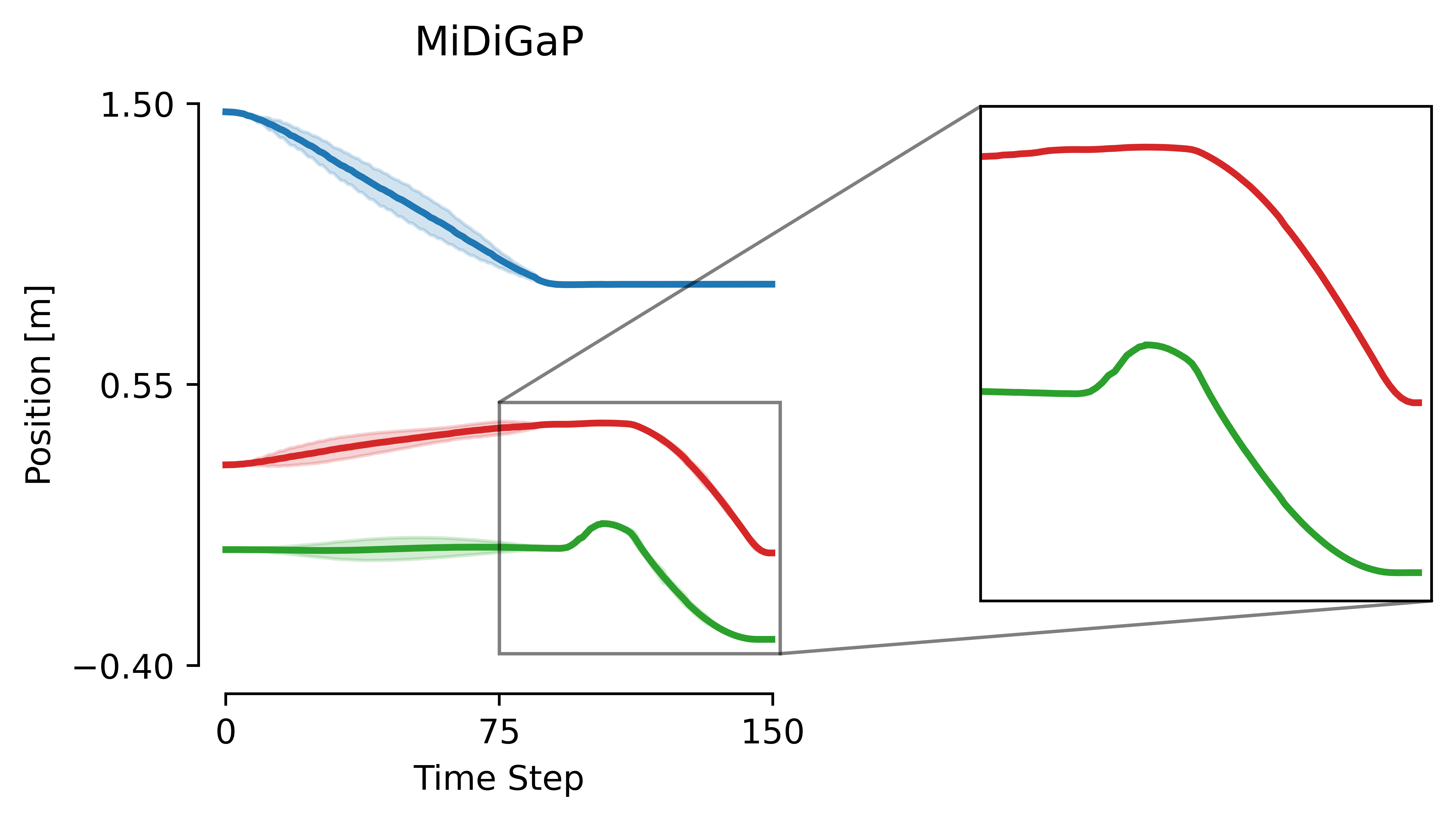}
        \caption{Joint model and prediction of the discrete-time Gaussian Process.}\label{fig:tp_gp_pred}
    \end{subfigure}\\ %
    \begin{subfigure}[t]{\columnwidth}
        \includegraphics[width=0.89\linewidth]{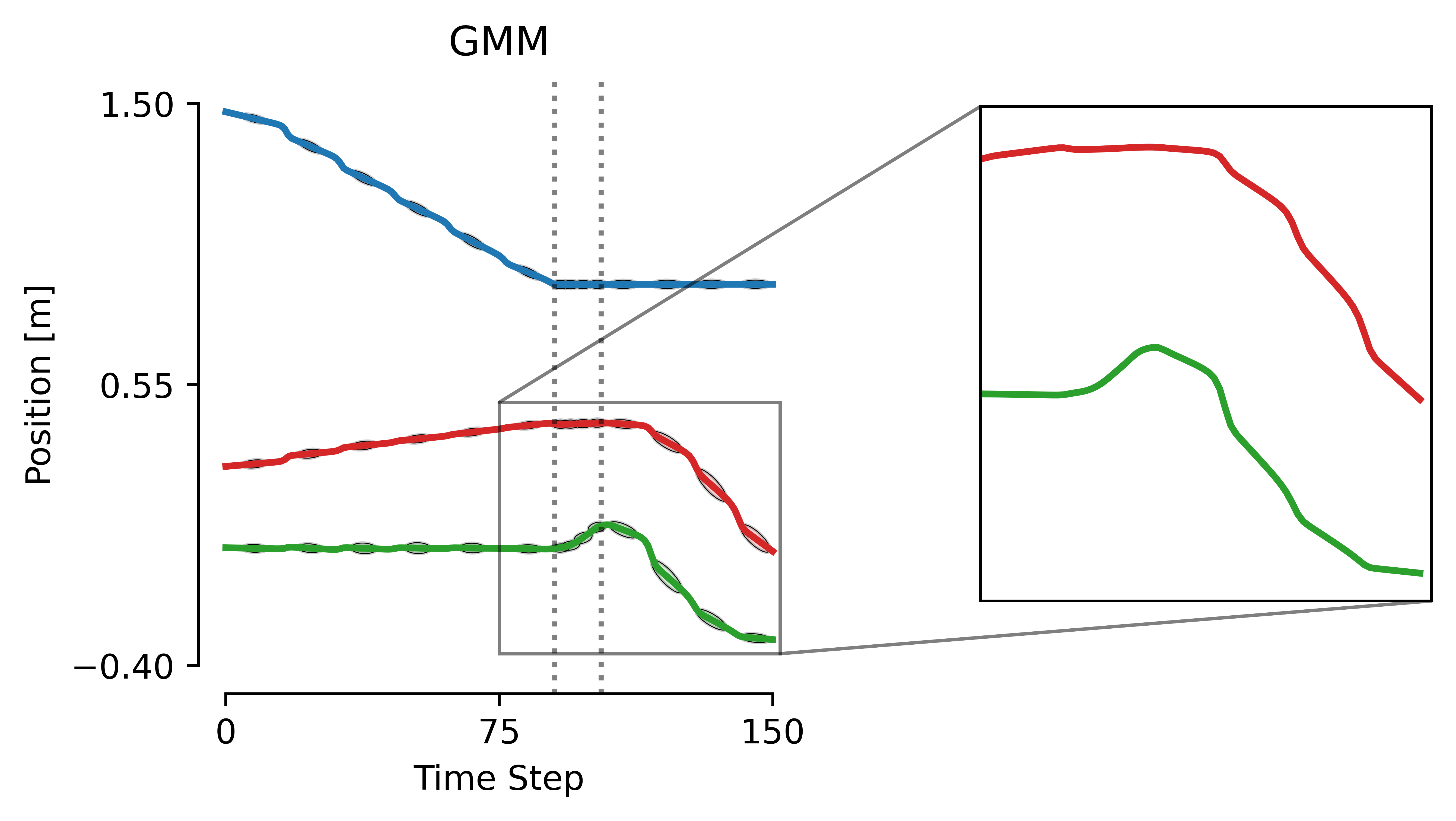}
        \caption{Joint model and prediction of the task-parameterized Gaussian Mixture Model.}\label{fig:tp_gmm_pred}
    \end{subfigure}
    \caption{Task-parameterized policies for the \texttt{OpenMicrowave} task depicted in \figref{fig:uni_tasks}.
    For \rebuttal{clarity}, we only plot the end-effector position.
    \subfiguresubref{fig:tp_gp} shows the local per-frame models for a discrete-time Gaussian Process.
    When the end-effector approaches the microwave, both the initial end-effector pose and the microwave's pose inform the trajectory.
    Afterwards, when grasping the handle and opening the microwave, only the microwave's pose determines the movement.
    TAPAS~\cite{vonhartz2024art} automatically finds these skill borders (vertical lines) and selects the relevant task parameters.
    For a specific task instance, we transform these local models into the world frame and combine them using the product of Gaussians.
    This process yields the joint model and prediction shown in \subfiguresubref{fig:tp_gp_pred}.
    \subfiguresubref{fig:tp_gmm_pred} plots a GMM's prediction for the same task instance.
    Note how the Gaussian Process predicts a smooth trajectory suited for opening the microwave's door, whereas the GMM does not.
    }\label{fig:tp_models}
\vspace{-0.5cm}
\end{figure}

\para{Task-Parameterized GMMs}~\cite{calinon2007learning} model end-effector motion relative to multiple coordinate frames, called \emph{task-parameters}.
These frames are typically anchored to objects, the robot, or the environment.
They can be automatically extracted from visual observations~\cite{vonhartz2024art}.
Given specific object poses, these local models are then transformed into world frame and combined.
\figref{fig:tp_models} illustrates this \emph{multi-stream} modeling.
Calinon~\textit{et~al.} gives an excellent introduction to task parameterization~\cite{calinon2007learning} and Zeestraten~\textit{et~al.} extends it to Riemannian data~\cite{zeestraten2018programming}.
GMMs are sample efficient and learn long-horizon tasks from only five demonstrations~\cite{vonhartz2024art}.
However, as \figref{fig:tp_gmm_pred} illustrates, their expressivity is limited by assuming local linearity.
Moreover, when modeling the covariance between data dimensions, they are prone to learning spurious correlations, which degrades prediction quality~\cite{vonhartz2024art}.

\begin{figure*}
    \centering
    \includegraphics[width=.9\linewidth]{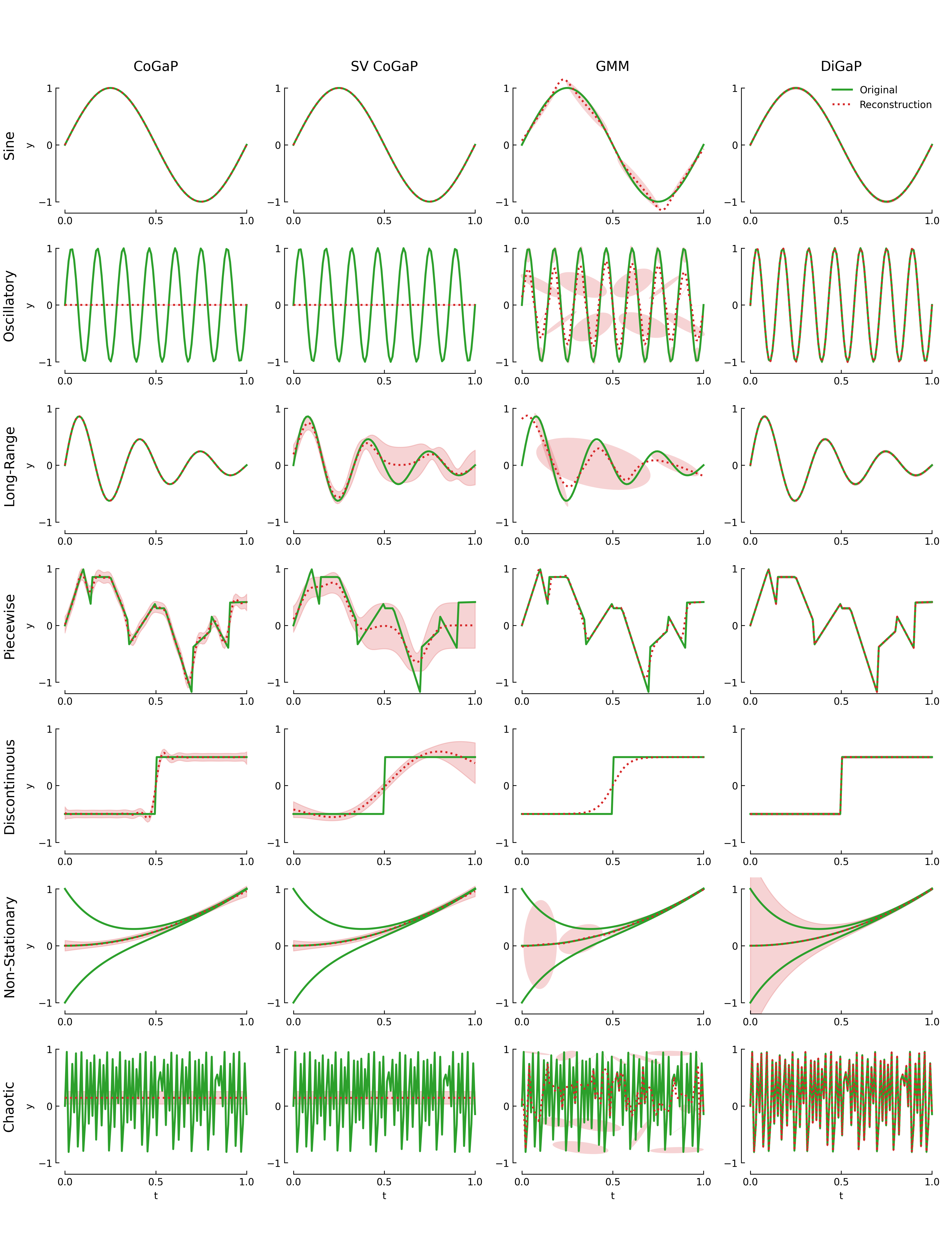}
    \vspace{-0.5cm}
    \caption{Limitations of Continuous Gaussian Processes (\cgp) and Gaussian Mixture Models (GMM).
    Shaded areas indicate the 95\% confidence intervals.
    The continuous GP and Stochastic Variational Gaussian Process (SV \cgp  ) were fitted with the RBF kernel and a manually tuned length scale of \(l=0.1\).
    Although the \text{Mat\'ern} kernel yields similar results on the presented functions, Gaussian Mixture Models tend to smooth the trajectories between components, and the model quality is highly dependent on its initialization.
    Here, we used a time-based initialization scheme~\cite{vonhartz2024art} and manually tuned the number of parameters.
    Kernel-based or continuous Gaussian Processes are well-suited for simple smooth functions, but struggle with periodicity (oscillation), non-stationarity, chaotic functions, and piecewise (linear) functions.
    In contrast, Discrete-time Gaussian Processes (\dtgp) can model all these function classes.
    }
    \label{fig:gp-kern}
\end{figure*}

\para{A continuous-time Gaussian Process} (\cgp) with domain \(\mani\) defines a distribution over functions, characterized by a mean function \(m:\mani\to\mathbb{R}\) and covariance function \(k:\mani\times\mani\to\mathbb{R}\), i.e.\
\begin{equation}
    f(\boldsymbol x)\sim\GP_\text{cont}\left(m(\boldsymbol x), k(\boldsymbol x, \boldsymbol x^\prime)\right).
\end{equation}
The mean \(m\) is typically set to zero~\cite{franzese2024generalization} and the expressivity of the \cgp{} is determined by the kernel function \(k\).
Popular options are the squared exponential kernel and the Mat\'ern kernel.
\figref{fig:gp-kern} shows that \cgp{}s struggle to model functions that are oscillatory, piecewise linear, discontinuous, non-stationary, chaotic, or have long-range dependencies.
For example, periodic (oscillatory) movements and long-horizon dependencies arise in household tasks such as stirring, scrubbing, and pouring liquids, as well as in industrial tasks such as screwing and polishing.
Moreover, piecewise movements occur in assembly tasks and when interacting with articulated objects.
Additionally, \cgp{}s suffer from cubic training and squared inference complexity, limiting their scalability.
Although they are designed to learn from (temporally) sparse data, this advantage diminishes in imitation learning, where dense sampling is readily available.
\cgp{}s are instead limited by their homoscedasticity, limited expressivity, and poor scalability.

We address these challenges by introducing a discrete-time Gaussian process that models trajectories densely and flexibly.
To this end, we make minimal assumptions: for any given task, there is a fixed frequency sufficient for sensing, modeling, and control.
In all our experiments, this frequency is \SI{20}{\hertz}, however, high-precision applications may require a higher frequency.
At this rate, data points are conditionally independent given their sequence position, and each is well-approximated by a Gaussian or a mixture thereof.
Observations are assumed to be noisy, yet unbiased.
Building on these minimal assumptions, we introduce a powerful, flexible, yet efficient imitation learning method.
\section{Technical Approach}\label{sec:approach}
We develop an imitation learning approach that is computationally inexpensive and sample-efficient, but at the same time models complex and multimodal tasks.
Additionally, our resulting policies can be updated in response to new scene information at inference time, such as new obstacles.

We structure our approach as follows.
We begin by modeling unimodal trajectory distributions using discrete-time Gaussian Processes  (\secref{sec:dtgp}).
As we show, these require as few as five demonstrations and surpass Gaussian Mixture Models in both expressivity and efficiency.
To capture multimodal behavior, we introduce modal partitioning to group demonstrations into subsets (see \secref{sec:mm_det}), each representing a mode of the trajectory distribution.
We then model the partition using Mixtures of Gaussian Processes (\secref{sec:gpm}).
To solve long-horizon tasks, we automatically segment demonstrations into shorter skills and then chain the corresponding models (\secref{sec:seq}).
Inference-time adaptation, e.g., for obstacle avoidance, is achieved via constrained Gaussian updating (\secref{sec:updating}).
Finally, we ensure joint-level kinematic consistency through \vaporfull{} (\secref{sec:traj_opt}). 

Techniques such as task parameterization~\cite{calinon2007learning}, automatic parameter selection~\cite{vonhartz2024art}, and modeling Riemannian data~\cite{zeestraten2018programming} are directly compatible with our approach. We refer the reader to prior work~\cite{calinon2006learning, zeestraten2018programming, vonhartz2024art} for details.

\subsection{Discrete-time Gaussian Process}\label{sec:dtgp}
A \emph{discrete-time Gaussian Process (\dtgp)} is a finite sequence of \(T\) Gaussian components
\begin{equation}
    \GP = \left(\left(\boldsymbol\mu_t, \boldsymbol\Sigma_t\right)\right)_{t=1}^T.
\end{equation}
Let
\begin{equation}\label{eq:data}
    \mathcal{D}=\{\left(\boldsymbol z_t^n\right)_{t=1}^T\}_{n=1}^N    
\end{equation}
be a dataset of \(N\) trajectories, each of length \(T\), where the data points \(\boldsymbol z_t^n\) lie on a Riemannian manifold \(\mani\) \(\left(\boldsymbol z_t^n\in\mani\right)\).
To achieve consistent length, we subsample a given set of demonstrations to its mean length.
In our experiments, we model the pose of the robot's end-effector \(\boldsymbol\xi=\begin{bmatrix}\boldsymbol x & \boldsymbol \theta\end{bmatrix}^T\) on the manifold \(\mani_{\text{pose}}=\mathbb{R}^3\times\mathcal{S}^3\).
We fit \(\GP\) to \(\mathcal{D}\) by estimating the empirical mean and covariance for each Gaussian component as
\begin{equation}\label{eq:gp_fit}
    \begin{aligned}
        \boldsymbol{\mu}_t &= \arg \min_{\mathbf{x} \in \mathcal{M}} \sum_{n=1}^N d_{\mathcal{M}}(\mathbf{x}, \boldsymbol{z}_t^n)^2,\\
        \boldsymbol{\hat\Sigma}_t &= \frac{1}{N - 1} \sum_{n=1}^N (\text{Log}_{\boldsymbol{\mu}_t}(\boldsymbol{z}_t^n))(\text{Log}_{\boldsymbol{\mu}_t}(\boldsymbol{z}_t^n))^\top,\\
        \boldsymbol{\Sigma}_{t} &= \operatorname{diag}(\boldsymbol{\hat\Sigma_t}).
    \end{aligned}
\end{equation}
Here, \( d_{\mathcal{M}}(\cdot, \cdot) \) denotes the geodesic distance on \(\mathcal{M}\) and \(\text{Log}_{\boldsymbol x}\) denotes \(\mathcal{M}\)'s logarithmic map applied at \(\boldsymbol x\).
For additional implementation details, please refer to~\cite{zeestraten2018programming, vonhartz2024art}.
\finalchanges{We optionally diagonalize the covariances, hence assuming conditional independence of the data dimensions.
While not strictly necessary, diagonalization prevents learning of spurious correlations between manifold dimensions~\cite{vonhartz2024art}, and simplifies downstream handling of the covariance matrices, see \eqref{eq:to_deriv_sig_diag}.
It also ignores the \emph{true} couplings between dimensions within each Gaussian, but at a sufficient temporal resolution this does not degrade policy performance.
In our experiments, a resolution of \SI{20}{\hertz} proved adequate across all tasks.
Alternative, adding a small constant (\(\lambda =1e^{-6}\)) to the diagonal \(\boldsymbol{\Sigma}_{t} = \boldsymbol{\hat\Sigma_t} + \lambda I\)  usually provides sufficient regularization for multi-stream learning.
}

Note further that we make no assumptions regarding the global smoothness or shape of the trajectory distribution.
In contrast, continuous-time Gaussian Processes need to define a mean and covariance function, thereby limiting their expressivity by the choice of function class.
Gaussian Mixture Models are also limited in their expressivity and tend to smooth the trajectories, which is not always desirable.
Conversely, our approach can model both smooth and erratic trajectories, if required, and is expressed in the demonstrations.
We discuss these limitations in detail in \secref{sec:background}.

For inference, we can sample any trajectory through \(\GP\), but we use the most likely trajectory given by \(\{\boldsymbol\mu_t\}_{t=1}^T\).
\figref{fig:tp_models} exemplifies a fitted \dtgp{} on the \texttt{OpenMicrowave} task, which is introduced in \figref{fig:uni_tasks}.
We evaluate the policy learning capabilities of discrete-time Gaussian Processes in \secref{sec:exp_uni_rlbench}.

\subsection{Modal Partitioning}\label{sec:mm_det}
The fitting procedure for discrete-time Gaussian Processes presented in \eqref{eq:gp_fit} is very efficient.
However, it assumes that the data distribution is unimodal.
To effectively model multimodal trajectory distributions, we estimate the modes of the data distribution prior to fitting a Mixture of Gaussian Processes.
In contrast to prior work~\cite{paraschos2013probabilistic}, we do not assume the number of modes to be known a priori.

\begin{figure}
    \centering
    \includegraphics[scale=0.5]{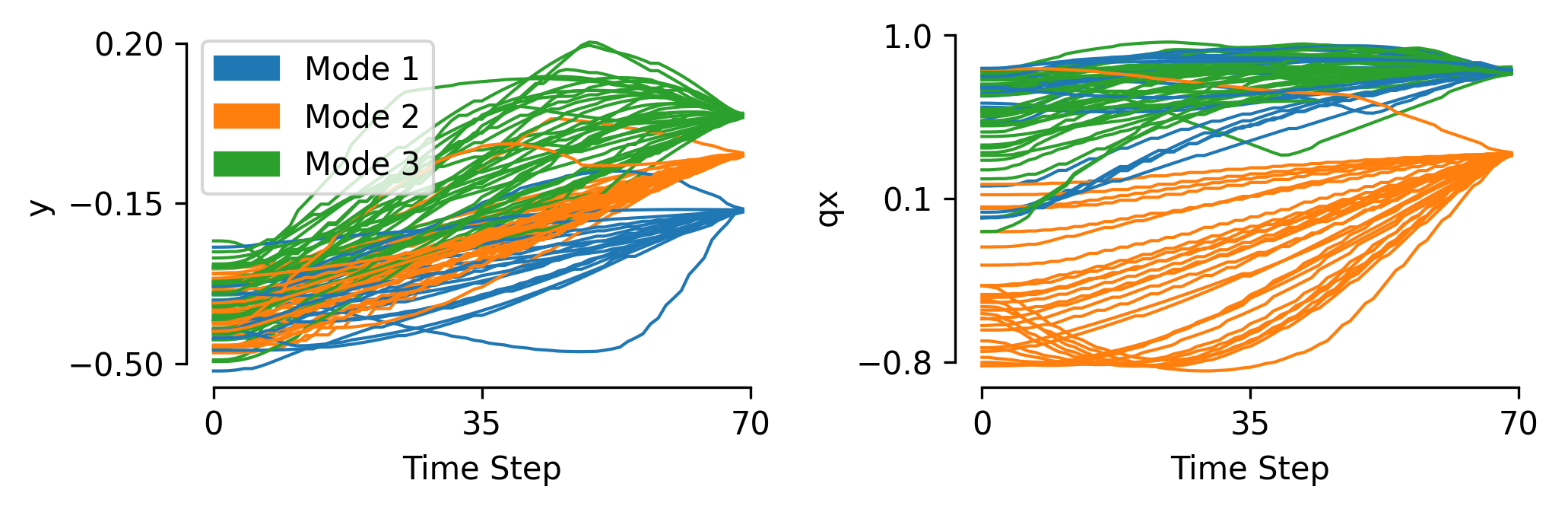}
    \caption{Modal partitioning of the \emph{mug placing} skill from the multimodal \texttt{PlaceCups} task.
    We plot two dimensions of the end-effector pose over time: the \(y\) position and the \(qx\) quaternion component.
    The data dimensions complement one another in differentiating the modes.
    For example, modes 1 and 3 are difficult to distinguish in \(qx\), but can be separated more easily in \(y\).
    }
    \label{fig:modal_part}
   \vspace{-0.4cm}
\end{figure}

Given a dataset \(\mathcal{D}\) of \(N\) trajectories, we compute a partition
\begin{equation}\label{eq:part_n}
    \{1,\ldots, N\} =\bigsqcup_{m=1}^M P_i
\end{equation}
of the \(N\) trajectories into \(M\) modes.
To this end, we first (optionally) subsample \(\mathcal{D}\) to a shorter length \(T^\prime\) to speed up the procedure.
For each trajectory \(n\), we then concatenate the sequence of \(T^\prime\) data points \(\{\boldsymbol z_1^n,\ldots\boldsymbol z_{T^\prime}^n\}\) into a single vector \(\boldsymbol v_n\).
For \(\boldsymbol z_t^n\in\mani\), this yields a set of vectors
\begin{equation}
    \mathcal{V} = \left\{\mathbf{v}_n \in \mani^{T^\prime}\right\}_{n=1}^N.
\end{equation}
\rebuttal{Clustering the full trajectories prevents the discovered modes from collapsing when fitting the modes’ trajectory distributions.}
We then cluster \(\mathcal{V}\) using one of the following three strategies.
\begin{enumerate}
    \item Fitting a series of Riemannian GMMs~\cite{zeestraten2018programming} with increasing number of components and selecting the best fitting GMM using the Bayesian Information Criterion (BIC).
    We diagonalize the GMM's covariance matrices, which both capture the diagonalization of the covariances in \eqref{eq:gp_fit} and suppress correlations between time steps.
    \item Performing Riemannian \(k\)-means with increasing number of components \(k\) and selecting the best model according to BIC.
    \item Applying density-based clustering (DBSCAN)~\cite{ester1996density} using the geodesic distance on \(\mani^{T^\prime}\).
\end{enumerate}
The cluster labels then yield the partition in \eqref{eq:part_n}.
\figref{fig:modal_part} shows an exemplary result.
The first two clustering approaches are more rigorous because they take the variance of the data into account, whereas DBSCAN uses a fixed distance threshold.
However, for DBSCAN the geodesic distances are only computed once.
In contrast, \(k\)-means and Expectation Maximization of Riemannian GMMs involve iterative Maximum Likelihood optimization~\cite{zeestraten2018programming}, rendering them computationally more expensive.
We evaluate and compare these strategies experimentally in \secref{sec:exp_multi_rlbench}.

\subsection{Mixture of Discrete-time Gaussian Processes}\label{sec:gpm}
A \emph{Mixture of Discrete-time Gaussian Processes (\ourmethod)} consists of \(M\) Gaussian Processes with prior weights \(\pi^m\in [0, 1]\) , i.e.\
\begin{equation}
    \GPM=\left\{\pi^m, \left(\boldsymbol\mu_t^m,\boldsymbol\Sigma_t^m\right)_{t=1}^T\right\}_{m=1}^M,
\end{equation}
where \(\sum_{m=1}^M \pi^m =1\).
For brevity, we also call the \ourmethod{}'s components its modes.
We fit \(\GPM\) to a dataset \(\mathcal{D}\), as defined in \eqref{eq:data}, by first performing the modal partitioning described in \secref{sec:mm_det}.
This yields a partition of \(\mathcal{D}\) with \(M\) parts
\begin{equation}
    \mathcal{D} = \bigsqcup_{m=1}^M \DM_m.
\end{equation}
We then instantiate a \ourmethod{} with \(M\) modes and fit each of its modes \(m\) to the corresponding part \(\DM_m\).
We define the prior \(\pi_m\) of a mode \(m\) as the fraction of demonstrations belonging to the associated part, i.e.\
\begin{equation}
    \pi^m=\frac{|\DM_m|}{|\mathcal{D}|}.
\end{equation}
We regress the \ourmethod{} by first sampling a mode \(m\) and then following the most likely trajectory \(\{\boldsymbol\mu_t^m\}_{t=1}^T\) under \(m\).
We evaluate the multimodal policy learning capabilities of \ourmethod{} in \secref{sec:exp_multi_rlbench}.

\begin{figure}
    \centering
    \includegraphics[scale=0.4]{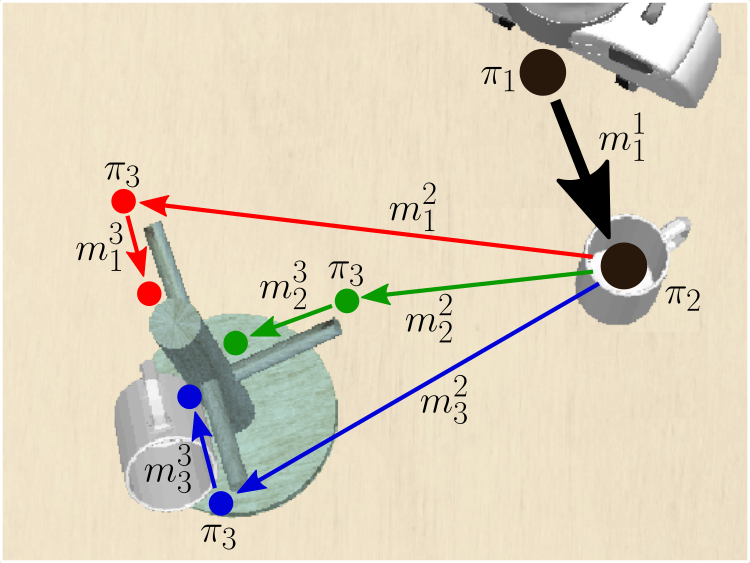}
    \caption{A sequence \(\{\GPM_1, \GPM_2, \GPM_3\}\) of three Gaussian Process Mixtures for solving the multimodal \texttt{PlaceCups} tasks.
    Arrows indicate the modes of the GPMs, nodes their start and end points.
    An arrow's or node's thickness indicates its likelihood under the sequence of GPMs.
    \(\GPM_1\) is unimodal, i.e.\ has prior \(\pi_1(m_1^1)=1\).
    \(\GPM_2\) has three modes \(m_1^2, m_2^2, m3_2\), each with prior \(\frac{1}{3}\).
    Therefore, sequencing yields \(\pi_2(1,j)=\frac{1}{3}\) for all \(j\in\{1,2,3\}\).
    \(\GPM_3\) has three modes as well.
    Here, we have \(\pi_3(j,j)=1\) for all \(j\in\{1,2,3\}\) and \(\pi_3(i,j)=0\) for all \(i\neq j\).
    }
    \label{fig:trans_model}
   \vspace{-0.3cm}
\end{figure}

\subsection{Segmenting and Sequencing of Skills}\label{sec:seq}
We model long-horizon manipulation tasks by automatically segmenting them into a series of atomic skills~\cite{vonhartz2024art}.
Each skill is modeled by a \ourmethod{}, and skill models are sequenced to solve long-horizon tasks.
Such sequences can either be learned from demonstrations or constructed later from previously learned skills~\cite{vonhartz2024art}.
\figref{fig:trans_model} shows an example of a task consisting of three skills.

If a dataset \(\mathcal{D}\) demonstrates a sequence of skills, we leverage \emph{Task Auto-Parametrization And Skill Segmentation} ({TAPAS})~\cite{vonhartz2024art} to automatically segment it into a sequence of demonstration sets \(\left(\mathcal{D}_1,\ldots,\mathcal{D}_n\right)\) to which we fit a sequence of  models \(\left(\GPM_1,\ldots,\GPM_n\right)\).
To sequence these models for inference, we define a transition model, overloading notation for \(\pi\).
For each \(j\in[1,\ldots,n-1]\) and for each pair of parts \(\DM_k\in\mathcal{D}_j\) and \(\DM_l\in\mathcal{D}_{j+1}\), we define the transition probability from mode \(k\) in model \(j\) to mode \(l\) in model \(j+1\) as
\begin{equation}
    \begin{aligned}
        \pi_1(k) &= \pi_1^k\\
        \pi_{j}(k, l) &= \frac{|\DM_k\cap\DM_l|}{|\DM_k|}.
    \end{aligned}
\end{equation}
The probability \(\pi\) of a modal path \(\left(m_1,\ldots,m_n\right)\) through a sequence of models \(\left(\GPM_1,\ldots,\GPM_n\right)\) is then given by
\begin{equation}\label{eq:modal_path_pi}
    \pi(m_1,\ldots,m_n) = \pi_1(m_1) \prod_{j=2}^n \pi_j(m_j, m_{j+1}).
\end{equation}
We regress over the sequence of models by sampling a modal path from \(\pi\) and following the most likely trajectory under this path.
Specifically, we concatenate the trajectory segments predicted by the selected mode of each model in the sequence, as described in \secref{sec:gpm}.
\figref{fig:trans_model} illustrates this cascading of Gaussian Process Mixtures.

To construct a novel skill sequence that was not demonstrated, we iteratively sequence two \ourmethod{}s using the KL-divergence between their Gaussian components~\cite{rozo2020learning}.
Given
\begin{equation*}
\begin{aligned}
    \GPM_i&=\left\{\pi^k, \{\boldsymbol\mu_t^k,\boldsymbol\Sigma_t^k\}_{t=1}^{T_i}\right\}_{\raisebox{0.7ex}{\scriptsize \(k=1\)}}^{\raisebox{-0.7ex}{\scriptsize \(K\)}},\\
    \GPM_j&=\left\{\pi^l, \{\boldsymbol\nu_t^l,\boldsymbol\Lambda_t^l\}_{t=1}^{T_j}\right\}_{\raisebox{0.7ex}{\scriptsize \(l=1\)}}^{\raisebox{-0.7ex}{\scriptsize \(L\)}},
\end{aligned}
\end{equation*}
for each pair of modes \(1\le k\le K\), \(1\le l\le L\), we compute
\begin{equation}
   \pi_{j}(k, l) \propto \exp\left(- D_{KL}\mathcal{N}(\boldsymbol\mu_{T_i}^k, \boldsymbol\Sigma_{T_i}^k)\parallel \mathcal{N}(\boldsymbol\nu_{0}^l, \boldsymbol\Lambda{0}^l)\right).
\end{equation}
The probability of a model path through the novel sequence of models is again given by \eqref{eq:modal_path_pi}.

\subsection{Constrained Gaussian Updating}\label{sec:updating}
Thus far, we have focused on learning and executing sequences of skills.
However, real-world deployment requires adapting these policies at \emph{inference-time} — for instance, to avoid novel obstacles.
Traditional approaches often lack mechanisms to incorporate such unforeseen evidence.
In contrast, our probabilistic framework enables effective inference-time updating.

\begin{figure}[tb]
    \centering
    \includegraphics[width=\columnwidth]{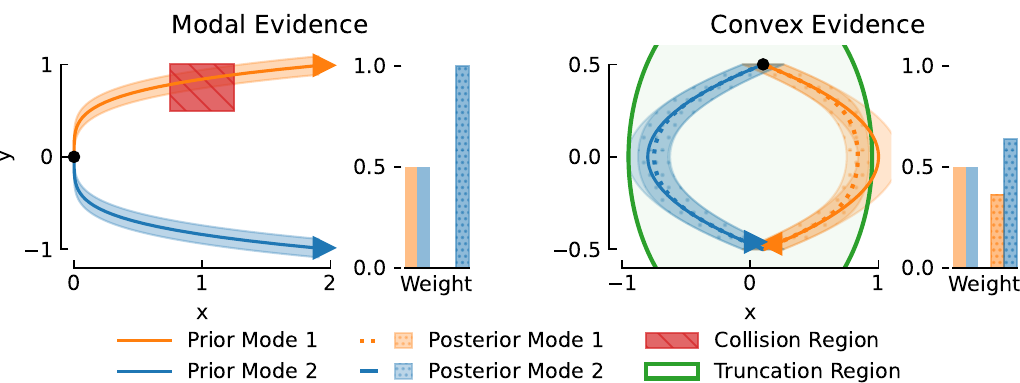}
    \caption{\textit{Left:} Modal evidence only updates the weights of the Gaussian Process Mixture.
    Here, the evidence is an unreachable region, leading to a collision.
    Under it, Mode 1 has zero probability.
    \textit{Right:} Convex evidence also updates the Gaussian parameters via moment matching.
    Here, the evidence approximates the robot's reachable workspace.
    The intersection of the prior (shaded area) and posterior (dotted area) indicates the likelihood of the evidence under the prior and is used to update the weight of the modes.
    }\label{fig:evidence}
   \vspace{-0.3cm}    
\end{figure}

In this section, we develop the tool set for applying additional evidence to \ourmethod{}.
As illustrated in \figref{fig:evidence}, we consider two forms of updating.
First, \emph{modal} updating adjusts the likelihoods of mixture components.
For example, if a mode is unreachable under the current scene configuration, its posterior likelihood should be zero.
Second, \emph{convex} updating adjusts the parameters of the Gaussian Processes themselves.
When only part of a mode’s probability mass lies outside the feasible region, the update shifts this mass within bounds.
To enable continual adaptation — e.g., in dynamic scenes — the posterior must also be a Mixture of Gaussian Processes.
If the evidence is Gaussian, this posterior can be computed analytically via the product rule. 
For more general cases, such as scalar bounds, the result is a constrained Gaussian, which we approximate via \emph{moment matching} to preserve mean and covariance.
When constraints are nonlinear or involve multivariate Gaussians, closed-form solutions are generally unavailable.
We therefore employ a Monte Carlo approximation scheme, which is also applicable to Riemannian Gaussians. 
We begin by defining moment matching for Riemannian Gaussians, then specify the class of constraints it supports, and finally explain its application to our Gaussian Mixture Process.

\para{Moment Matching:}
Let \(\mathcal{N}(\boldsymbol{\mu}, \boldsymbol{\Sigma})\) be a Gaussian prior over some Riemannian manifold \(\mathcal{M}\) with diagonal covariance \(\boldsymbol\Sigma\).
And let \(R\subseteq\mathcal{M}\) be the permissible region of a given constraint. %
To estimate the posterior \(\mathcal{N}(\boldsymbol{\mu}_R, \boldsymbol{\Sigma}_R)\) under \(R\), we sample \( \{ \mathbf{x}_i \}_{i=1}^N \sim \mathcal{N}(\boldsymbol{\mu}, \boldsymbol{\Sigma})\) and compute
\begin{equation}\label{eq:moment_matching}
\rebuttal{
    \begin{aligned}
        \mathbb{S}_R &= \{ \mathbf{x}_i \mid \mathbf{x}_i \in R, \; i = 1, \dots, N \},\\
        \boldsymbol{\mu}_R &\approx \arg \min_{\mathbf{x} \in \mathcal{M}} \sum_{\mathbf{x}_i \in \mathbb{S}_R} d_{\mathcal{M}}(\mathbf{x}, \mathbf{x}_i)^2,\\
        \boldsymbol{\hat\Sigma}_{R} &\approx \frac{1}{|\mathbb{S}_R| - 1} \sum_{\mathbf{x}_i \in \mathbb{S}_R} (\text{Log}_{\boldsymbol{\mu}_R}(\boldsymbol{x}_i))(\text{Log}_{\boldsymbol{\mu}_R}(\boldsymbol{x}_i))^\top,\\
        \boldsymbol{\Sigma}_R &= \operatorname{diag}(\boldsymbol{\hat\Sigma}_R).
    \end{aligned}}
\end{equation}
We further estimate the strength \(p_R\) of the truncation as the fraction of the probability mass of \(\mathcal{N}(\boldsymbol{\mu}, \boldsymbol{\Sigma})\) under \(R\), i.e.\
\begin{equation}\label{eq:trunc_strength}
    p_R(\boldsymbol \mu, \boldsymbol \Sigma) \approx \frac{|\mathbb{S}_R|}{N}.
\end{equation}
It serves as an estimate for the likelihood of \(R\) under the Gaussian, and we leverage it in \eqref{eq:pi_update} to update our prior weights over the modes of the \ourmethod{}.

\para{Convex Constraints:}
To ensure that moment matching provides a meaningful approximation of the constrained Gaussian, the feasible region \(R\) should be convex, yield a unimodal posterior, and avoid the distribution’s extreme tails. 
Moreover, constraints should not distort the end-effector trajectory in ways that compromise task execution.
For instance, while joint limits are convex in joint space, they can alter the end-effector path and thus threaten task success.
We implement the following \textbf{convex constraints}, which satisfy these criteria.
This list is neither exhaustive nor does it include all the constraints required for every task.
\begin{enumerate}
    \item \emph{Reachability} approximates the robot's workspace as a sphere around the robot's base position \(\boldsymbol b\) with radius \(r\)
    \begin{equation}\label{eq:reach_constr}
        R_\text{Reach} = \{ \mathbf{x} \in \mathbb{R}^3 \mid \lVert \mathbf{x} - \mathbf{b} \rVert \le r \}.
    \end{equation}
    \textmd{Example:} An object can be grasped from two sides, but one side is out of the robot's reach.
    \item \emph{Collision} evidence can be leveraged under some conditions.
    Let an obstacle \(O\) be given by its point \(\mathbf{p}\) closest to the end-effector, and by the normal vector \(\mathbf{n}\) pointing outward.
    We use a margin \(d_\text{uni}\) to ensure \(\boldsymbol p\) is far enough from the prior mean \(\mu\) for the posterior to be approximately unimodal.
    I.e.\ \(\mathbf{n}_i^T (\boldsymbol{\mu} - \mathbf{p}_i) \ge d_{\text{safe}} + d_\text{uni}\).
    If this is given, we enforce a safety distance \( d_{\text{safe}} \) via
    \begin{equation}\label{eq:convex-obst}
        R_\text{CC} = \{ \mathbf{x} \in \mathbb{R}^3 \mid \mathbf{n}^T (\mathbf{x} - \mathbf{p}) \ge d_{\text{safe}} \}.
    \end{equation}
    This constraint extends to a set of obstacles \( \{ O_1, O_2, \dots, O_k \} \), provided the feasible region remains sufficiently connected.
    For all \(1\le i,j\le k\) we require \(\| \mathbf{p}_i - \mathbf{p}_j \| \ge d_{\text{safe}} + d_\text{uni}\).
    The combined constraint is given by
    \begin{equation}
        R_\text{CCC} = \bigcap_{i=1}^k \{ \mathbf{x} \in \mathbb{R}^3 \mid \mathbf{n}_i^T (\mathbf{x} - \mathbf{p}_i) \ge d_{\text{safe}} \}.
    \end{equation}
    \textmd{Example:} If the scene is only sparsely occupied, this constraint performs some basic obstacle avoidance.
\end{enumerate}

\para{Modal Constraints:}
The following constraints may violate unimodality of the posterior or the convexity of the feasible region \(R\).
They are therefore unsuited for updating the Gaussian parameters.
However, these \textbf{modal constraints} remain useful for adjusting the prior weights of the Gaussian Process Mixture.
As before, this set of constraints is neither exhaustive nor required for every task.
\begin{enumerate}
    \item \emph{Heuristic self-collision} enforces a minimum distance \(d_\text{min}\) between the End-Effector and the robot's base
    \begin{equation}
        R_\text{HSC} = \{ \mathbf{x} \in \mathbb{R}^3 \mid \lVert \mathbf{x} - \mathbf{b} \rVert \ge d_\text{min}\}.   
    \end{equation}
    \textmd{Example:} An object can be grasped from two sides, but one side lies too close to the robot's base.
    \(R_\text{HSC}\) ensures the mode leading to self-collision is avoided.
    \item \emph{Scene object collision} avoidance leverages an occupancy representation of the scene, such as a voxel grid or point cloud, to avoid modes likely leading to collision.
    The occupancy likelihood function \(\mathrm{occ}(\mathbf{x}) : \mathbb{R}^3 \rightarrow [0, 1]\) can be estimated, for example, from a depth camera.
    Using some occupancy threshold \(\tau_\mathrm{occ}\), we then define
    \begin{equation}
        R_\text{SOC} = \{ \mathbf{x} \in \mathbb{R}^3 \mid \mathrm{occ}(\mathbf{x}) < \tau_\mathrm{occ}\}.   
    \end{equation}
    \textmd{Example:} A clutter object or some other agent blocks one of the modes of solving the current task.
    \(R_\text{SOC}\) ensures a feasible mode is chosen instead.
    \item {Kinematic feasibility checking} ensures that the predicted end-effector trajectory can actually be executed by the robot arm.
    In \secref{sec:traj_opt} we develop a feasibility checking approach that leverages the probabilistic structure of our policies.
    As shown in \secref{sec:exp_traj_opt} this even supports effective embodiment transfer.
\end{enumerate}

\para{Constrained Updating of Mixture of Gaussian Processes:}
Given a \ourmethod{} \(\left\{\pi^i, \{\boldsymbol\mu_t^i,\boldsymbol\Sigma_t^i\}_{t=1}^T\right\}_{\raisebox{0.7ex}{\scriptsize \(i=1\)}}^{\raisebox{-0.7ex}{\scriptsize \(N\)}}\), we apply a given \emph{convex} constraint \(R\) in two steps.
First, we exclude the tails of the distribution by testing for all modes \(m\) and time steps \(t\) if the the 95\% confidence interval (\(z=1.96\)) of the corresponding Gaussian \(\mathcal{N}(\boldsymbol\mu_t^m, \boldsymbol\sigma_t^m)\) intersects with \(R\).
Let \( d_{\mathcal{M}}\) denote the geodesic distance on \( \mathcal{M} \) and let \(\chi^2_{n, 0.95}\) be the critical value for the chi-squared distribution.
Then we test for
\begin{equation}
    \mathbbm{1}_{\{\mathbf{x} \in \mathcal{M} \mid d_{\mathcal{M}}(\mathbf{x}, \boldsymbol{\mu})^2 \leq \chi^2_{n, 0.95} \} \cap R \neq \emptyset}.
\end{equation}

E.g.\ for the reachability constraint in \eqref{eq:reach_constr}, we test for
\begin{equation}
\rebuttal{
    \text{reachable}(m) = \mathbbm{1}_{\left( \boldsymbol \mu_t^m - \boldsymbol b \right)^\top (\boldsymbol \Sigma_t^m)^{-1} \left( \boldsymbol \mu_t^m - \boldsymbol b \right) \le z^2}.}
\end{equation}
The likelihood \(\pi^m\) of a mode \(m\) failing this test is set to zero.
Second, we truncate all Gaussian components \(\mathcal{N}(\boldsymbol\mu_t^m, \boldsymbol \Sigma_t^m)\) with \(R\), and moment match the posterior according to \eqref{eq:moment_matching}. 
We then update the prior weight \(\pi^m\) of each mode \(m\) given the evidence \(R\) using the truncation strength \(p_R\) defined in \eqref{eq:trunc_strength}.
To estimate the truncation strength along the Gaussian Process \(m\), we use the normalized \(L_q\)-norm with \(q\in\mathbb{N}\).
This norm allows us to trade off the average truncation of the Gaussian components (\(q=1\)) with the maximum truncation of any Gaussian component (\(q\to\infty\)).
We calculate the posterior weight as
\begin{equation}\label{eq:pi_update}
    \tilde{\pi}^m \propto \pi^m \left(\frac{1}{T}\sum_{t=1}^T {p_R(\boldsymbol \mu_t^m,\boldsymbol \Sigma_t^m)}^q\right)^{1/q}.
\end{equation}
In our experiments, we use \(q=1\).

If \(R\) is just a \emph{modal} constraint, we only update the prior weights and skip the moment matching.
Finally, we re-normalize the prior \(\pi\).

Note that we do not apply any further smoothness constraints to the Gaussian Processes during moment matching.
Instead, the diagonal covariance structure and the convexity of \(R\) preserve the smoothness of the prior Gaussian Process - or its lack of smoothness.
This can be appreciated in \figref{fig:evidence}.

Recall from \secref{sec:seq} that for a sequence \(\left(\GPM_1,\ldots,\GPM_n\right)\) of models, \(\pi_{j}(k, l)\) encodes the transition probability from mode \(k\) in model \(\GPM_{j}\) to mode \(l\) in model \(\GPM_{j+1}\).
To update such a sequence of models, we use the modal evidence available for \(l\) to update all its incoming transition probabilities.
I.e.\ for all components \(k\) in \(\GPM_{j}\), we update \(\pi_{i}(k, l) \) according to \eqref{eq:pi_update}.

\subsection{\vapor: \vaporfull}\label{sec:traj_opt}
Manipulation policies are typically learned in end-effector space to generalize across varying object poses.
However, these trajectories must be converted into joint-space commands via inverse kinematics (IK) to control the robot.
For redundant arms, such as the Franka Emika, multiple IK solutions can exist for a given pose, and not all predicted end-effector trajectories are kinematically feasible across different task instances or embodiments.
Unlike methods that produce only a point estimate, MiDiGaP predicts probabilistic trajectories \( \{\boldsymbol\mu_t, \boldsymbol\Sigma_t\}_{t=1}^T\), where \(\boldsymbol \Sigma_t\)  encodes allowable \emph{variance} in each pose dimension. 
We leverage this information to guide IK: the optimizer is permitted to deviate from  \( \{\boldsymbol\mu_t\}_{t=1}^T\) in directions and magnitudes indicated by \(\{\boldsymbol \Sigma_t\}_{t=1}^T\), ensuring both task fidelity and kinematic feasibility.
Consider the \texttt{OpenMicrowave} task shown in \figref{fig:tp_gp_pred}.
While the end-effector is allowed to deviate from the most likely path when approaching the microwave, it cannot deviate as much once the handle is grasped without slipping off the handle.
This fact is captured precisely by the predicted variance \(\{\boldsymbol\Sigma_t\}_{t=1}^T\).

We name this approach \vapor: \vaporfull.
As we show in \secref{sec:exp_traj_opt}, \vapor{} even enables effective cross-embodiment transfer of learned \ourmethod{} policies.
In contrast, path optimization alone (without a probabilistic policy such as ours) is \emph{not} enough for solving many robot manipulation tasks.
We demonstrate this experimentally in \secref{sec:exp_uni_rlbench}; ARP, a method that only predicts a small set of key poses and uses motion planning in between, fails on constrained tasks, such as opening a door or wiping a desk.

We now formulate the \vapor{} optimization problem, beginning with standard IK and progressing to implementation and integration into our modal updating framework.

\para{Inverse Kinematics} takes a goal pose \(\boldsymbol\xi _{\text{des}}=\begin{bmatrix}\boldsymbol x_\text{des} & \boldsymbol \theta_\text{des}\end{bmatrix}^T\), a forward kinematics model \(\boldsymbol{\xi}(\cdot)\), joint limits \(\mathbf{q}_{\text{min}}, \mathbf{q}_{\text{max}}\), and a distance function \(d(\cdot,\cdot)\), and solves the optimization problem
\begin{align}\label{eq:ik_opt}
    & \min_{\mathbf{q}} & &  d\left(\boldsymbol\xi _{\text{des}}, \boldsymbol\xi(\mathbf{q})\right) \\
    & \textrm{s.t.} & & \mathbf{q}_{\text{min}} \leq \mathbf{q} \leq \mathbf{q}_{\text{max}}.
\end{align} 
Numerical IK solvers typically compute the 6D error vector
\begin{equation}\label{eq:6derror}
    \mathbf{e}_\mathbf{q} = 
    \begin{bmatrix}
    \mathbf{x}_{\text{des}} - \mathbf{x}(\mathbf{q}) \\
    \text{Log}(\boldsymbol \theta_{\text{des}}^{-1} \boldsymbol \theta(\mathbf{q}))
    \end{bmatrix},
\end{equation}
and iteratively minimize the quadratic loss function
\begin{equation}\label{eq:poseCost}
    d\left(\boldsymbol\xi _{\text{des}}, \boldsymbol\xi(\mathbf{q})\right) = \frac{1}{2} \mathbf{e}_\mathbf{q}^T\boldsymbol W \mathbf{e}_\mathbf{q}
\end{equation}
with diagonal weight matrix \(\boldsymbol W\), until a given threshold \(\tau\) is reached.
To make the IK variance-aware, we set \(\boldsymbol\xi _{\text{des}}=\boldsymbol\mu_t\) and \(\boldsymbol W=\boldsymbol\Sigma^{-1}_t\) for each time step \(t\).
With this cost function, minimizing the cost of the IK solution is equivalent to maximizing the IK solution's likelihood under \(\mathcal{N}(\boldsymbol\mu_t, \boldsymbol\Sigma_t)\).
This can be seen as a further form of Gaussian updating that takes into account the non-convex robot kinematics.
Yet, optimizing local IK solutions does not guarantee a continuous joint trajectory.
For this, we need to solve a global optimization problem.

\para{Path Optimization} simultaneously optimizes the IK solution for all time steps, also taking into account the continuity and smoothness of the joint trajectory.
We pose the optimization problem as
\begin{align}\label{eq:trajCost}
     & \min_{\mathbf{q}_1, \dots, \mathbf{q}_T} &&
     \sum_{t=1}^{T} \Tilde{d}\left(\boldsymbol\xi ^{\text{des}}_t, \boldsymbol\xi(\mathbf{q}_t)\right)
    + 
    \lambda_q
    \| \mathbf{q}_{t+1} - \mathbf{q}_t \|^2 \\
    & \textrm{s.t.} && \mathbf{q}_{\text{min}} \leq \mathbf{q}_t \leq \mathbf{q}_{\text{max}}, \quad\quad\ \forall 1\le t\le T,\label{eq:to_qbounds} \\
    &&& \rebuttal{\vert\mathbf{e}_{\mathbf{q}_t}\vert \le z \, \sqrt{\operatorname{diag}(\boldsymbol\Sigma_t)},  \quad \forall 1\le t\le T.}\label{eq:to_sigmabounds}
\end{align}
The cost function in \eqref{eq:trajCost} trades off the likelihood of the end-effector poses \(\{\boldsymbol\xi(\mathbf{q}_t)\}_{t=1}^T\) under the model against the smoothness of the resulting joint trajectory \(\{\boldsymbol q_t\}_{t=1}^T\).
Smoothness is achieved by penalizing the distance of consecutive joint positions $\boldsymbol q_t$ and $\boldsymbol q_{t+1}$.
The likelihood of the poses \(\{\boldsymbol\xi (\boldsymbol q_t)\}_{t=1}^T\) under the model is maximized by minimizing the pose cost in \eqref{eq:poseCost}.
We derive this objective in the next paragraph.
However, we found that naively minimizing \eqref{eq:poseCost} can incur arbitrarily high costs, making it challenging to balance the two objectives in \eqref{eq:trajCost}.
As a remedy, we normalize the variances \(\boldsymbol\Sigma_t\) by their largest value $\sigma_{\max} = \max_{t, i} \boldsymbol \Sigma_{t,i, i}$, i.e.
\begin{equation}\label{eq:modPoseCost}
    \Tilde{d}\left(\boldsymbol\xi ^{\text{des}}_t, \boldsymbol\xi(\mathbf{q}_t)\right) = \mathbf{e}_{\boldsymbol q_t}^T  \left(\frac{\boldsymbol\Sigma_t}{\sigma_{\max}}\right)^{-1} \mathbf{e}_{\boldsymbol q_t}.
\end{equation}
While \eqref{eq:modPoseCost} encourages closeness to \(\{\boldsymbol\xi _t^{\text{des}}\}_{t=1}^T=\{\boldsymbol\mu_t\}_{t=1}^T\), we additionally constrain the permissible deviation in \eqref{eq:to_sigmabounds}. %
This ensures that the optimized trajectory \(\{\boldsymbol\xi (\boldsymbol q_t)\}_{t=1}^T\) is likely under the predicted  distribution \( \{\boldsymbol\mu_t, \boldsymbol\Sigma_t\}_{t=1}^T\), therefore satisfying task constraints.
The factor $z$ determines the width of the imposed confidence interval.
As before, \eqref{eq:to_qbounds} enforces the joint limits, ensuring physical plausibility.

To solve the optimization problem, we use the forward kinematics and Jacobians provided by Kineverse~\cite{rofer2022kineverse} and apply the Augmented Lagrangian solver~\cite{toussaint2017tutorial}, which is well-suited for constrained problems.
To aid convergence, we initialize the optimizer with a linear interpolation between the initial joint configuration \(\boldsymbol q_0\) and the IK solution for the final target pose  $\boldsymbol \xi_T^{des}$.
The cost terms are balanced using $\lambda_e=1, \lambda_q=0.1$, and the deviation limit is set to $z=1.96$.

\para{Derivation of the optimization objective:}
We maximize the likelihood of a trajectory \(\boldsymbol{\Xi} = \{\boldsymbol{\xi}_t\}_{t=1}^T\) under a discrete-time Gaussian Process \(G=\{\boldsymbol\mu_t, \boldsymbol\Sigma_t\}_{t=1}^T\) by maximizing
\begin{equation}\label{eq:path_prob}
    \mathcal{L}(\boldsymbol\Xi; G) =  \prod_{t=1}^T f(\boldsymbol\xi_t; \boldsymbol{\mu}_t, \boldsymbol{\Sigma}_t).
\end{equation}
Due to the monotonicity of the logarithm, maximizing \eqref{eq:path_prob} is equivalent to minimizing its negative log-likelihood
\begin{equation}\label{trajopt_nll}
   \text{NLL}(\mathbf{\Xi}) = -\log \mathcal{L}(\boldsymbol{\Xi};G) = -\sum_{t=1}^T \log f(\boldsymbol\xi_t; \boldsymbol{\mu}_t, \boldsymbol{\Sigma}_t). 
\end{equation}
\rebuttal{The density function \(f\) of a \(k\)-dimensional Riemannian Gaussian contains a manifold volume term~\cite{said2017riemannian}.
Since it is constant with respect to our optimization objective, we omit it, yielding}
\begin{multline}
    f(\boldsymbol\xi_t; \boldsymbol{\mu}_t, \boldsymbol{\Sigma}_t) \propto \frac{1}{\sqrt{(2\pi)^k |\boldsymbol{\Sigma}_t|}} \\ \exp\left(-\frac{1}{2} \logmap_{\boldsymbol{\mu}_t}(\boldsymbol{\xi}_t)^T\boldsymbol\Sigma_t^{-1} \logmap_{\boldsymbol{\mu}_t}(\boldsymbol{\xi}_t)\right).
\end{multline}
Applying the logarithm to the PDF, we get
\begin{align}
    \begin{split}\label{eq:log_pdf_1}
        \log f(\boldsymbol\xi_t; \boldsymbol{\mu}_t, \boldsymbol{\Sigma}_t) \propto{}& -\frac{1}{2} \logmap_{\boldsymbol{\mu}_t}(\boldsymbol{\xi}_t)^T\boldsymbol\Sigma_t^{-1} \logmap_{\boldsymbol{\mu}_t}(\boldsymbol{\xi}_t)\\
             & - \frac{k}{2} \log(2\pi) - \frac{1}{2} \log |\boldsymbol{\Sigma}_t|
    \end{split}\\
    \propto{}& -\frac{1}{2} \logmap_{\boldsymbol{\mu}_t}(\boldsymbol{\xi}_t)^T\boldsymbol\Sigma_t^{-1} \logmap_{\boldsymbol{\mu}_t}(\boldsymbol{\xi}_t).
\end{align}
The terms \(\frac{k}{2} \log(2\pi)\) and \(\frac{1}{2} \log |\boldsymbol{\Sigma}_t|\) are independent of \(\boldsymbol{\xi}_t\), hence they can be omitted when optimizing  for \(\boldsymbol{\Xi}\).
Because all \(\boldsymbol\Sigma_t\) are diagonal, we have
\begin{align}\label{eq:to_deriv_sig_diag}
    \logmap_{\boldsymbol{\mu}_t}(\boldsymbol{\xi}_t)^T\boldsymbol\Sigma_t^{-1} \logmap_{\boldsymbol{\mu}_t}(\boldsymbol{\xi}_t) &= \sum_{i=1}^k \frac{\logmap_{\boldsymbol{\mu}_t}(\boldsymbol\xi_{t})_i^2}{\sigma_{t,i}^2},
\end{align}
where \(\sigma_{t,i}^2\) are the diagonal variances of \(\boldsymbol{\Sigma}_t\).
Therefore, our simplified NLL objective is given by
\begin{equation}\label{eq:to_deriv_final}
    C(\boldsymbol{\Xi}) = \frac{1}{2} \sum_{t=1}^T \sum_{i=1}^k \frac{\logmap_{\boldsymbol{\mu}_t}(\boldsymbol\xi_{t})_i^2}{\sigma_{t,i}^2}
    = \sum_{t=1}^{T} {d}\left(\boldsymbol\xi ^{\text{des}}_t, \boldsymbol\xi(\mathbf{q}_t)\right),
\end{equation}
when setting \(\boldsymbol{W}=\boldsymbol{\Sigma_t}^{-1}\) for all \(t\).
Hence, minimizing \eqref{eq:to_deriv_final} is equivalent to maximizing the likelihood of the trajectory \(\Xi\) under the predicted Gaussian Process \(G\).
The IK objective in \eqref{eq:poseCost} is just a special case with \(T = 1 \).

\para{Modal Updating: }\label{para:traj_opt_update}
We integrate \vaporfull{} into our Gaussian updating framework (\secref{sec:updating}) by treating the likelihood of an optimized trajectory along a modal path as modal evidence
Given an optimized joint trajectory \(\mathbf{q}_1^m, \dots, \mathbf{q}_T^m\), we compute the corresponding end-effector trajectory \(\boldsymbol\xi(\mathbf{q}_1^m), \dots, \boldsymbol\xi(\mathbf{q}_T^m)\).
Its likelihood under mode \(m\) with densities \(\{\boldsymbol\mu_t^m, \boldsymbol\Sigma_t^m\}_{t=1}^T\) is evaluated via \eqref{eq:path_prob} and the mode's likelihood is updated via \eqref{eq:pi_update}.

\section{Experimental Evaluation}
We evaluate our approach following the structure laid forth in \secref{sec:approach}.
\secref{sec:exp_uni_rlbench} examines unimodal, simulated policy learning:
How expressive is \ourmethod{}?
How well does it handle constrained movements? How sample- and computationally efficient is it, and how does it scale?
In \secref{sec:exp_multi_rlbench}, we turn to multimodal tasks, evaluating the effectiveness of different partitioning strategies and MiDiGaP's ability to model multimodal behaviors.
\secref{sec:exp_constraints} investigates inference-time updating: Can \ourmethod{} incorporate novel evidence, such as obstacles, while still achieving the task?
Can it generalize to an expanded task space?. 
\secref{sec:exp_traj_opt} assesses whether \vaporfull{} improves policy generalization across embodiments. Does performance stem from optimizing the end-effector or the joint trajectory? 
Does optimization help identify the most likely mode in multimodal tasks?. 
Finally, \secref{sec:exp_real} validates these findings in extensive real-world experiments.

\begin{table}
        \caption{Training and inference wallclock times across RLBench tasks.
        }\label{tab:inf_times}
        \centering
        \setlength{\tabcolsep}{3pt}
        \begin{threeparttable}
        \begin{tabular}{l c c c}
            \toprule
             & \multicolumn{2}{c}{\textbf{Training Duration}~[\unit{\hour}:\unit{\minute}]} & \multirow{2}{*}{\textbf{Inference}~[\unit{\milli\second}]} \\
             \cmidrule(lr){2-3}
            &  5 Demos & 100 Demos &  \\
            \midrule
            \textbf{\ourmethod{}  (Ours)} & \textbf{00:00 {--} 00:01} &  \textbf{\phantom{0}00:00\phantom{0} {--} \phantom{0}00:02\phantom{0}} &  \textbf{\phantom{00}1}\phantom{\textsuperscript*}\\
            └─ + \textbf{Updating} & - & - &  \textbf{\phantom{00}1}\phantom{\textsuperscript*}\\
            TAPAS-GMM~\cite{vonhartz2024art}  & 00:01 {--} 00:02 & \phantom{0}00:02\phantom{0} {--} \phantom{0}00:12\phantom{0}  & \phantom{00}5\phantom{\textsuperscript*}\\
            LSTM~\cite{hochreiter1997long} & 00:40 {--} 01:00 & \phantom{0}02:00\phantom{0} {--} \phantom{0}04:30\phantom{0} & \phantom{00}7\phantom{\textsuperscript*}\\
            Diffusion Policy~\cite{chi2023diffusionpolicy} & 01:00 {--} 05:00 & \phantom{0}07:00\phantom{0} {--} \phantom{0}72:00\phantom{0}  & \phantom{0}50\phantom{\textsuperscript*}\\
            └─ + ITPS~\cite{wang2024inference} & - & - & 200\phantom{\textsuperscript*}\\
            ARP~\cite{zhang2024arp} & 19:15 {--} 45:00 & 385:00\phantom{0} {--} 900:00\phantom{0}  & 200\textsuperscript* \\
            \bottomrule
        \end{tabular}
        \begin{tablenotes}[para,flushleft]
        \footnotesize
            Inference times are normalized by the policy's prediction horizon.
            Diffusion Policy, LSTM, and ARP are trained and rolled out on an A6000 GPU, while TAPAS-GMM and \ourmethod{} only use a CPU.
            The reported inference time of ARP ({*}) is for the keypose prediction without path-finding (and thus not normalized) and for a warmed-up model.
            Initial predictions take around 1\si{\second}.
            Dashes (-) indicate that inference-time updating does not affect training times.
         \end{tablenotes}
        \end{threeparttable}
      \vspace{-0.3cm}
\end{table}

\begin{figure*}[tb]
    \begin{tikzpicture}
        \def\imgwidth{0.2\textwidth}
        \def\imgheight{0.2\textwidth}
        \def\borderoffset{1pt}
        \def\borderwidth{2pt}
         \node[anchor=north west,inner sep=0pt] (firstrow) {
             \includegraphics[width=0.2\textwidth]{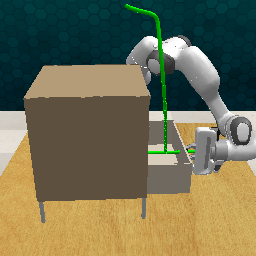}%
            \includegraphics[width=0.2\textwidth]{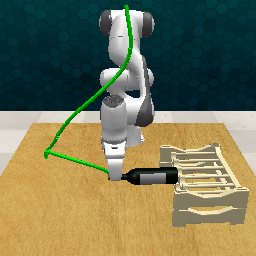}%
            \includegraphics[width=0.2\textwidth]{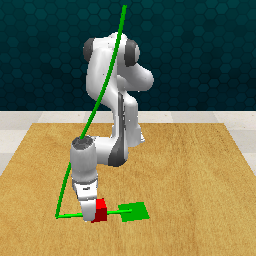}%
            \includegraphics[width=0.2\textwidth]{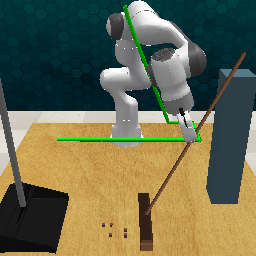}%
            \includegraphics[width=0.2\textwidth]{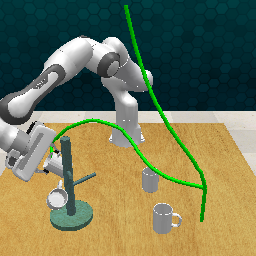}%
            };
            \node[anchor=north west,inner sep=0pt, below=\borderwidth of firstrow] (secondrow) {
            \includegraphics[width=0.2\textwidth]{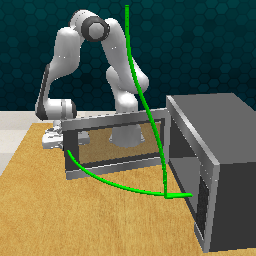}%
            \includegraphics[width=0.2\textwidth]{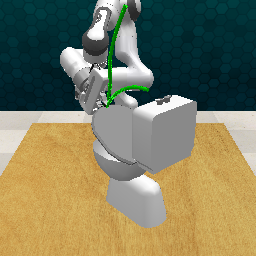}%
            \includegraphics[width=0.2\textwidth]{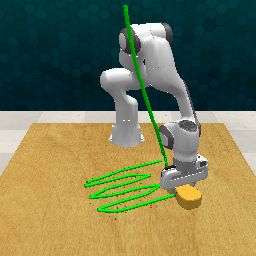}%
            \includegraphics[width=0.2\textwidth]{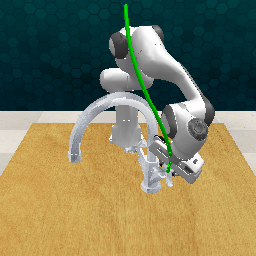}%
            \includegraphics[width=0.2\textwidth]{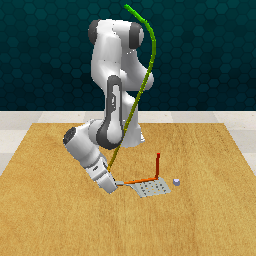}
         };
        \draw[RoyalBlue, line width=2pt] 
            (0, 0) -- (5*\imgwidth, 0)  -- (5*\imgwidth, -\imgheight) -- (0, -\imgheight) -- (0,0);
        \draw[CrimsonRed, line width=2pt] 
            (0, -\imgheight-\borderwidth) -- (5*\imgwidth, -\imgheight-\borderwidth) -- (5*\imgwidth, -2*\imgheight-\borderwidth) -- (0, -2*\imgheight-\borderwidth) -- (0, -\imgheight-\borderwidth);
    \end{tikzpicture}
    \caption{Unimodal RLBench tasks.
    \textit{Top} \textcolor{RoyalBlue}{(blue)}: The \textit{mildly constrained} tasks  \texttt{OpenDrawer}, \texttt{StackWine}, \texttt{SlideBlock}, \texttt{SweepToDustpan}, and \texttt{PlaceCups} can be solved using a sequence of linear movements.
    \textit{Bottom} \textcolor{CrimsonRed}{(red)}: In contrast, the  \emph{highly constrained} tasks include more complex object interactions (\texttt{OpenMicrowave}, \texttt{ToiletSeatUp} \texttt{TurnTap}), complex trajectory shapes (\texttt{WipeDesk}), or velocity constraints (\texttt{ScoopWithSpatula}).
    The \textcolor{Green}{green} lines indicate end-effector trajectories predicted by \ourmethod{}.
    For all tasks, object poses are randomized between task instances.
    }\label{fig:uni_tasks}
\end{figure*}
\begin{table*}
        \caption{Policy success rates on unimodal RLBench tasks. }\label{tab:success_rates_uni}
        \centering
        \setlength{\tabcolsep}{5pt}
        \begin{threeparttable}
        \begin{tabular}{ll cccccc cccccc}
            \toprule
            \multirow{2}{*}{\textbf{Demos}} & \multirow{2}{*}{\textbf{Method}} & \multicolumn{6}{c}{\textbf{Mildly Constrained Tasks}} & \multicolumn{6}{c}{\textbf{Highly Constrained Task}} \\
            \cmidrule(lr){3-8} \cmidrule(lr){9-14}
            & & \makecell{Open \\ Drawer} & \makecell{Stack \\ Wine} & \makecell{SlideBlock \\ ToTarget} & \makecell{SweepTo \\ Dustpan} & \makecell{Place \\ Cups} & Avg. & \makecell{Open \\ Microwave} & \makecell{Toilet \\ SeatUp} & \makecell{Wipe \\ Desk} & \makecell{Turn \\ Tap} & \makecell{ScoopWith \\ Spatula} & Avg. \\
            \midrule
            100 & Diffusion Policy~\cite{chi2023diffusionpolicy} & 0.88 & 0.95 & 0.21 & 0.90 & 0.40 & 0.67 & 0.92 & 0.58 & 0.06 & 0.82 & 0.02 & 0.48 \\
            & LSTM~\cite{hochreiter1997long} & 0.00  & 0.00 & 0.00 & 0.00 & 0.00 & 0.00 & 0.00 & 0.00 & 0.00 & 0.00 & 0.00 & 0.00 \\
            \cmidrule{1-14}
            5 & Diffusion Policy~\cite{chi2023diffusionpolicy} & 0.11 & 0.18 & 0.02 & 0.36 & 0.01 & 0.14 & 0.01 & 0.11 & 0.00 & 0.32 & 0.01 & 0.09 \\
            & ARP + RRTC\textsuperscript{*}~\cite{zhang2024arp} & 0.38 & 0.64 & 0.05 & 0.05 & 0.02 & 0.23 & 0.07 & 0.09 & 0.00 & 0.84 & 0.23 & 0.25\\
            & TAPAS-GMM\cite{vonhartz2024art} & 0.88 & \textbf{1.00} & 0.97 & 0.86 & 0.93 & 0.93 & 0.00 & 0.00 & 0.01 & 0.37 & 0.57 & 0.19 \\
            & \textbf{\ourmethod{} (Ours)} & \textbf{0.96} & \textbf{1.00} & \textbf{0.98} & \textbf{1.00} & \textbf{0.97} & \textbf{0.98} & \textbf{0.99} & \textbf{0.93} & \textbf{0.87} & \textbf{0.97} & \textbf{0.97} & \textbf{0.95} \\
            \bottomrule
        \end{tabular}
        \begin{tablenotes}[para,flushleft]
           \footnotesize  
           Bold values indicate the best-performing model for each task.
           The asterisk ({*}) indicates that ARP only predicts keyposes, not full trajectories. The trajectory to the next keypose is then generated using RRT-Connect~\cite{kuffner2000rrtc}, which has ground-truth access to the collision information of the scene.
         \end{tablenotes}
        \end{threeparttable}
\end{table*}
\begin{table*}
        \caption{Total positional end-effector acceleration per episode [$\si{\meter/\second^2}$] on unimodal RLBench tasks as mean and standard deviation. Lower is better.}\label{tab:ee_acc_uni}
        \centering
        \begin{threeparttable}
        \setlength{\tabcolsep}{5pt}
        \begin{tabular}{l c c c c c c c c c c}
            \toprule
            \multirow{2}{*}{\textbf{Method}} & \multicolumn{5}{c}{\textbf{Mildly Constrained Tasks}} & \multicolumn{5}{c}{\textbf{Highly Constrained Task}}\\
            \cmidrule(lr){2-6} \cmidrule(lr){7-11}
            & \makecell{Open \\ Drawer} & \makecell{Stack \\ Wine} & \makecell{SlideBlock \\ ToTarget} & \makecell{SweepTo \\ Dustpan} & \makecell{Place \\ Cups} & \makecell{Open \\ Microwave} & \makecell{Toilet \\ SeatUp} & \makecell{Wipe \\ Desk} & \makecell{Turn \\ Tap} & \makecell{ScoopWith \\ Spatula}\\
            \midrule
            Diffusion Policy~\cite{chi2023diffusionpolicy} & 414$\pm$638 & 256$\pm$259 & 558$\pm$436 & 313$\pm$382 & 494$\pm$420 & 213$\pm$\phantom{0}54 & 442$\pm$680 & 976$\pm$830 & \phantom{0}96$\pm$\phantom{0}79& 444$\pm$522 \\
            TAPAS-GMM~\cite{vonhartz2024art} & 346$\pm$\phantom{0}48 & 248$\pm$\phantom{0}18 & 222$\pm$\phantom{0}26 & 151$\pm$\phantom{0}38 & 529$\pm$114 & 417$\pm$140 & 273$\pm$\phantom{0}15 & 508$\pm$\phantom{0}21 & 300$\pm$\phantom{0}47 & 312$\pm$\phantom{0}56 \\
            \textbf{\ourmethod{} (Ours)} & \textbf{\phantom{0}61$\pm$\phantom{0}28} & \textbf{\phantom{0}84$\pm$\phantom{00}5} &  \textbf{\phantom{0}63 $\pm$\phantom{0}42} &  \textbf{\phantom{0}70$\pm$\phantom{00}4} & \textbf{231$\pm$166} & \textbf{146$\pm$\phantom{0}99} & \textbf{\phantom{0}85$\pm$\phantom{0}11} & \textbf{300$\pm$\phantom{0}26} & \textbf{\phantom{0}63$\pm$\phantom{0}44} & \textbf{116$\pm$\phantom{0}40}\\
            \bottomrule
        \end{tabular}
            \begin{tablenotes}[para,flushleft]
           \footnotesize 
           Results for Diffusion Policy are with 100 demonstrations.
           We forgo reporting LSTM performance, as it fails to solve the given tasks in the first place.
           We also do not report results for ARP as the trajectory cost is dictated by the used path planner, and not by ARP itself.
         \end{tablenotes}
        \end{threeparttable}
\end{table*}

\subsection{Unimodal Task-Parameterized Policy Learning}\label{sec:exp_uni_rlbench}
\para{Tasks: }
We evaluate the few-shot policy learning capabilities of discrete-time Gaussian processes (DiGaP) using two sets of RLBench tasks.
The first set comprises \textit{mildly constrained tasks}~\cite{vonhartz2024art}, such as \texttt{OpenDrawer} and \texttt{PlaceCups}.
While some of these tasks have long horizons or require precise grasps, they can all be solved by learning a sparse set of key poses~\cite{goyal2023rvt}, as they involve straight or unconstrained motions between object contacts. 
We therefore add a second set of \textit{highly constrained tasks}.
These include articulated object interactions (\texttt{OpenMicrowave}, \texttt{ToiletSeatUp},  \texttt{TurnTap}), densely constrained motion paths (\texttt{WipeDesk}), and dynamic tasks imposing constraints on the velocity profile (\texttt{ScoopWithSpatula}).
All tasks are shown in \figref{fig:uni_tasks}.

\para{Baselines:} 
In line with prior work~\cite{vonhartz2024art}, we compare our DiGaP against a set of deep learning methods.
These include the Diffusion Policy and an LSTM trained on 100 demonstrations.
We further compare against Auto-regressive Policy (ARP)~\cite{zhang2024arp}, the current incumbent transformer-based policy architecture.
Note that ARP only predicts a next keypose for the end-effector, not a dense trajectory.
Instead, it leverages the RRT-Connect planner~\cite{kuffner2000rrtc} with access to ground-truth collision information.
We further compare it against TAPAS-GMM~\cite{vonhartz2024art}, a state-of-the-art few-shot learning method, trained on five demonstrations.
To disentangle policy learning and representation learning, all methods receive ground-truth object pose observations.
We evaluate learning from RGB-D observations in \secref{sec:exp_real}.
ARP \emph{additionally} relies on four RGB-D cameras for 3D scene reconstruction.

\para{Metrics: }
We report task success rates over 200 evaluation episodes (\tabref{tab:success_rates_uni}) and quantify trajectory smoothness using the total positional acceleration of the end-effector (\tabref{tab:ee_acc_uni}).
We also measured the rotational acceleration, as well as jerk, which show analogous results and are omitted for brevity.

\para{Results: }
Diffusion Policy fails on all tasks when trained on five demonstrations and struggles even with 100 samples, particularly on tasks that require precision (\texttt{PlaceCups}), constrained movements (e.g., \texttt{WipeDesk}), or dynamic movements (\texttt{ScoopWithSpatula}).
As hypothesized, ARP fails on highly constrained tasks such as \texttt{OpenMicrowave}, because key pose prediction is insufficient when task success constrains the \emph{shape} of the trajectory.
More surprisingly, it also underperforms on many mildly constrained tasks, such as \texttt{SweepToDustpan} and \texttt{PlaceCups}.
Similar to Diffusion Policy, it seems to struggle with the trifecta of high variance, high precision, and data scarcity.
TAPAS-GMM performs well on mildly constrained tasks but fails under strong constraints.
We observe three key limitations:
 (1) limited Gaussian components can misalign gripper timing, (2) limited expressivity prevents accurate modeling of constrained motions (e.g., \texttt{OpenMicrowave}, see \figref{fig:tp_gmm_pred}), and (3) jerky motions due to assuming linear dynamics (see \figref{fig:tp_gmm_pred} and \tabref{tab:ee_acc_uni}).
In contrast, \ourmethod{} achieves strong performance across all task, benefiting from its dense trajectory modeling.
It consistently outperforms all baselines, even those trained with significantly more data.
On average, \ourmethod{} surpasses Diffusion Policy by 31 percentage points on the mildly constrained tasks and by 47 percentage points on the highly constrained tasks, despite using only 1/20th of the data.

\rebuttal{To assess whether Diffusion Policy can match \ourmethod{} provided substantially more data, we conduct scaling experiments with up to 400 demonstrations per task on \texttt{ToiletSeatUp} and \texttt{PlaceCups}.
Although performance tends to increase with more data, Diffusion Policy still fails to perform on par with \ourmethod{}.
It reaches a 79\% success rate on \texttt{ToiletSeatUp} and 40\% on \texttt{PlaceCups}.
Full scaling curves are available on our project website.
The project website further reports additional RoboSuite experiments, which indicate that \ourmethod{}’s performance is robust to the choice of simulator.
}

Furthermore, \ourmethod{} produces significantly smoother trajectories.
\tabref{tab:ee_acc_uni} shows a 67\% reduction in trajectory cost compared to TAPAS-GMM.
The improvement is even more pronounced relative to the Diffusion Policy.
Qualitative examples are shown in our supplementary video.
Compared to TAPAS-GMM, \ourmethod{} also improves success rates by five percentage points on mildly constrained tasks and by 76 points on highly constrained tasks, all with the same amount of demonstration data.

\begin{figure*}[tb]
    \centering
            \includegraphics[width=0.2\textwidth]{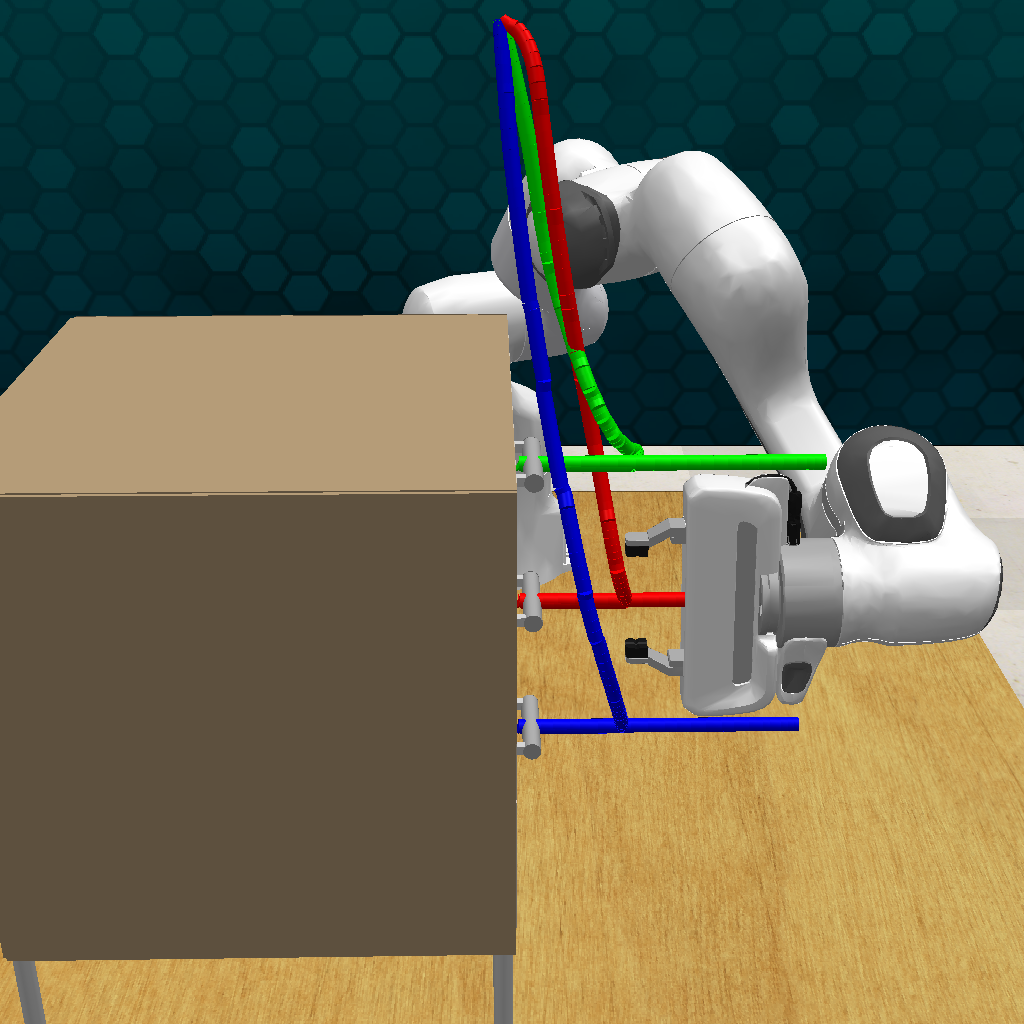}%
            \includegraphics[width=0.2\textwidth]{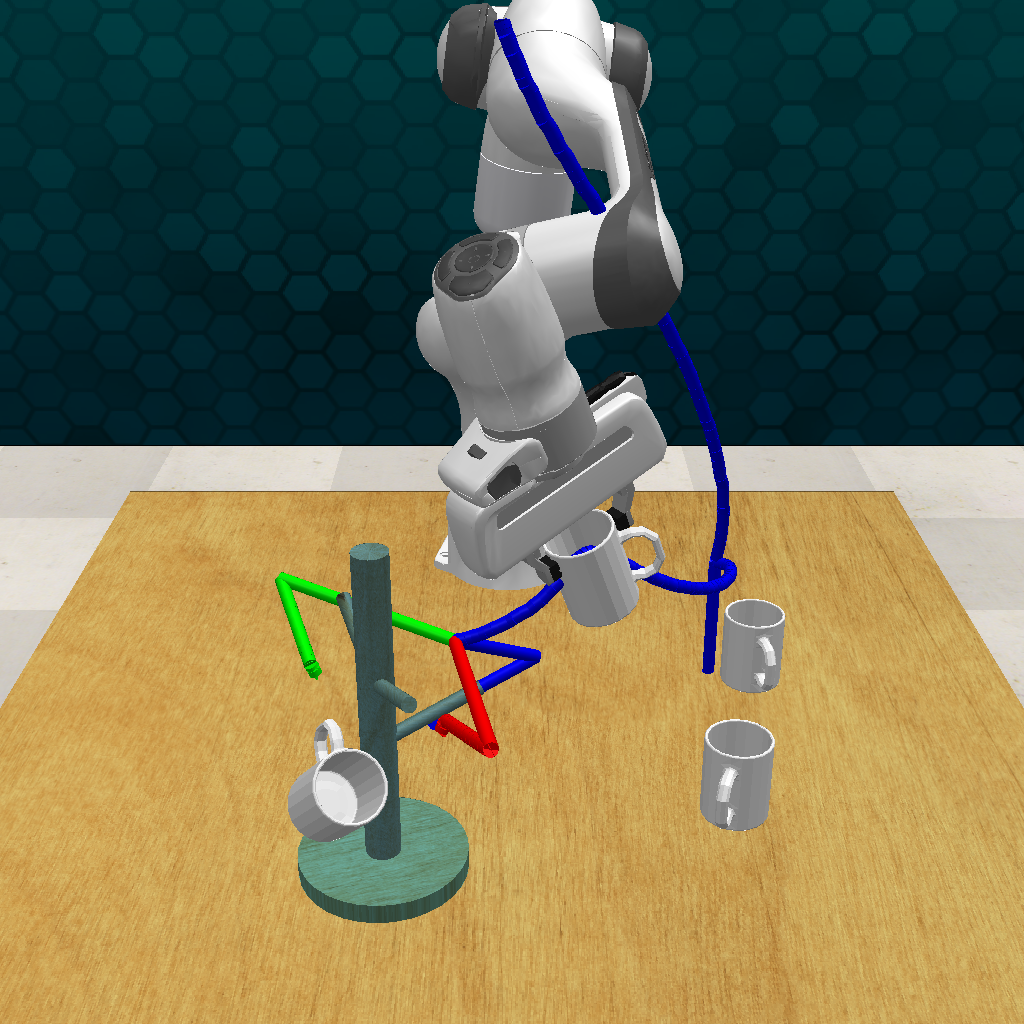}%
            \includegraphics[width=0.2\textwidth]{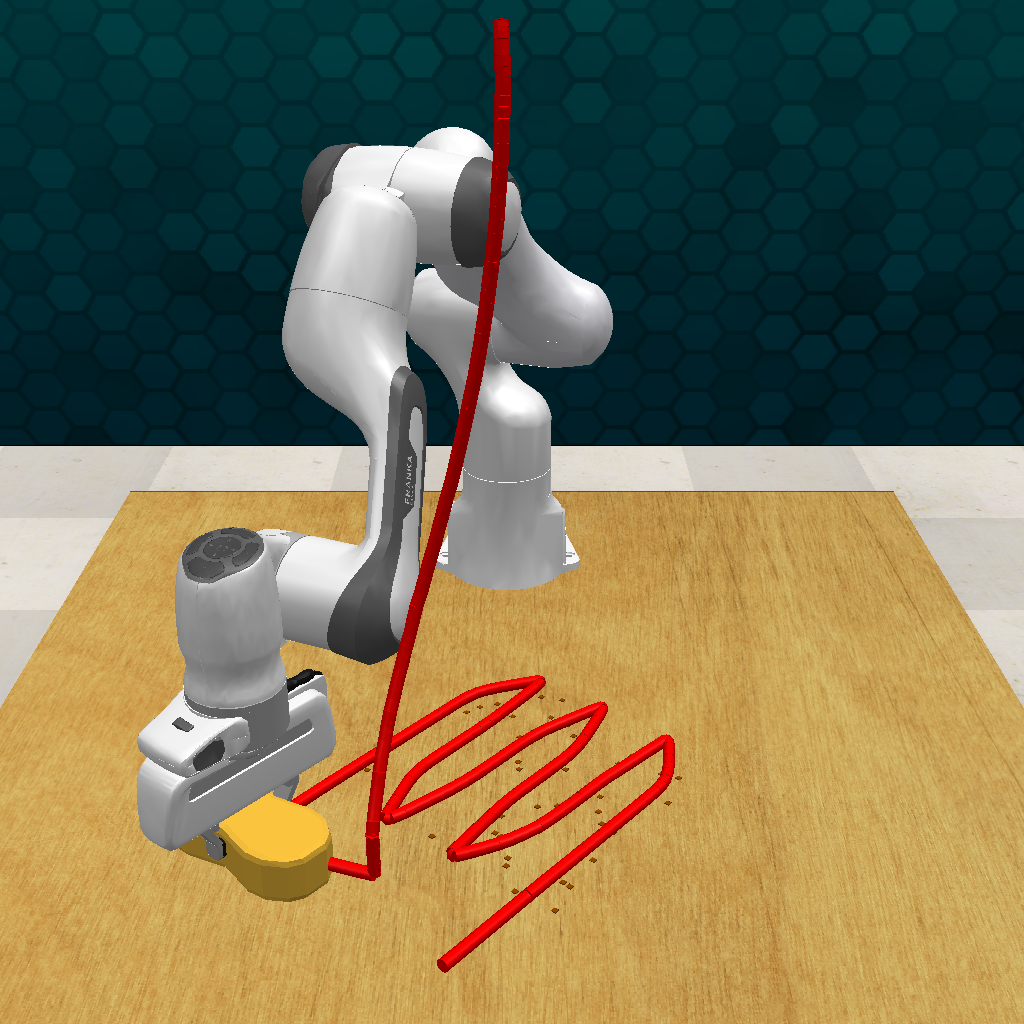}%
            \includegraphics[width=0.2\textwidth]{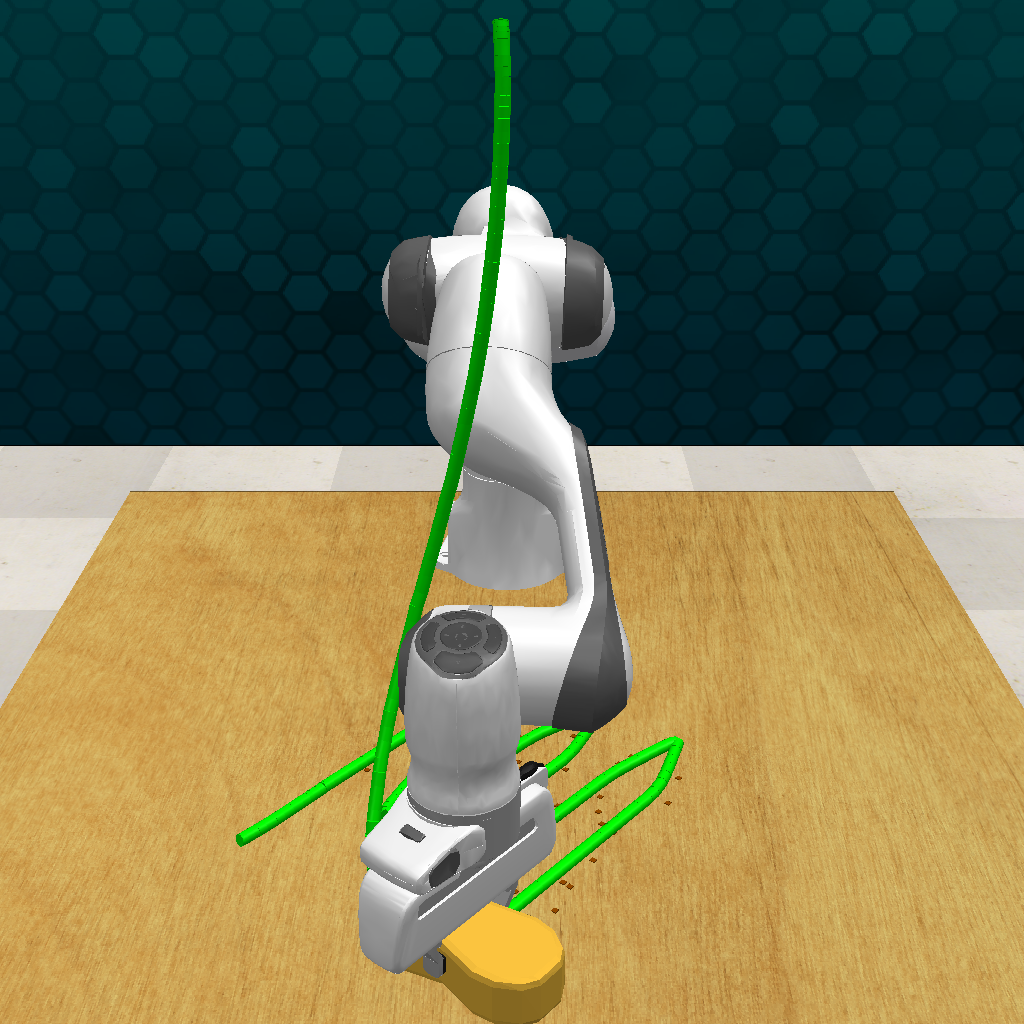}%
            \includegraphics[width=0.2\textwidth]{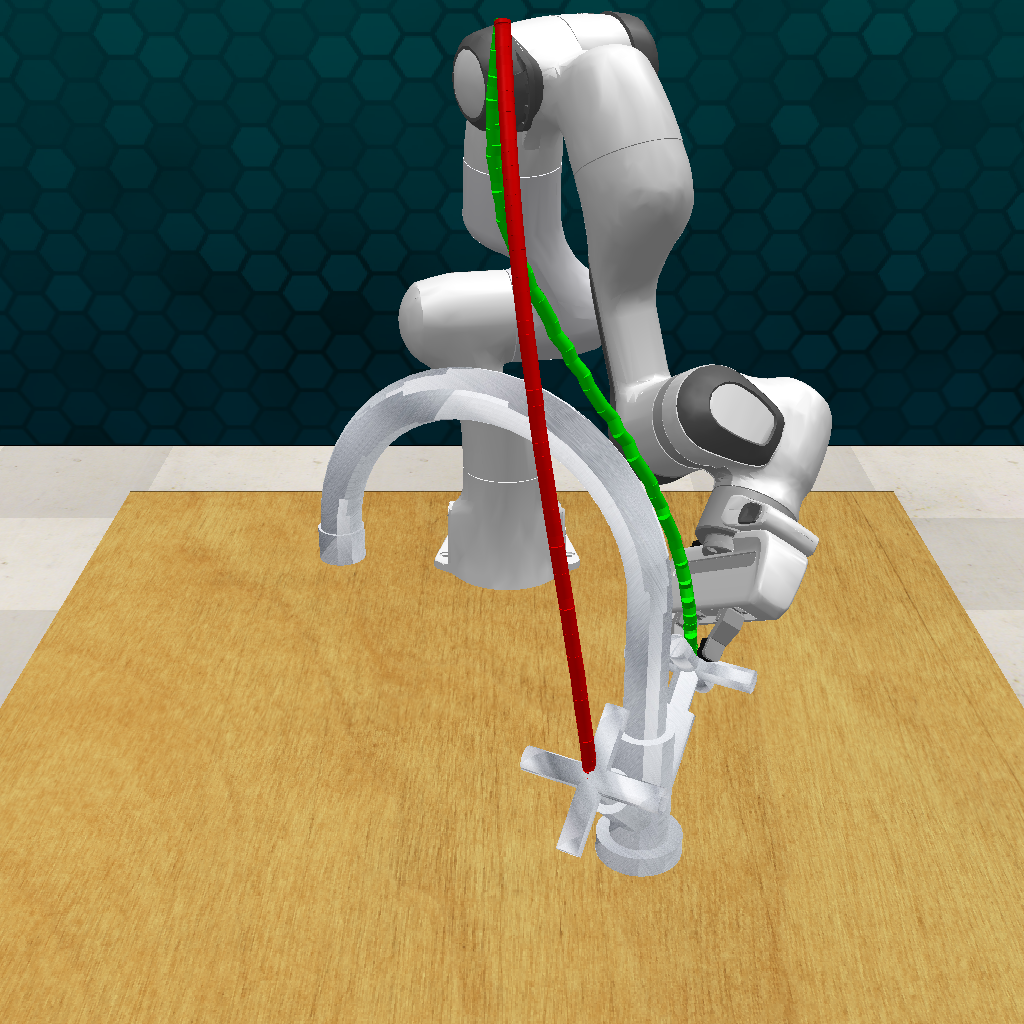}%
    \caption{Multimodal RLBench tasks.
    In \texttt{OpenDrawer} either of the cabinet's three drawers may be opened, and  in \texttt{PlaceCups} the cup can be placed on either of the three free spokes.
    In \texttt{WipeDesk}, the desk may be wiped in either direction, and in \texttt{TurnTap}  either of the two valves can be turned.
    The colored lines indicate one suitable end-effector trajectory per mode, predicted by \ourmethod.
    The object poses are randomized between task instances.
    }\label{fig:multi_tasks}
   \vspace{-0.3cm}
\end{figure*}

Finally, \tabref{tab:inf_times} summarizes training and inference times for all evaluated policy models.
\ourmethod{} is not only more accurate but also markedly more efficient.
Compared to Diffusion Policy, training is three \emph{orders of magnitude} faster, and inference is about 50 times faster.
And this speedup is despite our method running on a CPU, while Diffusion \rebuttal{Policy} uses an A6000 GPU.
Even compared to the low-cost TAPAS-GMM, both training and inference of \ourmethod{} are about five times faster.

\begin{table}
        \caption{Wall clock time [\SI{}{\second}] for partitioning of multimodal RLBench tasks. }\label{tab:modal_part_times_multi}
        \centering
        \setlength{\tabcolsep}{4pt}
        \begin{threeparttable}
        \begin{tabular}{l c c c c }
            \toprule
            \multirow{2}{*}{\textbf{Task}} & \multirow{2}{*}{\textbf{Demos}} & \multicolumn{3}{c}{\textbf{Method}}\\
            \cmidrule(lr){3-5}
            & & GMM-EM~\cite{zeestraten2018programming} & \(k\)-Means~\cite{steinhaus1957} & DBSCAN~\cite{ester1996density} \\
            \midrule
            OpenDrawer & \phantom{0}25 & 1.37 & \textbf{0.34}& 0.03\textsuperscript{\textdagger}\\
            \midrule
            WipeDesk & \phantom{0}25 & 0.98 & 0.37 & \textbf{0.04}\phantom{\textsuperscript{\textdagger}}\\
            \midrule
            TurnTap & \phantom{0}25 & 61.5 & 2.78 & \textbf{0.04}\phantom{\textsuperscript{\textdagger}} \\
            & \phantom{0}50 & 69.3 & 3.99 & \textbf{0.10}\phantom{\textsuperscript{\textdagger}} \\
            & 100 & 66.4 & 3.31 & \textbf{0.41}\phantom{\textsuperscript{\textdagger}} \\
            \midrule
            PlaceCups & \phantom{0}25 & 123 & 6.42 & \textbf{0.09}\phantom{\textsuperscript{\textdagger}}\\
            & \phantom{0}50 & 125 & 12.6 & \textbf{0.32}\phantom{\textsuperscript{\textdagger}} \\
            & 100 & 233 & 24.1 & \textbf{0.67}\phantom{\textsuperscript{\textdagger}} \\
            \bottomrule
        \end{tabular}
          \begin{tablenotes}[para,flushleft]
           \footnotesize  
           Partitioning times are summed up over the multiple skills of a task.\\
           A dagger (\textdagger) indicates partial failure to correctly partition.
         \end{tablenotes}
        \end{threeparttable}
\end{table}
\begin{table}
        \caption{Policy success rates on multimodal RLBench tasks. }\label{tab:success_rates_multi}
        \centering
        \setlength{\tabcolsep}{4pt}
        \begin{tabular}{l l c c c c c}
            \toprule
            \multirow{3}{*}{\textbf{Demos}} & \multirow{3}{*}{\textbf{Method}} & \multicolumn{5}{c}{\textbf{Multimodal Tasks}}\\
            \cmidrule(lr){3-7}
            & & \makecell{Open \\ Drawer} & \makecell{Place \\ Cups} & \makecell{Turn \\ Tap} & \makecell{Wipe \\ Desk} & Avg. \\
            \midrule
            100 & Diffusion Policy~\cite{chi2023diffusionpolicy} & 0.72 & 0.25 & 0.80 & 0.05 & 0.46\\
            \cmidrule{1-7}
            15 & ARP + RRTC~\cite{zhang2024arp} & 0.60 & \rebuttal{0.03} & \rebuttal{0.96} & 0.00 & \rebuttal{0.40} \\ 
            & \textbf{\ourmethod{} (Ours)} & {0.95} & {0.92} & 0.87 & 0.69 & {0.86}\\
            & └─ + \textbf{Modal Evidence} & \textbf{0.96} & \textbf{0.98} & \textbf{0.96} & \textbf{0.85} & \textbf{0.94} \\
            \bottomrule
        \end{tabular}
        \vspace{-0.3cm}
\end{table}

\begin{figure}[t]
    \centering
    \includegraphics[scale=0.5]{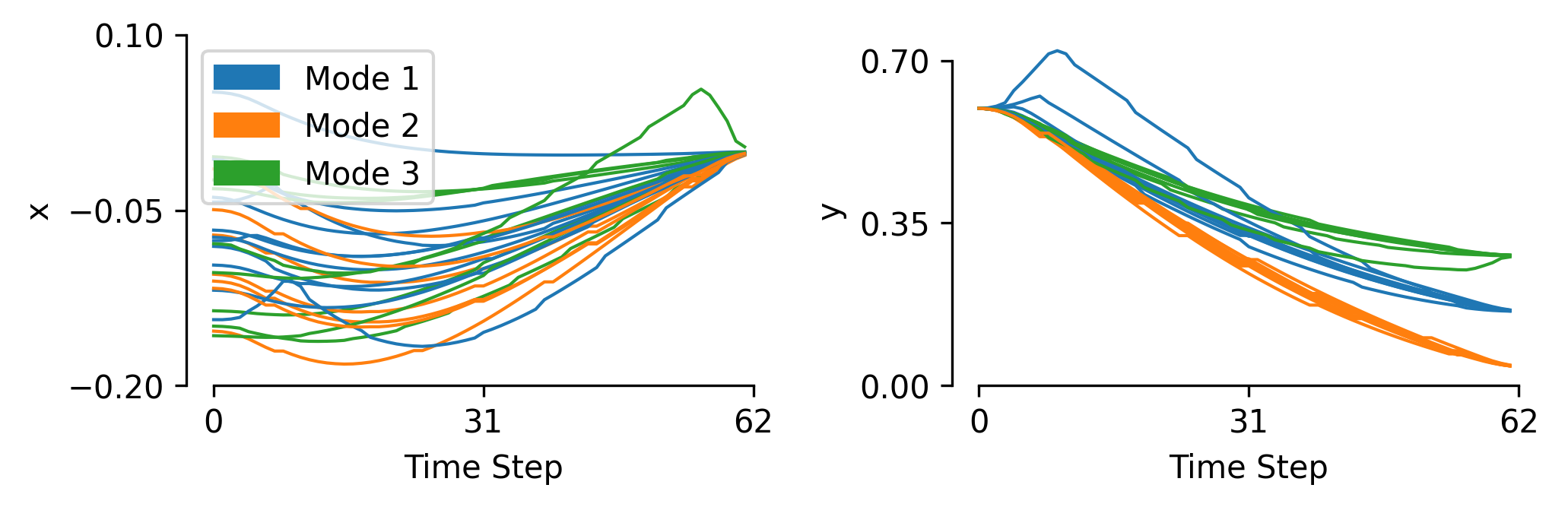}
    \caption{Modal partitioning of the first skill from the multimodal \texttt{OpenDrawer} task. %
    We plot the end-effector's \(x\) and \(y\) position over time.
    The modes are only separable in the \(y\) dimension and only in the second half of the trajectories.
    DBSCAN does not consider the variance of the data, thus failing to cluster this skill.
    In contrast, \(k\)-means successfully discovers the three plotted modes.
    }
    \label{fig:modal_part_drawer}
   \vspace{-0.5cm}
\end{figure}

\subsection{Multimodal Task-Parameterized Policy Learning}\label{sec:exp_multi_rlbench}
\para{Tasks: }
We evaluate \ourmethod{}'s capability to learn multimodal policies on the four RLBench tasks shown in \figref{fig:multi_tasks}.
All four tasks have well-defined modes.
For example, in the multimodal \texttt{PlaceCups} task, the cup may be hung on three spokes on the mug tree, giving rise to the three modes shown in \figref{fig:trans_model}. 

\para{Modal Partitioning: }
We begin by evaluating whether our modal partitioning, proposed in \secref{sec:mm_det}, can recover these modes from a set of 15 demonstrations.
Note that we again use TAPAS to segment the long-horizon tasks into a series of shorter skills and apply modal partitioning per skill.
As also shown in \figref{fig:trans_model}, the \emph{mug picking} skill in \texttt{PlaceCups} is unimodal, and only aligning the grasped mug with the mug tree and hanging it are multimodal.
Note further that we are performing task-parameterized policy learning.
The clustered data is therefore the per-frame data.
\tabref{tab:modal_part_times_multi} shows the wall clock time needed by the different partitioning methods.
Across all tasks, DBSCAN is the fastest in partitioning the modes, taking less than \SI{1}{\second} even for difficult tasks with 100 demonstrations, and less than \SI{0.1}{\second} for most tasks.
However, it fails to correctly cluster the first skill of the \texttt{OpenDrawer} task.
\figref{fig:modal_part_drawer} illustrates the reason: the multimodality only arise in one dimension and can only be separated in the second half of the trajectories.
Since DBSCAN uses a constant distance threshold and neglects the variance of the data, it thus fails to cluster this skill.
In contrast, we find \(k\)-means to be more robust, correctly clustering all multimodal skills.
It is somewhat slower than DBSCAN, taking between \SI{0.3}{\second} and \SI{24}{\second}.
Yet, its runtime is still negligible compared to training other multimodal policy learning methods, such as Diffusion Policy.
For new tasks, we therefore recommend starting with DBSCAN and switching to \(k\)-means when needed.
We do not find any benefit in fitting a full GMM over \(k\)-means.

\para{Scaling: }
We further cluster tasks with varying numbers of demonstrations to investigate the scaling properties of the clustering methods.
Note that DBSCAN has a theoretical complexity of \(\mathcal{O}(n^2)\) in the number \(n\) of demonstrations, while \(k\)-means scales \emph{linearly} with the number of demonstrations \emph{and} with the number of clusters.
\tabref{tab:modal_part_times_multi} indicates that in practice, DBSCAN and \(k\)-means both scale near linearly with the number of demonstrations.
If speed is of concern, estimating the modes from a smaller set of demonstrations and using the cluster model to label the full demonstration set might be a viable strategy.
Although the small set should also include all modes of the larger data distribution.
In contrast, the runtime of the Riemannian GMM is dominated by the needed number of maximum likelihood iterations, which is only weakly related to the number of demonstrations.
Overall, the data distribution of the task has as much of an influence on clustering runtime as the number of demonstrations.

\begin{figure}
    \centering
    \includegraphics[width=0.4\linewidth]{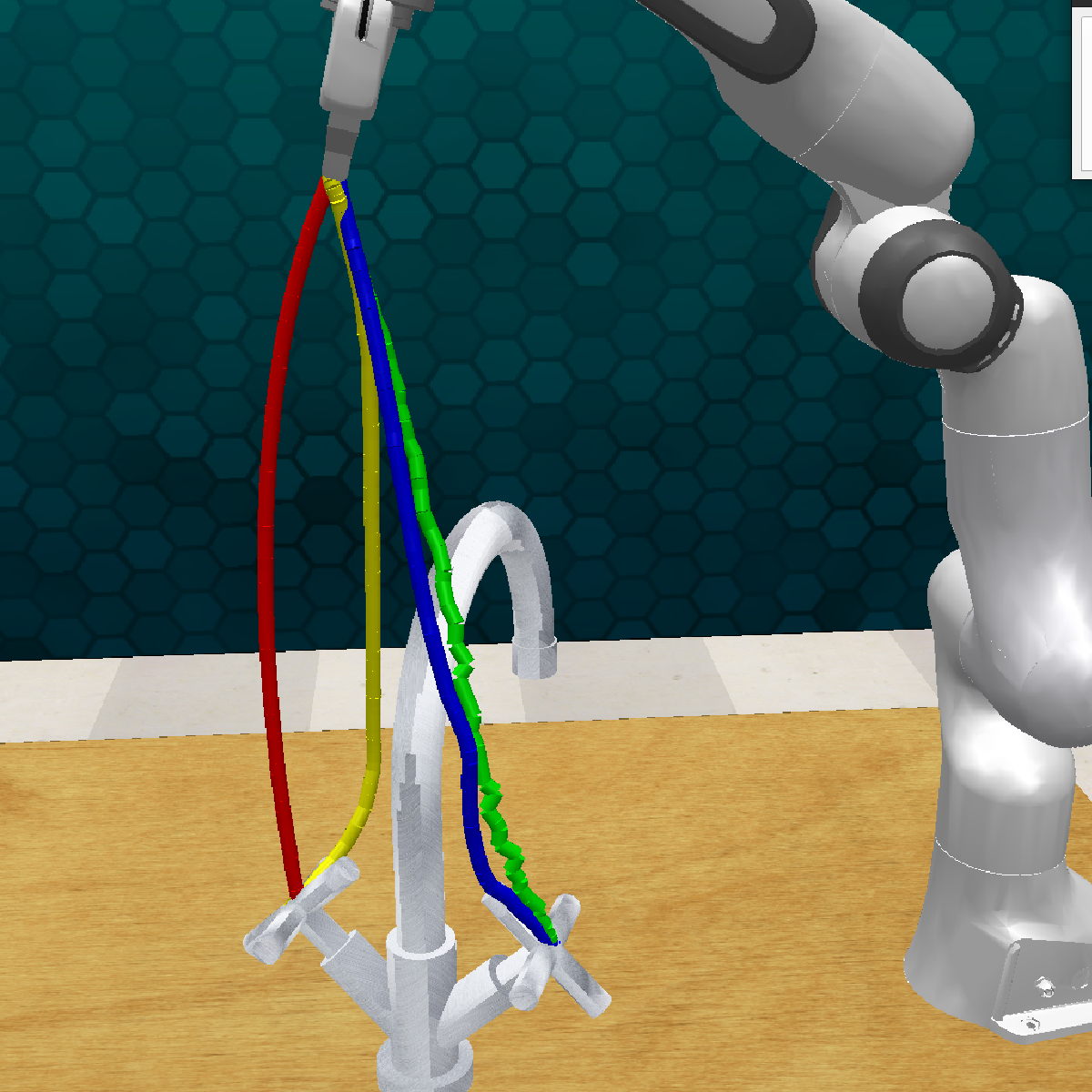}\hfil
    \includegraphics[width=0.4\linewidth]{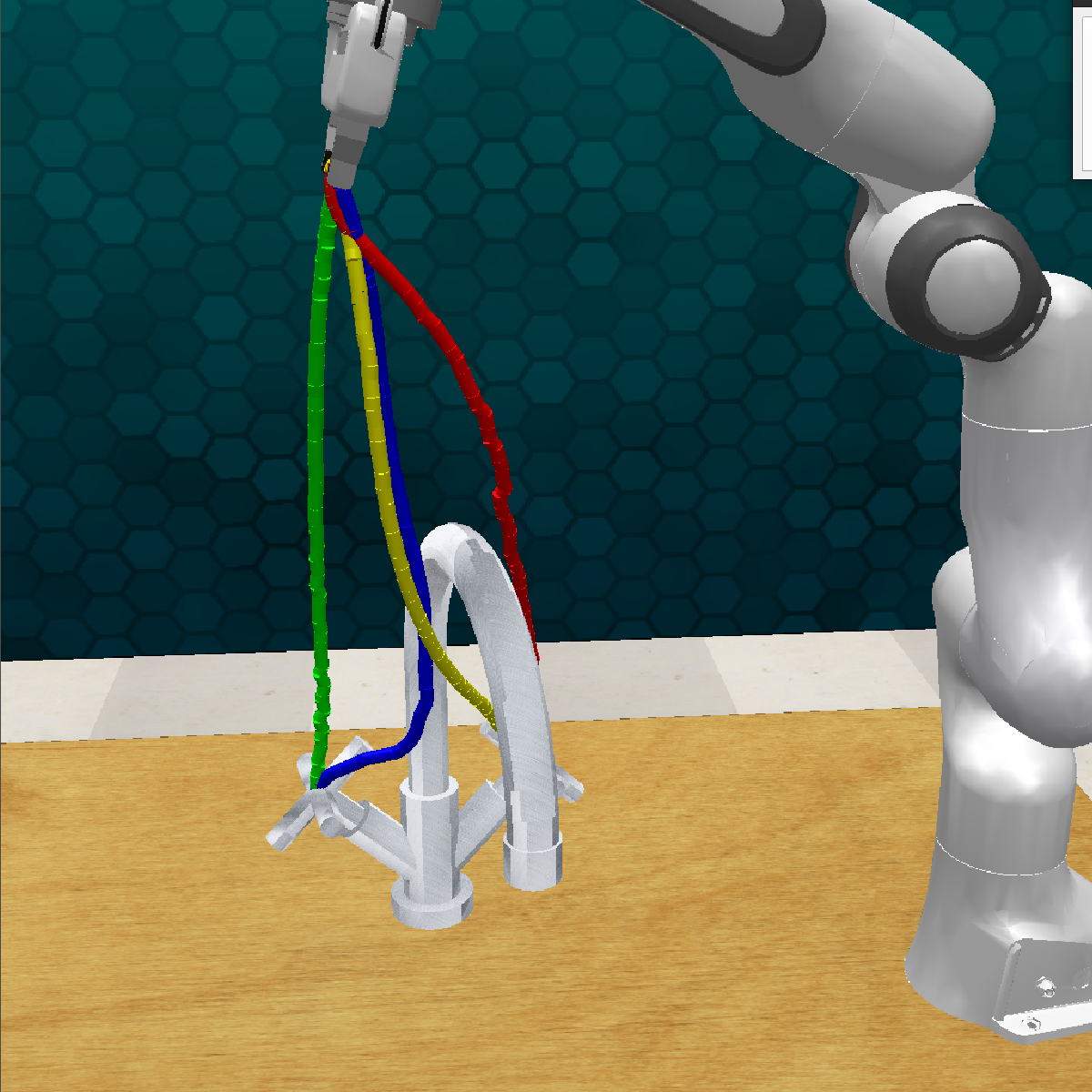}
    \caption{Two instances of the multimodal \texttt{TurnTap} task, highlighting the four different modes of approach the tap.
    Either valve might be approached in two different ways, depending on relative pose of the tap.
    In the left instance, the \textcolor{red}{red} and \textcolor{blue}{blue} modes are best suited for solving the task.
    In contrast, in the right instance, the \textcolor{green}{green} mode is better suited than the \textcolor{blue}{blue} mode.
    Constrained Gaussian Updating can easily select the suitable modes.
    }
    \label{fig:turntap_four_modes}
   \vspace{-0.3cm}
\end{figure}

\para{Mode Discovery: }
Our data-driven approach to modal partitioning has the great advantage of uncovering unexpected modes from the set of demonstrations.
For example, on \texttt{TurnTap}, we expected to find only two modes - one for each valve.
However, our modal partitioning revealed that there are actually \emph{four} modes, both for approaching the tap and for turning it.
\figref{fig:turntap_four_modes} shows that either valve is approached in two different ways depending on the relative pose of the tap.
We found that for a given task instance, the unsuitable approach generates end-effector velocities and accelerations that exceed the hardware thresholds.
Consequently, the unsuited modes are easy to detect and exclude using the modal updating introduced in \secref{sec:updating}.
As we discuss in the following paragraph, doing so improves policy success to be on par with the unimodal task variant.
The additional unexpected modes in \emph{turning} the valves resulted from the demo generator randomly turning a given valve in either direction, whereas we expected a consistent direction.
\ourmethod{} correctly identified and modeled these different modes of solving the task, which we had missed.

\para{Policy Learning:}
Afterward, we evaluate the multimodal policy learning introduced in \secref{sec:gpm} on the same four tasks.
In this experiment, we assume that in a given task instance all modes are available, and randomly sample any mode.
We test \ourmethod's capability to leverage additional evidence for mode selection in \secref{sec:exp_constraints}.
The training and evaluation protocol are the same as in \secref{sec:exp_uni_rlbench}, but we exclude baselines that are unsuited for multimodal policy learning.
\rebuttal{Having established that \ourmethod{} learns unimodal tasks with as little as five demonstrations, and expecting three modes per task, we sample 15 demonstrations for training \ourmethod{} on these multimodal tasks.
Note that we randomly sample these demonstrations and to not curate their distribution over the modes.}

\tabref{tab:success_rates_multi} shows the policy success rates.
As in the unimodal experiments, \ourmethod{} significantly outperforms Diffusion Policy, increasing policy success rates on average by 48 percentage points, while using only 15\% as much training data.
Compared to the unimodal experiments, ARP leverages the additional demonstrations to improve in \texttt{OpenDrawer} and \texttt{TurnTap}, but it still fails completely on \texttt{PlaceCups} and \texttt{WipeDesk}.
The ARP policies are still very noisy, making it difficult to judge whether they adequately model the multimodal trajectory distributions or rather overfit specific task instances.

Overall, \ourmethod{} performs about as well on the multimodal tasks, as on the unimodal variants.
Only on \texttt{WipeDesk} performance drops by 18 percentage points.
This is due to one of the modes being significantly more difficult kinematically and us sampling the modes randomly.
The second mode is not kinematically feasible for about half of the task instances.
If we use the modal evidence introduced in \secref{sec:updating} to update the modes' likelihood, policy success on \texttt{WipeDesk} improves from 69\% to 85\%.
Similarly, as discussed above, the evidence further allows us to select the mode of approaching the tap in \texttt{TurnTap} that is best suited for a given task instance, boosting policy success from 87\% to 96\%.
In both cases, this puts the policy success on par with the unimodal task variants.

\subsection{Constrained Gaussian Updating}\label{sec:exp_constraints}

\begin{table*}
    \centering
    \begin{threeparttable}
        \caption{Policy success with inference-time updating under difficult task conditions.}\label{tab:gu_collision}
        \centering
        \begin{tabular}{l cc cc cc cc c}
            \toprule
            \multirow{3}{*}{\textbf{Method}} & \multicolumn{4}{c}{\textbf{Collision Avoidance}} & \multicolumn{4}{c}{\textbf{Limited Reachability}} & \multirow{3}{*}{\textbf{Avg.}} \\
            \cmidrule(lr){2-5} \cmidrule(lr){6-9}
            & \multicolumn{2}{c}{OpenDrawer} &  \multicolumn{2}{c}{PlaceCups} &  \multicolumn{2}{c}{PlaceCups} & \multicolumn{2}{c}{TurnTap}\\
            \cmidrule(lr){2-3} \cmidrule(lr){4-5} \cmidrule(lr){6-7} \cmidrule(lr){8-9} 
            & Level 1 & Level 2 & Level 1 & Level 2 & Level 1 & Level 2 & Level 1 & Level 2 \\
            \midrule
            Diffusion Policy~\cite{chi2023diffusionpolicy} & 0.28 & 0.31 & 0.24 & 0.17 & 0.00 & 0.00 & 0.19 & 0.20 & 0.17 \\
            └─ + Inference-Time Policy Steering~\cite{wang2024inference} & 0.26 & 0.22 & 0.18 & 0.16 & 0.00 & 0.00 & 0.19 & 0.10 & 0.14 \\
            \textbf{\ourmethod{} (Ours)} & 0.66 & 0.34 & 0.63 & 0.61 & 0.40 & 0.41 & 0.75 & 0.76 & 0.57 \\
            └─ + \textbf{Constrained Gaussian Updating (Ours)} & \textbf{0.98} & \textbf{0.99} & \textbf{0.95} & \textbf{0.90} & \textbf{0.99} & \textbf{1.00} & \textbf{1.00} & \textbf{1.00} & \textbf{0.98} \\
            \bottomrule
        \end{tabular}
        \begin{tablenotes}[para,flushleft]
           \footnotesize  
           \textit{Collision Avoidance:} Additional obstacle objects are placed in the scene at inference-time, requiring collision avoidance.\\
           \textit{Limited Reachability:} The task space size is increased by placing the task objects farther away from the robot, rendering some modes unreachable.
         \end{tablenotes}
    \end{threeparttable}
       \vspace{-0.2cm}
\end{table*}

\begin{figure}
    \centering
    \includegraphics[width=0.25\linewidth]{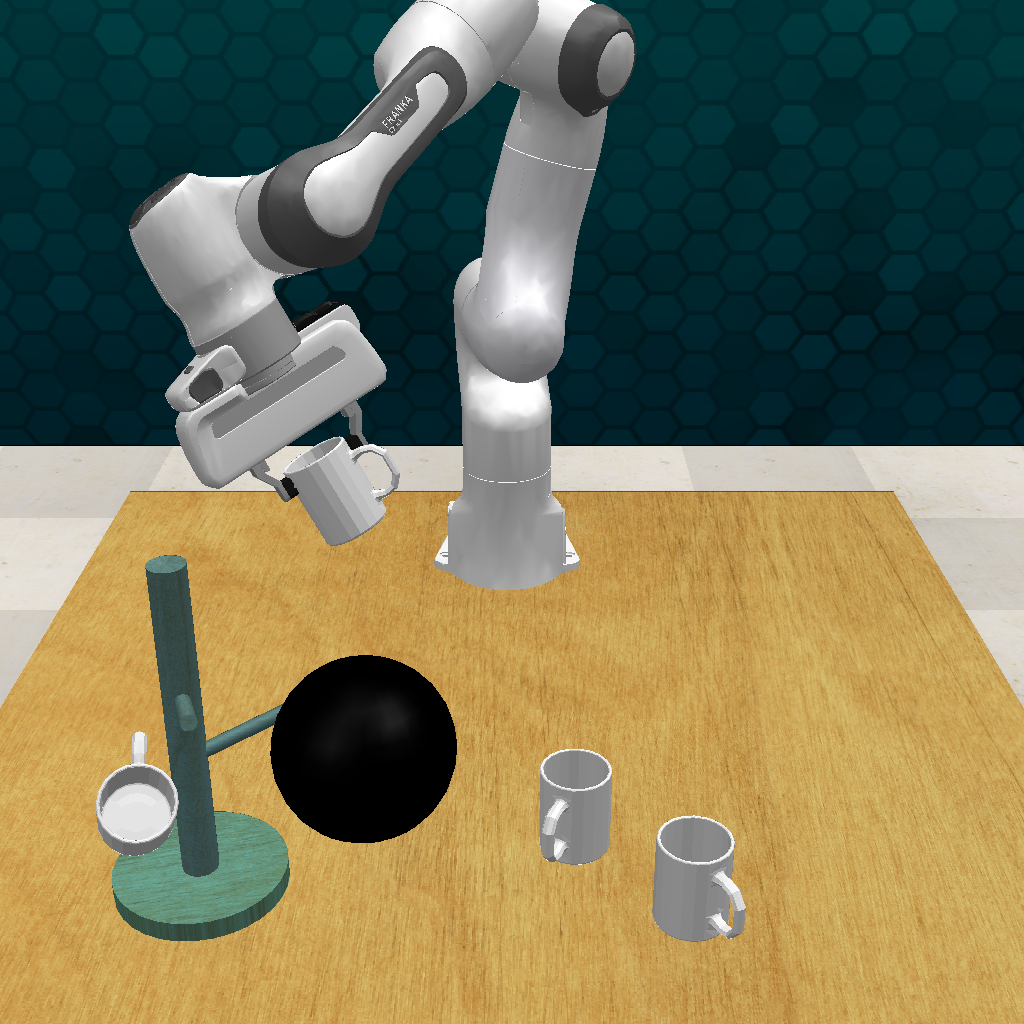}\hfil
    \includegraphics[width=0.25\linewidth]{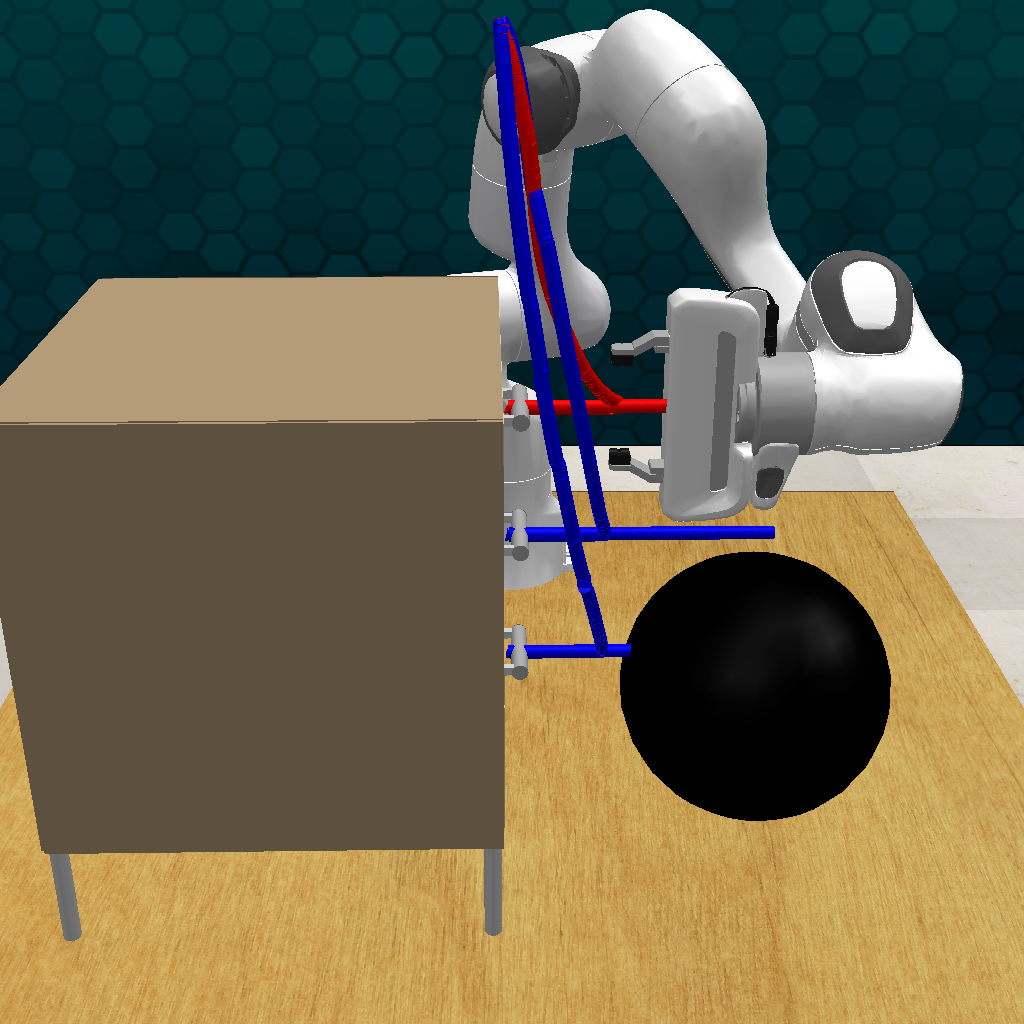}\hfil
    \includegraphics[width=0.25\linewidth]{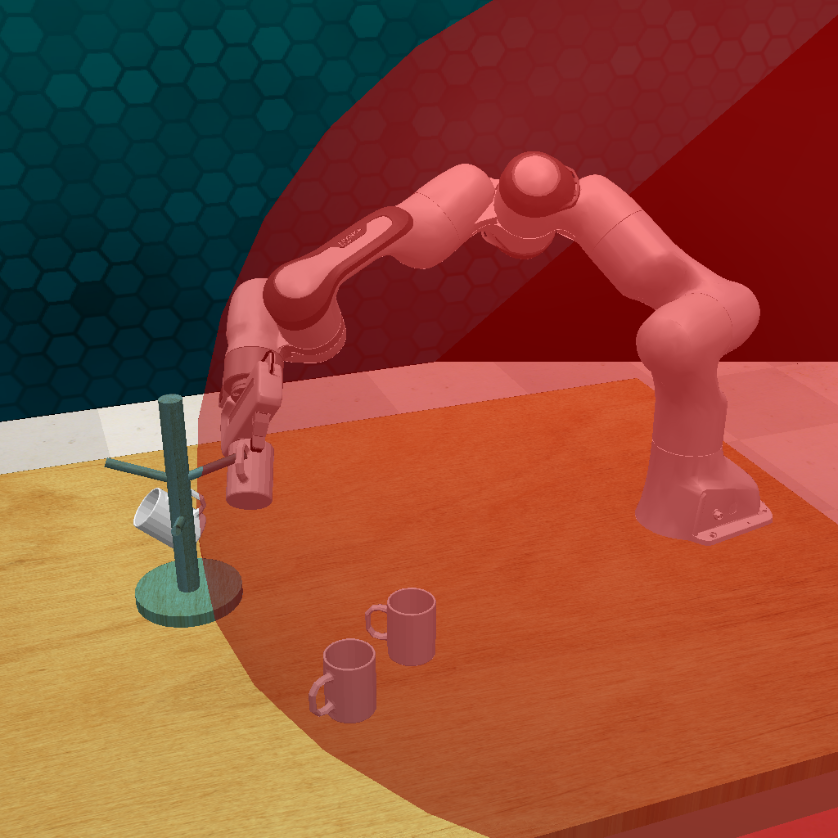}\hfil
    \includegraphics[width=0.25\linewidth]{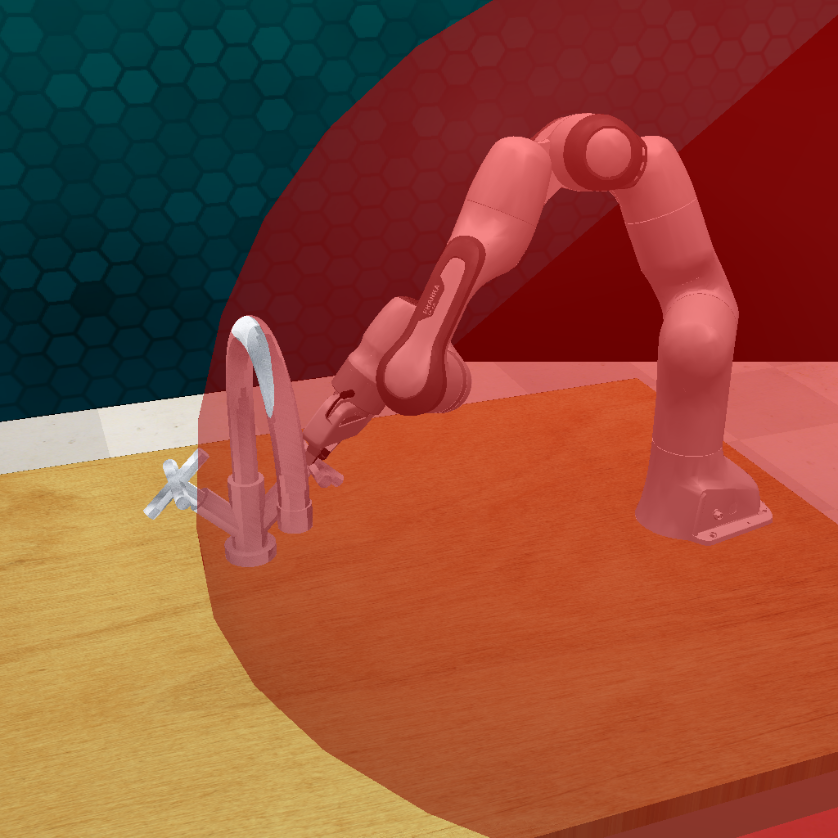}
    \caption{Task instances for inference-time updating.
    \textit{Left: } collision objects (black) are placed in \texttt{PlaceCups} and \texttt{OpenDrawer}.
    \textit{Right: } the changed object placement in \texttt{PlaceCups} and \texttt{TurnTap} makes some of the modes unreachable.
    The red spheres indicates the reach of the arm's tool center point.
    }\label{fig:collision_tasks}
   \vspace{-0.3cm}
\end{figure}

\begin{figure}
    \centering
    \includegraphics[scale=0.5]{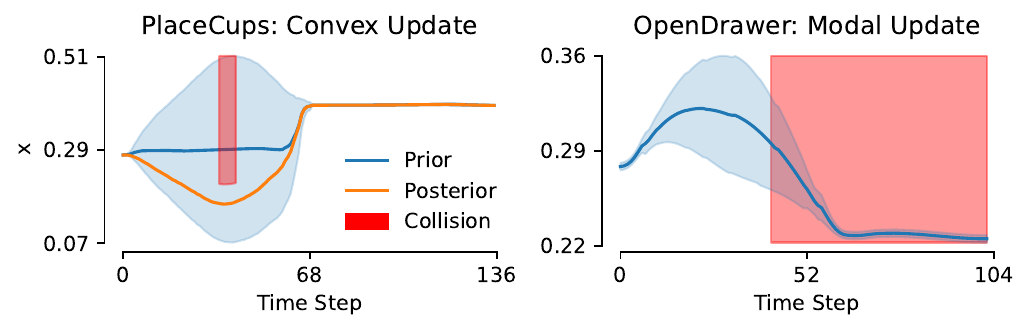}

    \caption{Gaussian updating using convex collision evidence.
    We plot the end-effector's \(x\) position over time.
    The collision region is shown in \textcolor{red}{red}.
    We compute it by taking the full end-effector position into account, but then project it onto the \(x\) dimension.
    In \texttt{PlaceCups}, the end-effector encounters the collision object in the high-variance region of the prior trajectory (\textcolor{blue}{blue}) and can evade it, leading to the \textcolor{orange}{orange} posterior.
    In contrast, in \texttt{OpenDrawer}, the collision object is encountered in the low-variance region of the trajectory - i.e.\ when pulling the handle.
    Consequently, evading the object is impossible under the prior trajectory distribution, and another mode is chosen instead.
    }
    \label{fig:collision}
   \vspace{-0.3cm}
\end{figure}

\para{Tasks: }
We evaluate the constrained Gaussian updating introduced in \secref{sec:updating} using two sets of experiments: collision avoidance and reachability.
We modify the multimodal tasks for the collision avoidance scenario by introducing additional collision objects during the inference process.
Example task instances are shown in \figref{fig:collision_tasks}.
These tasks pose two different challenges, both of which are illustrated in \figref{fig:collision}.
On the one hand, the robot might encounter an obstacle during a high-variance segment of the predicted trajectory.
In such cases, the robot can avoid the obstacle while still fulfilling the task.
In contrast, if the obstacle is encountered during a low-variance segment, deviation may jeopardize task success, and the policy should instead switch to a different mode.
For the reachability experiments, we expand the sampling space of object poses in the multimodal tasks, such that one or more modes become unreachable for the robot. 
For example, we place the tap in \texttt{TurnTap} further away from the robot, such that it can only turn one of the valves.

\begin{figure}
    \centering
    \includegraphics[trim={0 1.5cm 0 0},clip]{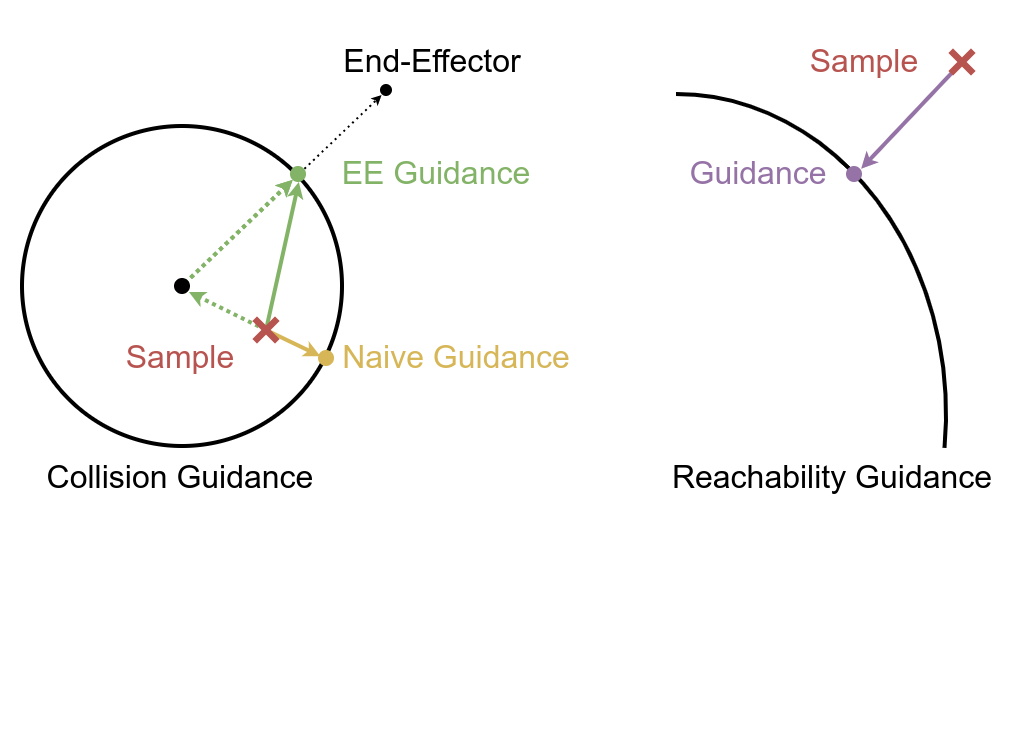}
    \caption{Guidance scheme for Diffusion models.
    \textit{Left: } the naive collision avoidance gradient (\textcolor{yellow}{yellow}) pushes the samples (\textcolor{red}{red}) towards the safe area using the surface normal, which can lead to discontinuous trajectory.
    Our EE-oriented guidance (\textcolor{green}{green}) resolves this issue by pushing all samples towards the current end-effector position.
    \textit{Right: } we ensure reachability by pushing samples towards the manipulable are using its surface normal (\textcolor{violet}{violet}).
    }
    \label{fig:itps_guidance}
   \vspace{-0.6cm}
\end{figure}

\para{Baseline:} We evaluate \ourmethod{} with and without constrained Gaussian updating against the Diffusion Policy.
To update Diffusion Policy online, we adapt Inference-Time Policy Steering (ITPS)~\cite{wang2024inference}.
ITPS uses an annealed Markov-Chain Monte-Carlo (MCMC) guidance scheme to steer the policy with additional gradients.
While the original ITPS uses a user input to steer a multimodal diffusion model \emph{towards} a specific mode, we leverage guidance to steer the policy away from a collision object or a workspace boundary.
For obstacle avoidance, we compute the guidance gradient as follows.
Given a trajectory point \(\boldsymbol x_i\), a collision sphere with center \(\boldsymbol c_i\) and radius \(r_\text{coll}\), and a safety distance \(d_{\text{safe}}\), the naive guidance gradient is the sphere's unit normal pointing towards \(\boldsymbol x_i\), scaled using the inverse distance \(\lVert \boldsymbol x_i - \boldsymbol c\rVert\), i.e.\
\begin{equation}
    \boldsymbol g_i^c = \frac{r_\text{coll} + d_{\text{safe}} - \lVert \boldsymbol x_i - \boldsymbol c\rVert}{d_\text{safe}} \frac{\boldsymbol x_i -\boldsymbol c}{\lVert \boldsymbol x_i - \boldsymbol c\rVert}.
\end{equation}
However, this naive gradient can push consecutive noisy diffusion samples towards different sides of the sphere, leading to discontinuous trajectories.
Instead, we use the current end-effector position \(\boldsymbol e\) as reference and push the diffusion sample to the outside of the sphere but towards the end-effector, i.e.\
\begin{equation}
    \boldsymbol g_i^e = \boldsymbol c - \boldsymbol x_i + \frac{\boldsymbol e - \boldsymbol c}{\lVert \boldsymbol e - \boldsymbol c\rVert} (r +d_\text{safe}).
\end{equation}
When multiple obstacles are in the scene, we sum their induced gradients.
To ensure reachability, we similarly apply a gradient pointing towards the center of the reachable sphere \((\boldsymbol a, r_\text{reach})\), if the point \(\boldsymbol x_i\) is outside of it.
I.e.
\begin{equation}
    \boldsymbol g_i^r = \begin{cases}
        \frac{\lVert\boldsymbol{x}_i - \boldsymbol{a}\rVert - r_\text{reach}}{r_\text{reach}}\frac{\boldsymbol a-\boldsymbol x_i}{\lVert\boldsymbol{x}_i - \boldsymbol{a}\rVert} &,\text{ if } \lVert \boldsymbol{x}_i - \boldsymbol{a}\rVert >r_\text{reach}\\
        \boldsymbol{0} &,\text{ otherwise}.
    \end{cases}
\end{equation}
\figref{fig:itps_guidance} illustrates these gradients.
We then follow prior work in applying the annealed MCMC guidance scheme~\cite{wang2024inference}.

\para{Results:} From the results in \tabref{tab:gu_collision}, we make the following notable observations.
First, Diffusion Policy (without ITPS) performs as expected on the collision experiments but fails entirely in the reachability scenario on \texttt{PlaceCups}.
Notably, it fails to place the cup and even grasp it.
Given that the reach experiments include out-of-distribution observations with the mug tree being placed further away from the robot, this indicates limited extrapolation capabilities of the Diffusion Policy.
In contrast, \ourmethod{} performs robustly in both scenarios. Without Gaussian updating, it fails only when it samples an infeasible mode.
With constrained Gaussian updating, the reachability task is nearly perfectly solved by adjusting the prior likelihoods to focus on feasible modes.

\begin{table*}
    \begin{threeparttable}
        \caption{Policy success rates for embodiment transfer from the Franka Emika arm to the UR5 on unimodal RLBench tasks. }\label{tab:success_rates_traj_opt_uni}
        \centering
        \setlength{\tabcolsep}{5pt}
        \begin{tabular}{l cccccc cccccc}
            \toprule
            \multirow{2}{*}{\textbf{Method}} & \multicolumn{6}{c}{\textbf{Mildly Constrained Tasks}} & \multicolumn{6}{c}{\textbf{Highly Constrained Task}} \\
            \cmidrule(lr){2-7} \cmidrule(lr){8-13}
            & \makecell{Open \\ Drawer} & \makecell{Stack \\ Wine} & \makecell{SlideBlock \\ ToTarget} & \makecell{SweepTo \\ Dustpan} & \makecell{Place \\ Cups} & Avg. & \makecell{Open \\ Microwave} & \makecell{Toilet \\ SeatUp} & \makecell{Wipe \\ Desk} & \makecell{Turn \\ Tap} & \makecell{ScoopWith \\ Spatula} & Avg. \\
            \midrule
            Diffusion Policy~\cite{chi2023diffusionpolicy} & 0.00 & 0.00 & 0.00 & 0.00 & 0.00 & 0.00 & 0.00 & 0.00 & 0.00 & 0.00 & 0.00 & 0.00 \\
            ARP + RRTC~\cite{zhang2024arp} & 0.00 & 0.01 & 0.00 & 0.00 & 0.00 & 0.00 & 0.00 & 0.00 & 0.00 & 0.00 & 0.00 & 0.00  \\
            TAPAS-GMM~\cite{vonhartz2024art} & 0.00 & 0.00 & 0.00 & 0.19 & 0.33 & 0.10 & 0.00 & 0.01 & 0.01 & 0.03 & 0.00 & 0.01\\
            \ourmethod{} Naive & 0.00 & 0.06 & 0.04 & 0.54 & 0.71 & 0.27 & 0.00 & 0.40 & \textbf{0.95} & 0.25 & 0.01 & 0.32 \\
            \ourmethod{} End-Effector & 0.06 & 0.06 & 0.05 & \textbf{0.56} & 0.72 & 0.29 & 0.01 & 0.55 & 0.92 & 0.25 & 0.02 & 0.35 \\
            \textbf{\ourmethod{} + \vapor{} (Ours)} & \textbf{0.40} & 0.65
            & \textbf{0.68} & 0.51 & \textbf{0.90} & \textbf{0.63} & \textbf{0.05} & \textbf{0.84} & 0.94 & \textbf{0.71} & \textbf{0.82} & \textbf{0.68} \\
            \midrule
            \ourmethod{} Trained on UR5 & - & \textbf{1.00} & - & 0.04 & - & 0.21 & - & - & 0.00 & 0.13 & - & 0.03 \\
            └─ + \vapor{}  & - & \textbf{1.00} & - & 0.19 & - & 0.24 & - & - & 0.94 & 0.29 & - & 0.23\\
            \bottomrule
        \end{tabular}
        \begin{tablenotes}[para,flushleft]
           \footnotesize
           A dash (–) indicates that RLBench’s RRTC planner failed to collect demonstrations with the UR5 and is counted as zero success in the average.
       \end{tablenotes}
    \end{threeparttable}
\end{table*}
\begin{table}
        \caption{Policy success rates for embodiment transfer from the Franka Emika arm to the UR5 on multimodal RLBench tasks. }\label{tab:success_rates_traj_opt_mult}
        \centering
        \setlength{\tabcolsep}{5pt}
        \begin{tabular}{l ccccc}
            \toprule
            \multirow{2}{*}{\textbf{Method}} & \multicolumn{4}{c}{\textbf{Multimodal Tasks}} \\
            \cmidrule(lr){2-6}
            & \makecell{Open \\ Drawer} & \makecell{Place \\ Cups} & \makecell{Turn \\ Tap} & \makecell{Wipe \\ Desk} & Avg. \\
            \midrule
            Diffusion Policy~\cite{chi2023diffusionpolicy} & 0.00 & 0.00 & 0.00 & 0.01 & 0.00\\
            ARP + RRTC~\cite{zhang2024arp} & 0.00 & 0.00 & 0.00 & 0.00 & 0.00\\
            \ourmethod{} Naive & 0.00 & 0.35 & 0.54 & 0.42 & 0.33\\
            \ourmethod{} Mode Selection & 0.00 & 0.38 & 0.43 & 0.68 & 0.37 \\
            \textbf{\ourmethod{} + \vapor{} (Ours)} & \textbf{0.20} & \textbf{0.92} & \textbf{0.90} & \textbf{0.91} & \textbf{0.73} \\
            \midrule
            \ourmethod{} Trained on UR5 & - & - & - & 0.00 & 0.00  \\
            └─ + \vapor{} & - & - & - & 0.77 & 0.19  \\
            \bottomrule
        \end{tabular}
        \vspace{-.2cm}
\end{table}
\begin{figure}
    \centering
    \includegraphics[width=0.4\linewidth]{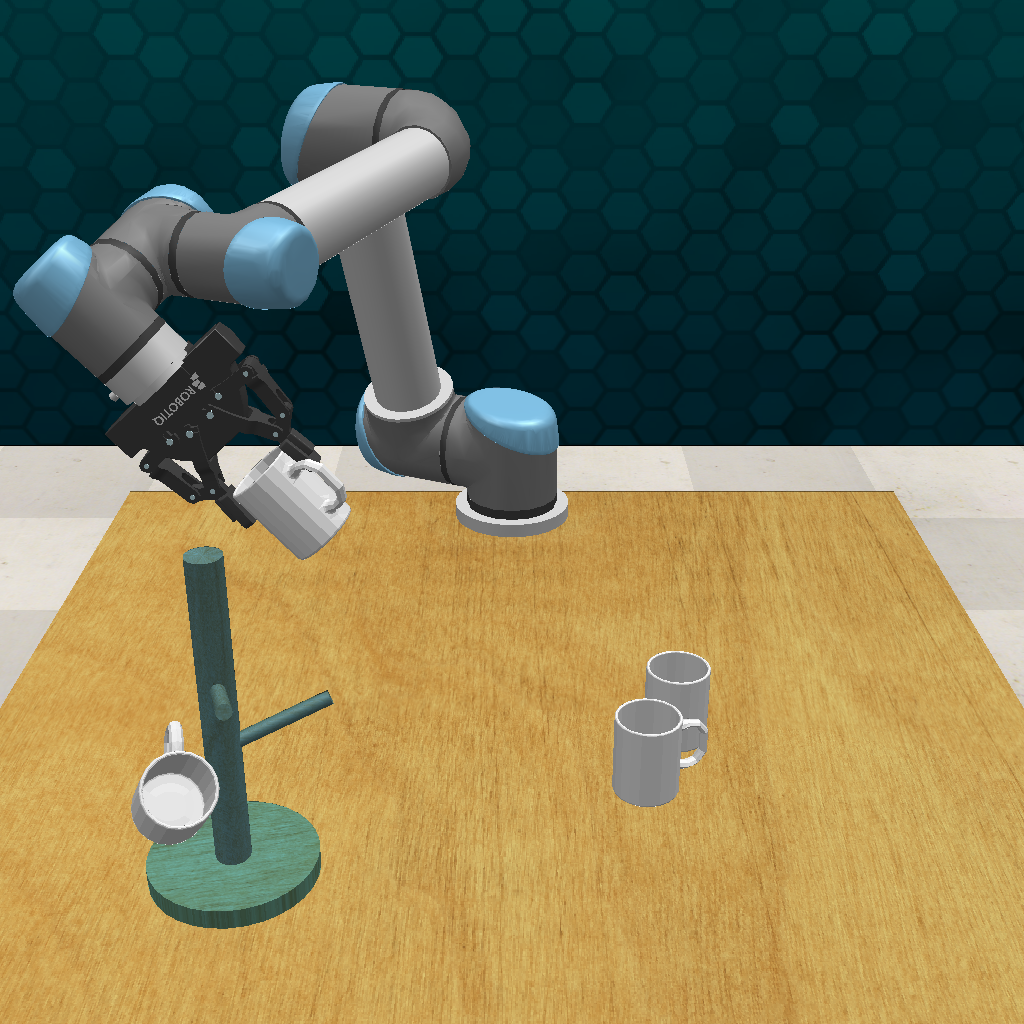}\hfil
    \includegraphics[width=0.4\linewidth]{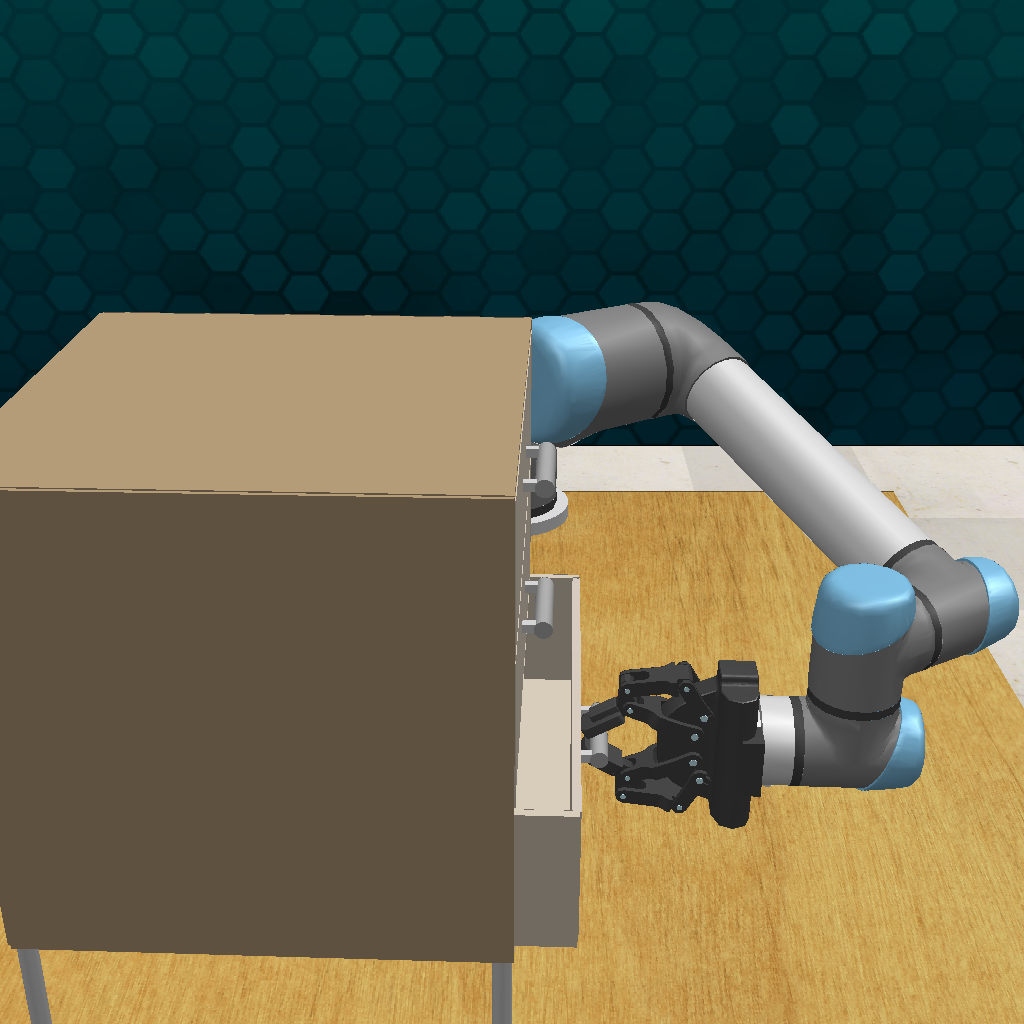}
    
    \caption{\textit{Left: } We evaluate policy cross-embodiment transfer from the Franka Emika arm to the UR5 with a Robotiq gripper on all the introduced tasks.
    \textit{Right: } In contrast to the Franka Emika arm, the UR5 has only six degrees of freedom, hence struggling to solve some of the tasks, like \texttt{OpenDrawer}.
    }\label{fig:ur5}
   \vspace{-0.6cm}
\end{figure}

Second, ITPS fails to improve the Diffusion policy's performance in all experimental conditions.
One likely reason is the receding horizon of only 16 actions, which prevents the policy from foreseeing obstacles further along the trajectory. 
More fundamentally, Diffusion Policy is highly sensitive to out-of-distribution samples, making it challenging to guide effectively at inference time.
\footnote{We verify that this is indeed a problem of OOD samples and not specific to the curved diffusion paths, by repeating the experiments with a Flow Matching (FM) Policy~\cite{chisari2024flowmatch}.
While FM is nudged into a different mode more easily due to its linear flows, it still fails the task about as frequently as Diffusion Policy, due to being OOD.}
While ITPS enjoys some success with mode \emph{attraction}~\cite{wang2024inference}, \emph{repulsion} does not seem to work well.

Third, \ourmethod{} successfully uses constrained Gaussian updating to incorporate collision information.
In cases where a mode becomes fully obstructed, as in \texttt{OpenDrawer}, its likelihood is reduced to zero, and other feasible modes are selected.
When only part of a mode’s trajectory distribution intersects with an obstacle, the Gaussians are updated accordingly to avoid the obstacle while maintaining task success (as in \texttt{PlaceCups}).
Both scenarios are shown in \figref{fig:collision}.
Recall that we achieve both these capabilities using the \emph{same} process of constrained Gaussian updating - introduced in \secref{sec:updating}.
The difference in behavior stems from the variance encoded by our probabilistic policy.
For example, in \texttt{OpenDrawer}, once the handle is grasped, little variance in the gripper position is permissible, lest the gripper slips off the handle.
Consequently, the movement needed to evade the obstacle is too unlikely under the policy, leading to the mode being discarded.
In contrast, when more variance is permitted under a policy's mode, it is adequately deformed to evade the obstacle.
For instance, this deformation can be observed in \texttt{PlaceCups}.

Fourth, \ourmethod{} accounts for long-horizon dependencies by propagating collision and reachability constraints along the full trajectory.
For example, if a collision would occur in the final phase of a multi-skill task, MiDiGaP can already avoid selecting modes in the earlier skills that \emph{lead} to the critical mode.
Such long-range dependencies are difficult to model using purely reactive policies.

Finally, our constrained Gaussian updating is computationally efficient.
While ITPS increases the already high computational cost of Diffusion Policy four-fold, the Gaussian updating adds only modest overhead.
The precise overhead depends on the number of samples used for moment matching.
For 1,000 samples, the wallclock time overhead is about 3\%; even at 10,000 samples, it is below 50\%. 
We find that one thousand samples are sufficient for strong performance, and report the full inference times in \tabref{tab:inf_times}.

\subsection{Trajectory Optimization and Cross-Embodiment Transfer}\label{sec:exp_traj_opt}

\begin{figure*}[tb]
    \begin{tikzpicture}
        \def\imgwidth{0.2\textwidth}
        \def\imgheight{0.15\textwidth}
        \def\borderoffset{1pt}
        \def\borderwidth{2pt}
         \node[anchor=north west,inner sep=0pt] (firstrow) {
         \includegraphics[width=0.2\textwidth]{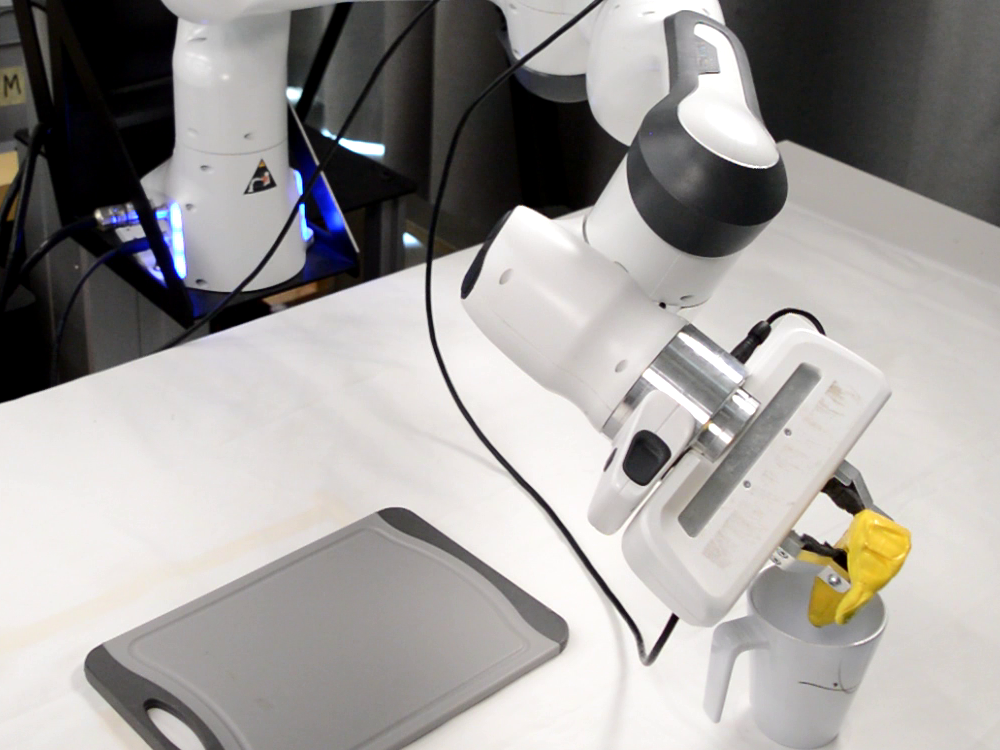}%
        \includegraphics[width=0.2\textwidth]{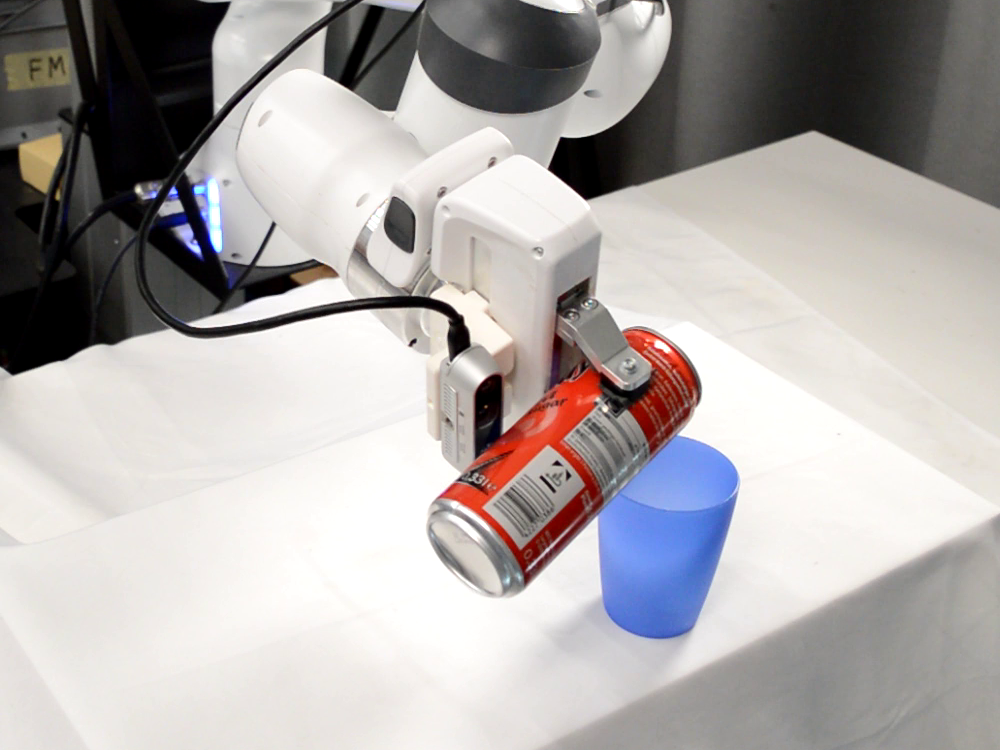}%
        \includegraphics[width=0.2\textwidth]{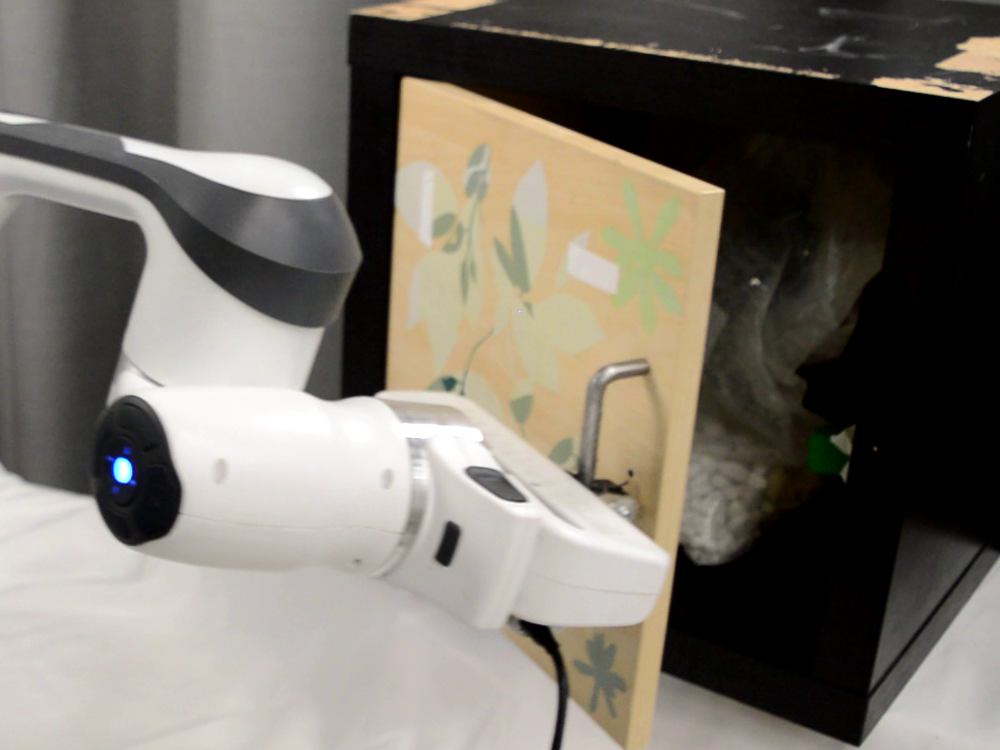}%
        \includegraphics[width=0.2\textwidth]{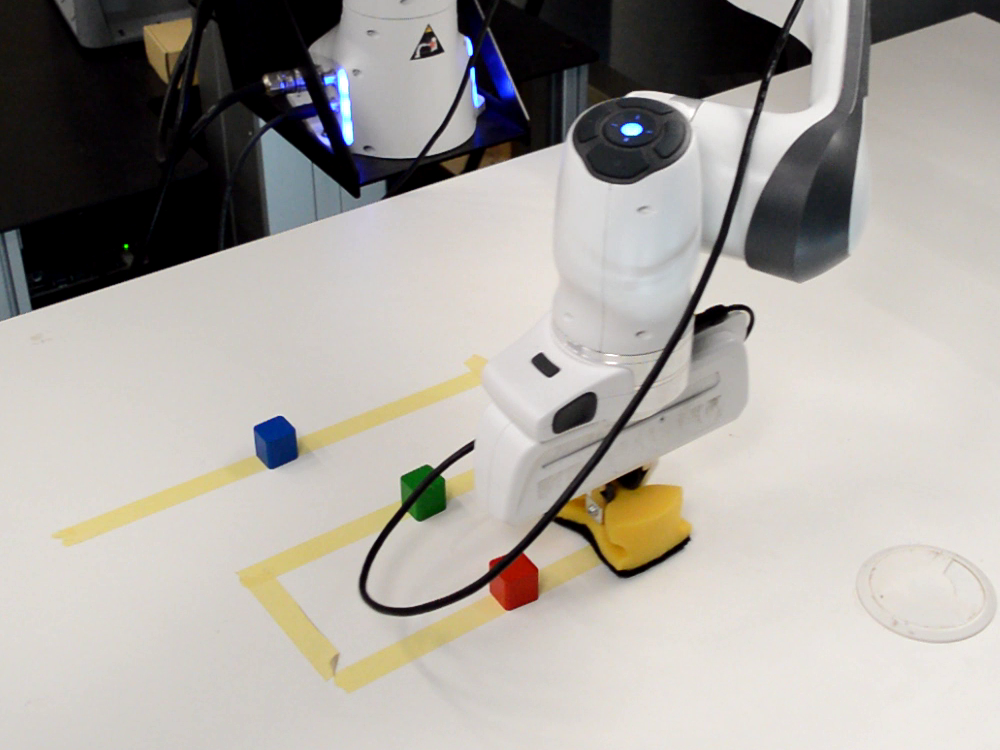}%
        \includegraphics[width=0.2\textwidth]{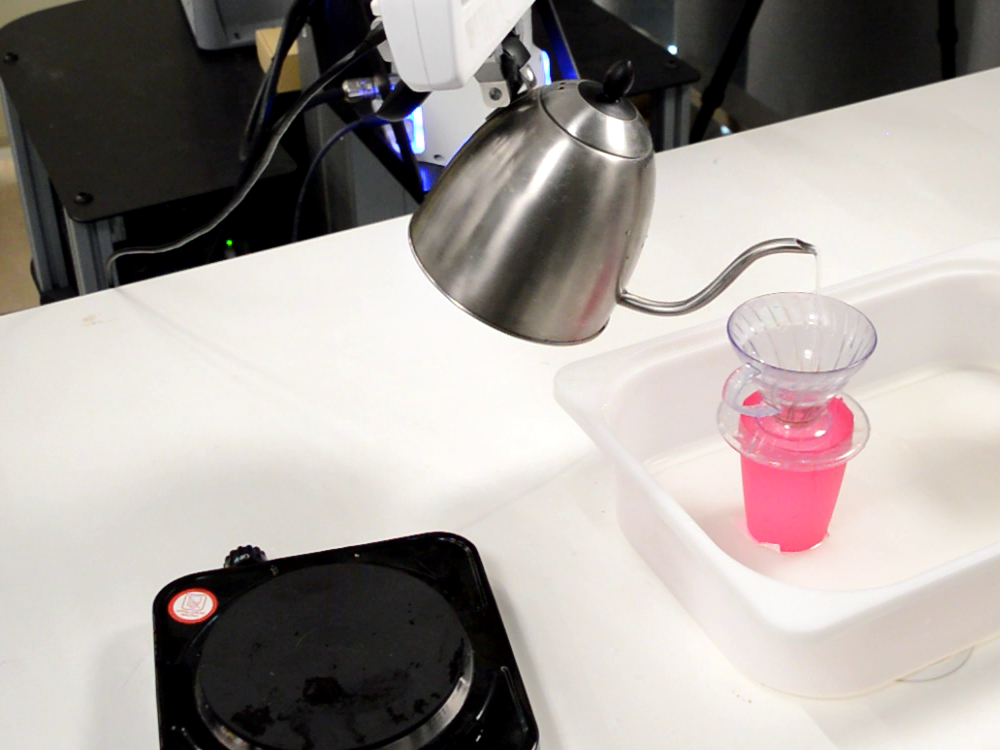}
    };
	\node[anchor=north west,inner sep=0pt, below=0pt of firstrow] (secondrow) {
        \includegraphics[width=0.2\textwidth]{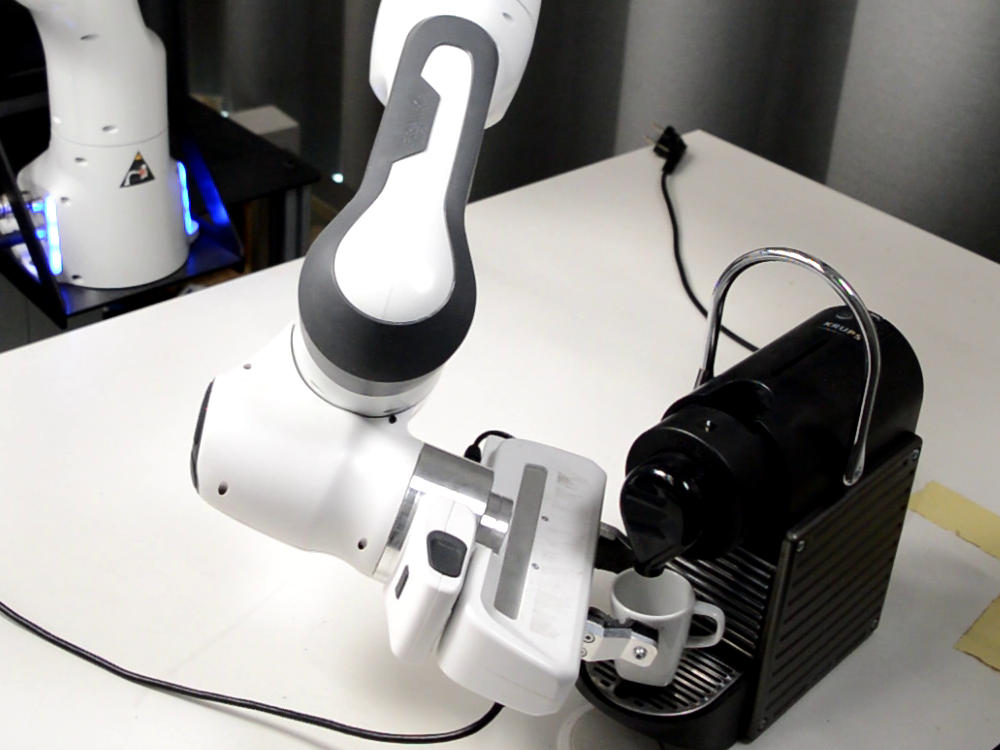}%
        \includegraphics[width=0.2\textwidth]{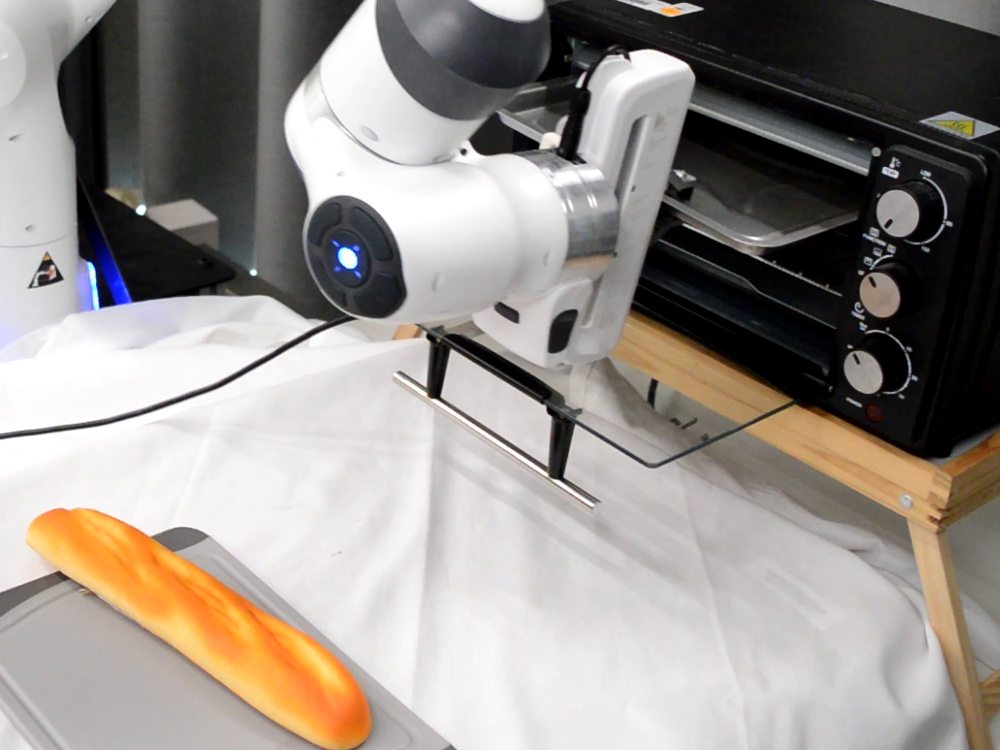}%
        \includegraphics[width=0.2\textwidth]{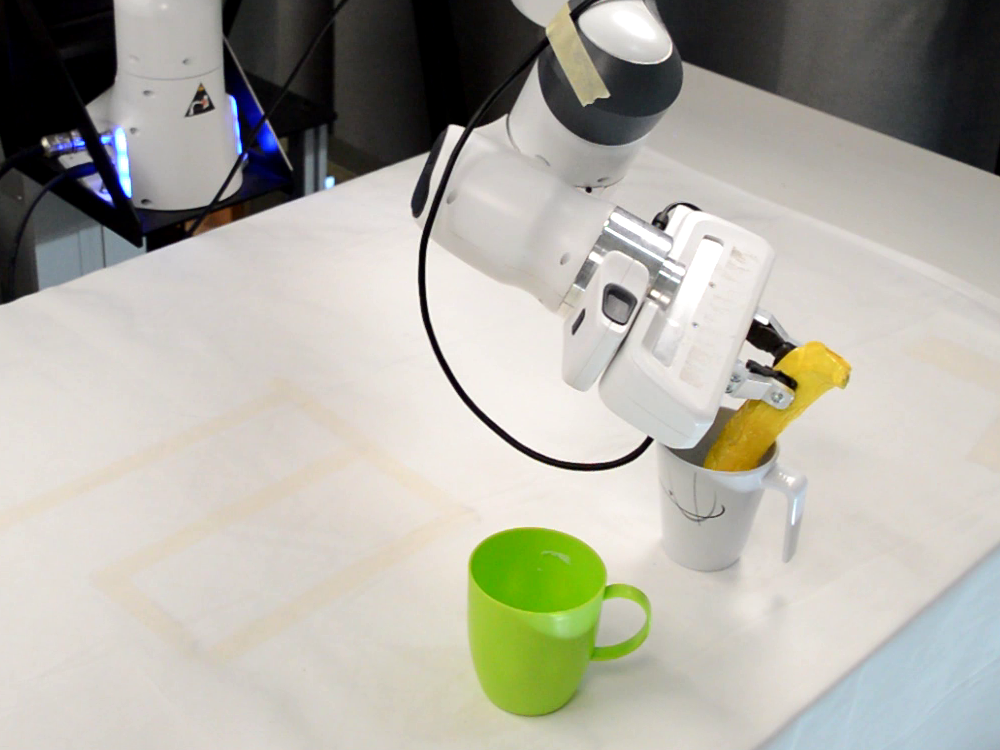}%
        \includegraphics[width=0.2\textwidth]{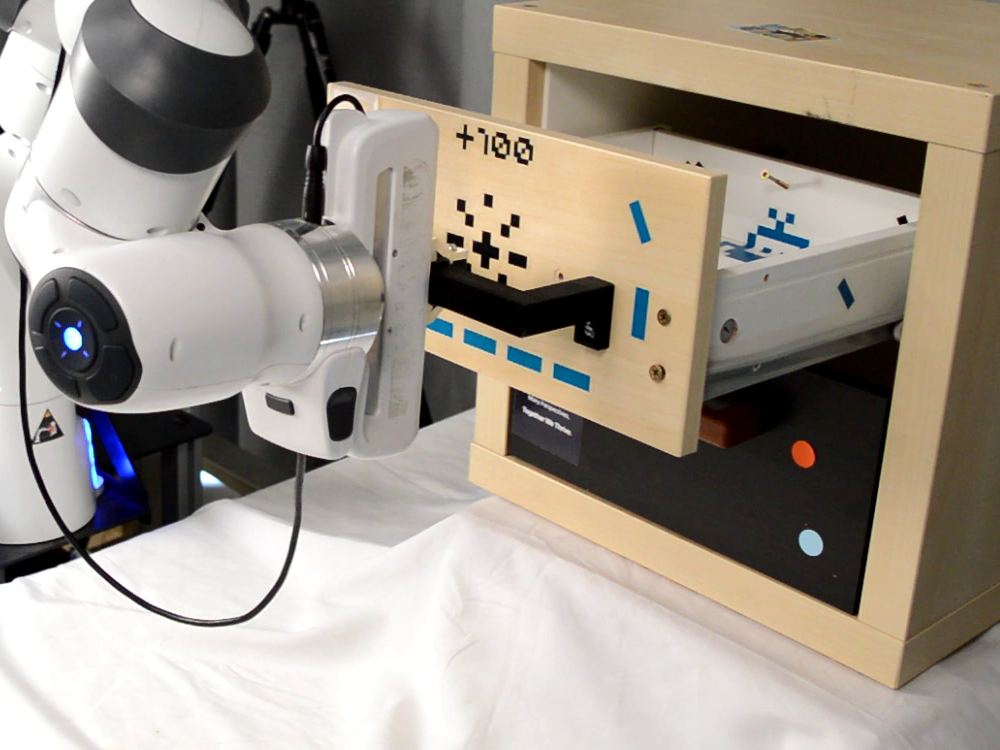}%
        \includegraphics[width=0.2\textwidth]{figures/tasks_real/wipeDesk.png}
    };
          \draw[RoyalBlue, line width=2pt] 
            (0, 0) -- (2*\imgwidth-\borderoffset, 0)  -- (2*\imgwidth-\borderoffset, -2*\imgheight-\borderoffset) -- (0, -2*\imgheight-\borderoffset) -- (0,0);
        \draw[CrimsonRed, line width=2pt] 
            (2*\imgwidth+\borderoffset, 0) -- (5*\imgwidth, 0) -- (5*\imgwidth, -\imgheight+\borderoffset) -- (2*\imgwidth+\borderoffset, -\imgheight+\borderoffset) -- (2*\imgwidth+\borderoffset, 0);
        \draw[ForestGreen, line width=2pt] 
            (2*\imgwidth+\borderoffset, -\imgheight-\borderoffset) -- (5*\imgwidth, -\imgheight-\borderoffset) -- (5*\imgwidth, -2*\imgheight-\borderoffset) -- (2*\imgwidth+\borderoffset, -2*\imgheight-\borderoffset) -- (2*\imgwidth+\borderoffset, -\imgheight-\borderoffset);
    \end{tikzpicture}    
    \caption{Real-world \rebuttal{policy} learning tasks.
    \textcolor{RoyalBlue}{\textit{Blue}}: The \textit{mildly constrained} tasks  \texttt{PickAndPlace}, \texttt{PourDrink}, \texttt{MakeCoffee}, and \texttt{BakeBread}.
    \textcolor{CrimsonRed}{\textit{Red}}: The  \emph{highly constrained} tasks include articulated object interactions (\texttt{OpenCabinet}), complex trajectory shapes (\texttt{WipeDesk}, \texttt{PourSpiral}), or velocity constraints (\texttt{ScoopWithSpatula}).
    \textcolor{ForestGreen}{\textit{Green}}: The  \emph{multimodal} tasks \texttt{PickAndPlace}, \texttt{OpenDrawer}, and \texttt{WipeDesk}.  
    }\label{fig:real_tasks}
   \vspace{-0.3cm}
\end{figure*}

\begin{figure}
    \centering
    \includegraphics[width=0.4\linewidth]{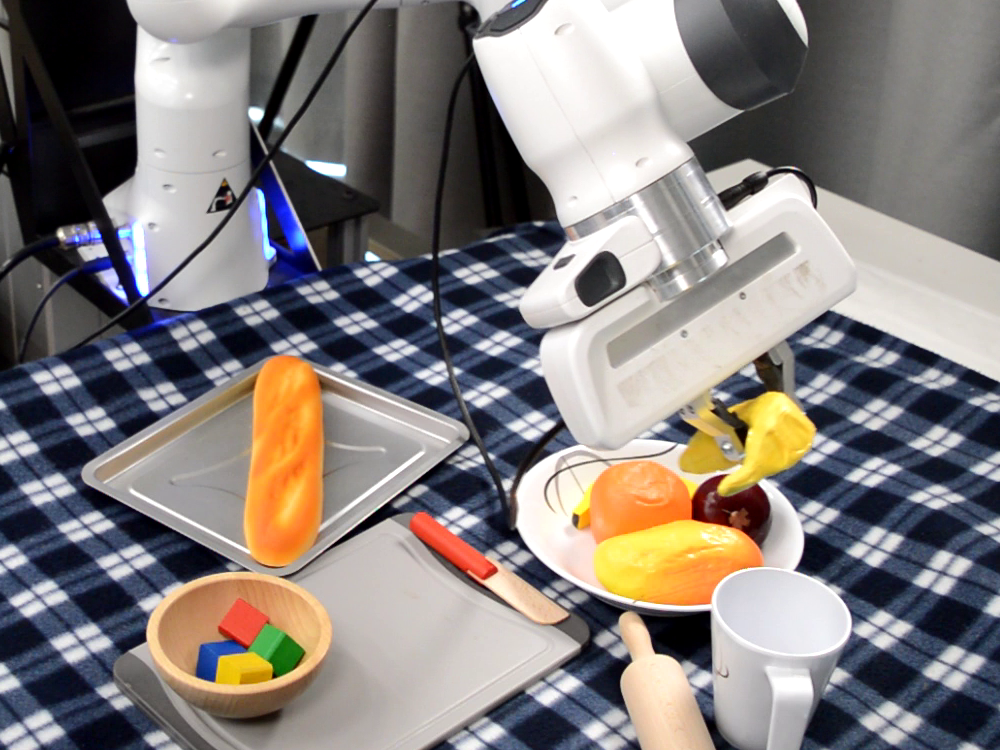}%
    \includegraphics[width=0.4\linewidth]{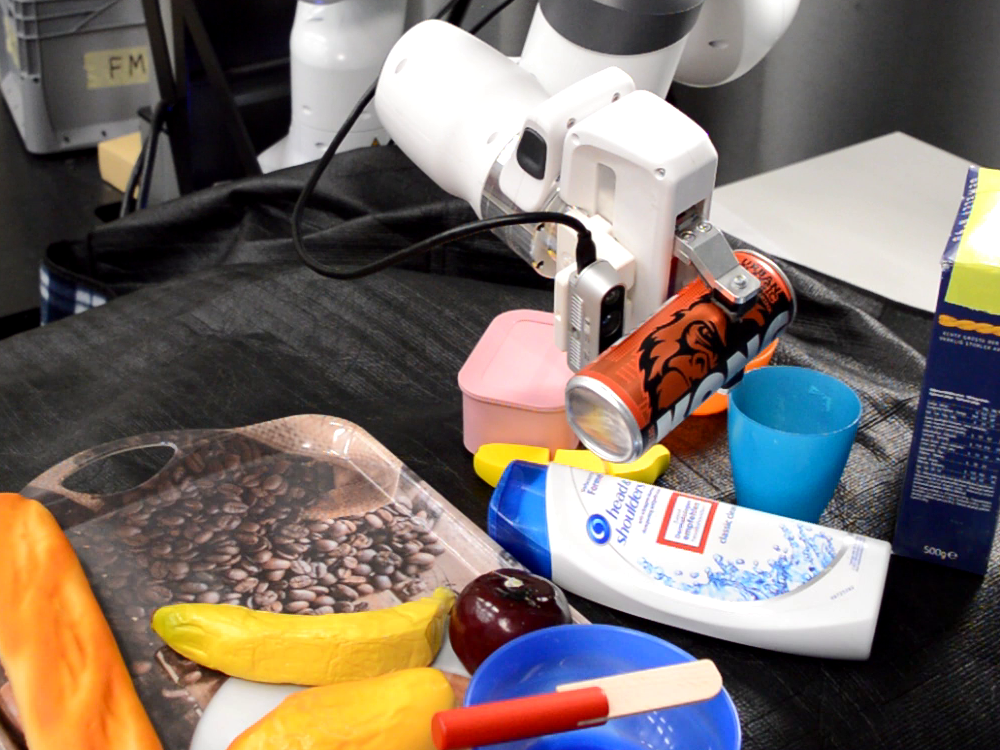}
    \includegraphics[width=0.4\linewidth]{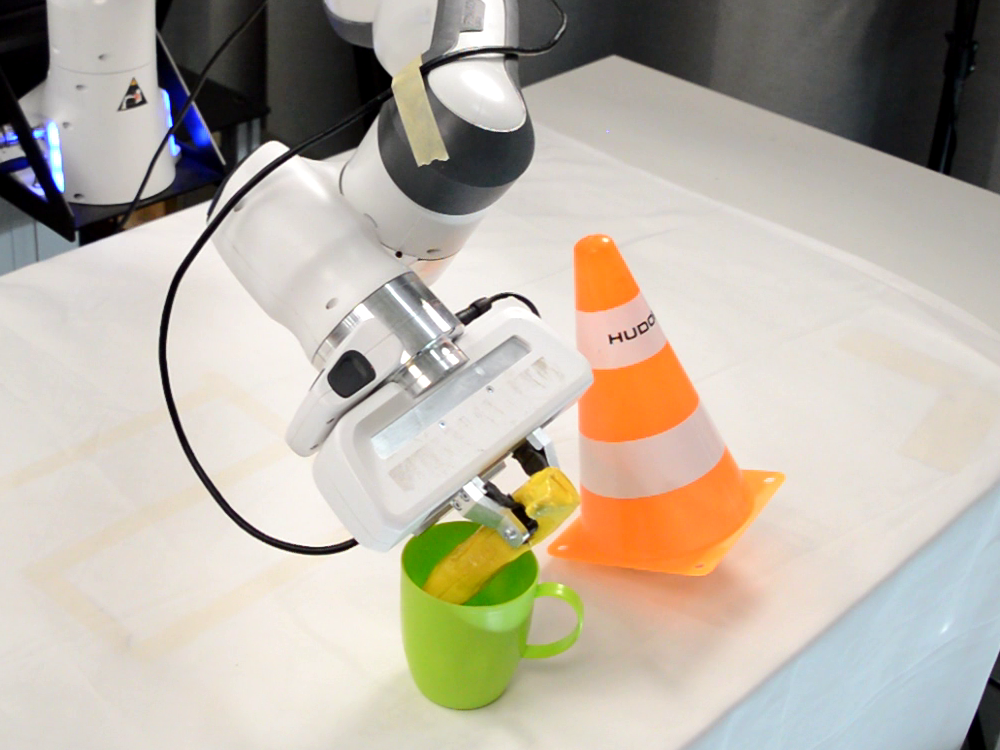}%
    \includegraphics[width=0.4\linewidth]{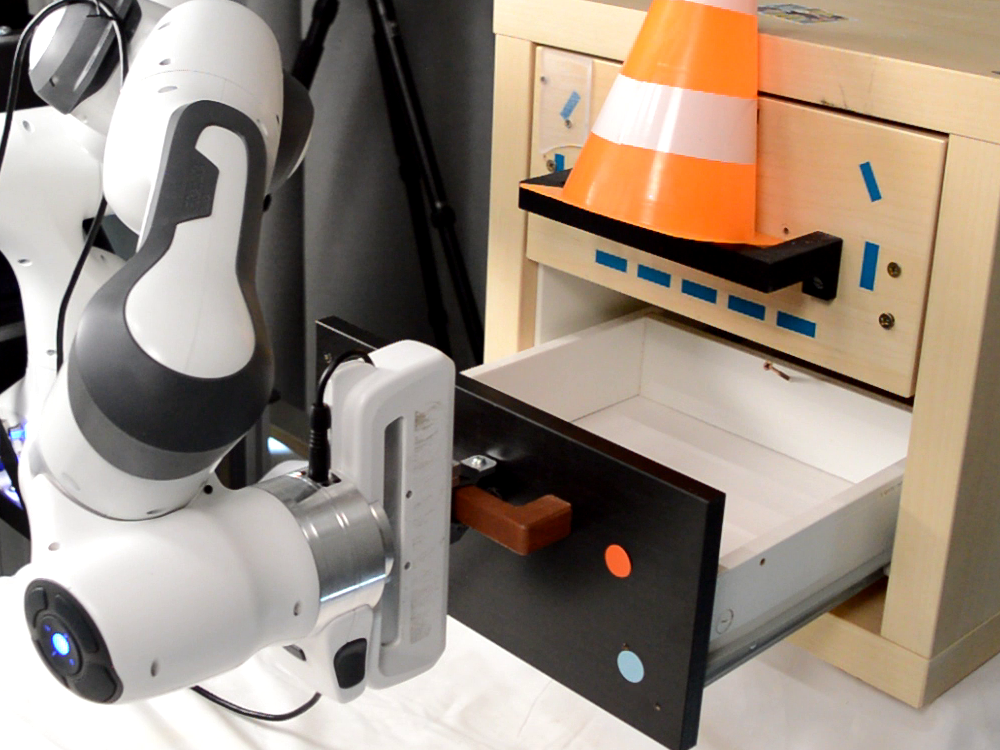}
    \caption{Example scenes from the visual generalization study (\textit{top}) and the collision avoidance experiments (\textit{bottom}).}
    \label{fig:viz_gen}
    \vspace{-.5cm}
\end{figure}

\para{Tasks: }
In \secref{sec:traj_opt} we hypothesized that \ourmethod{} enhances policy generalization and facilitates embodiment transfer through \vaporfull{} (\vapor{}).
We validate this hypothesis via an embodiment transfer experiment using the policies trained in \secref{sec:exp_uni_rlbench} (unimodal) and \secref{sec:exp_multi_rlbench} (multimodal).
Specifically, we replace the original Franka Emika arm and gripper with a UR5 arm and Robotiq 85mm gripper, as shown in  \figref{fig:ur5}.

\para{Baselines: }
We compare \vapor{} against naive transfer baselines by rolling out the learned policies from \secref{sec:exp_uni_rlbench} and \secref{sec:exp_multi_rlbench} on the new embodiment.\footnote{We use the Diffusion Policies trained on 100 demonstrations.}
In the unimodal case, we further contrast optimized joint-space trajectories with optimized end-effector-space trajectories.
In the multimodal case, recall from \ref{para:traj_opt_update} that we leverage the optimized trajectory's likelihood for mode selection.
We analyze whether performance gains stem solely from better mode selection or from trajectory optimization itself. 
Finally, consider that all tasks were designed with the Franka Emika arm in mind and might not even be feasible with the UR5.
We therefore train a set of policies using demonstrations collected with the UR5 to estimate the expected performance of the policy.

\para{Results:}
\tabref{tab:success_rates_traj_opt_uni} and \tabref{tab:success_rates_traj_opt_mult} reveal that the naive Diffusion baseline fails to transfer to the UR5, once again highlighting its vulnerability to out-of-distribution inputs. 
The UR5 has different kinematics than the Franka Emika arm and, therefore, a different initial end-effector pose, which pushes Diffusion out of distribution.
ARP suffers from the same issue.
In contrast, \ourmethod{} achieves respectable performance even under naive transfer.
Furthermore, on unimodal tasks, optimizing only the end-effector trajectory does not significantly improve performance, as it does not prevent the arm from encountering singularities during trajectory execution.
Similarly, on multimodal tasks, selecting the mode without optimizing the joint trajectory only improves policy success if one mode is clearly kinematically superior, such as on \texttt{WipeDesk}. 
In contrast, \vapor{} by optimizing full joint trajectories, successfully incorporates embodiment-specific kinematic evidence into the probabilistic policy.
As a result, compared to naive transfer, \vapor{} increases policy success on average by 36 percentage points on mildly constrained tasks, by 36 percentage points on highly constrained tasks, and by 40 percentage points on multimodal tasks.

Surprisingly, policies transferred from VAPOR often outperform those trained directly on UR5 demonstrations.
This occurs for two key reasons.
First, for many tasks, RLBench's RRTC-based demonstration collection fails on the UR5, which lacks the seventh degree of freedom that the Franka arm has.
Second, successful UR5 demonstrations exhibit low variance, which can impair generalization.
In contrast, Franka-trained policies encode greater variance, allowing VAPOR to explore alternative, feasible trajectories that still satisfy task constraints.
Nevertheless, applying \vapor{} to policies trained on UR5 demonstrations also significantly boosts their success rates.

Overall, VAPOR proves effective for embodiment transfer, achieving an average success rate of 63\%.
Remaining failure cases primarily stem from the UR5’s kinematic limitations.
For example, tasks like \texttt{OpenDrawer} and \texttt{OpenMicrowave} are frequently infeasible with the UR5, as reflected in the demonstration failures.
Additionally, the Robotiq gripper is suboptimal for tasks such as \texttt{TurnTap} and \texttt{SweepToDustpan}.
For feasible tasks, residual failures are often due to self-collisions, which are currently not accounted for in our optimization objective (\eqref{eq:trajCost}).
Future work could address this via explicit self-collision constraints or cost terms.

Finally, \vapor{} is computationally efficient.
When optimization succeeds, \emph{full} trajectory optimization completes in approximately \SI{50}{\milli\second}.
Optimization can be further sped up, for example, by optimizing only partial trajectories.

\subsection{Real World Experiments}\label{sec:exp_real}

\begin{table*}
    \centering
    \begin{threeparttable}
        \caption{Policy success rates on a real Franka Emika robot for unimodal tasks. }\label{tab:success_rates_real_uni}
        \centering
        \setlength{\tabcolsep}{6.5pt}
        \begin{tabular}{l ccccc ccc ccccc}
            \toprule
            \multirow{2}{*}{\textbf{Method}} & \multicolumn{5}{c}{\textbf{Mildly Constrained Tasks}} & \multicolumn{3}{c}{\textbf{Visual Generalization}} & \multicolumn{5}{c}{\textbf{Highly Constrained Task}} \\
            \cmidrule(lr){2-6}
            \cmidrule(lr){7-9}
            \cmidrule(lr){10-14}
            & \makecell{PickAnd \\ Place} & \makecell{Pour \\ Drink} & \makecell{Make \\ Coffee} & \makecell{Bake \\ Bread} & Avg. & \makecell{PickAnd \\ Place} & \makecell{Pour \\ Drink} & Avg. & \makecell{Open \\ Cabinet} & \makecell{Wipe \\ Desk} & \makecell{ScoopWith \\ Spatula} & \makecell{Pour \\ Spiral} & Avg. \\
            \midrule
            LSTM~\cite{hochreiter1997long} & 0.00 & 0.00 & 0.00 & 0.00 & 0.00 & 0.00 & 0.00 & 0.00 & 0.00 & 0.00 & 0.00 & 0.00 & 0.00 \\
            Diffusion Policy~\cite{chi2023diffusionpolicy} & 0.00 & 0.00 & 0.00 & 0.00 & 0.00 & 0.00 & 0.00 & 0.00 & 0.00 & 0.00 & 0.00 & 0.00 & 0.00  \\
            ARP~\cite{zhang2024arp} & 0.00 & 0.00 & 0.00 & 0.00 & 0.00 & 0.00 & 0.00 & 0.00 & 0.00 & 0.00 & 0.00 & 0.00 & 0.00 \\
            TAPAS-GMM~\cite{vonhartz2024art} & 0.96 & \textbf{1.00} & 0.92 & 0.92 & 0.95 & 0.92 & 0.96 & 0.94 & 0.00 & 0.12 & 0.04 & 0.00 & 0.04\\
            \textbf{\ourmethod{} (Ours)} & \textbf{1.00} & \textbf{1.00} & \textbf{1.00} & \textbf{1.00} & \textbf{1.00} & \textbf{1.00} & \textbf{1.00} & \textbf{1.00} & \textbf{0.96} & \textbf{1.00} & \textbf{0.88} & \textbf{0.92} & \textbf{0.95} \\
            \bottomrule
        \end{tabular}
    \end{threeparttable}
    \vspace{-.3cm}
\end{table*}
\begin{table}
    \centering
        \caption{Real-world policy success rates for multimodal tasks. }\label{tab:success_rates_real_multi}
        \setlength{\tabcolsep}{10pt}
        \begin{tabular}{l cccc} %
            \toprule
            \multirow{2}{*}{\textbf{Method}} & \multicolumn{4}{c}{\textbf{Multimodal Tasks}} \\ %
            \cmidrule(lr){2-5}
            & \makecell{PickAnd \\ Place} & \makecell{Open \\ Drawer} & \makecell{Wipe \\ Desk} & \multirow{2}{*}{Avg.}\\ %
            \midrule
            LSTM~\cite{hochreiter1997long} & 0.00 & 0.00 & 0.00 & 0.00 \\
            Diffusion Policy~\cite{chi2023diffusionpolicy} & 0.00 & 0.00 & 0.00 & 0.00 \\
            ARP~\cite{zhang2024arp} & 0.00 & 0.00 & 0.00 & 0.00 \\
            \textbf{\ourmethod{} (Ours)} &  \textbf{1.00} & \textbf{0.96}  & \textbf{1.00} & \textbf{0.99} \\
            \bottomrule
        \end{tabular}
    \end{table}
\begin{table}
    \centering
        \caption{Real-world policy success rates with inference-time updating. }\label{tab:success_rates_real_update}
        \setlength{\tabcolsep}{5pt}
        \begin{tabular}{l cccc c}
            \toprule
            \multirow{2}{*}{\textbf{Method}} & \multicolumn{5}{c}{\textbf{Collision Avoidance}} \\
            \cmidrule(lr){2-6}
            & \multicolumn{2}{c}{PickAndPlace} &  \multicolumn{2}{c}{OpenDrawer} & \multirow{2}{*}{Avg.}\\
            \cmidrule(lr){2-3} \cmidrule(lr){4-5}
            & Level 1 & Level 2 & Level 1 & Level 2 \\
            \midrule
            Diffusion Policy~\cite{chi2023diffusionpolicy} & 0.00 & 0.00 & 0.00 & 0.00 & 0.00  \\
            └─ + ITPS~\cite{wang2024inference} & 0.00 & 0.00 & 0.00 & 0.00 & 0.00  \\
            \textbf{\ourmethod{} (Ours)} & 0.50 & 0.50 & 0.48 & 0.50 & 0.49 \\
            └─ + \textbf{Updating (Ours)} & \textbf{1.00} & \textbf{1.00} & \textbf{0.96} & \textbf{1.00} & \textbf{0.99} \\ 
            \bottomrule
        \end{tabular}
    \vspace{-.3cm}
\end{table}
\begin{table}[t]
    \centering
    \caption{\rebuttal{DINO encoder inference costs.}}
    \rebuttal{
    \begin{threeparttable}
    \begin{tabular}{c cccc}
        \toprule
        \multirow{2}{*}{\textbf{Image Size} [px]} & \multicolumn{2}{c}{\textbf{Wall Clock Time} [\si{\milli\second}]} & \multicolumn{2}{c}{\textbf{Memory Usage} [\si{\mega\byte}]}\\
        \cmidrule(lr){2-3}\cmidrule(lr){4-5}
        & \makebox[1.2cm]{CPU} & \makebox[1.2cm]{GPU} & \makebox[1cm]{Model}  & \makebox[1cm]{Total Peak} \\
        \midrule
        $256\times256$ & \phantom{0}140 & 3 & 100 & \phantom{0}150 \\
        $480\times640$ & 4000 & 3 & 100 & 1200 \\
    \bottomrule
    \end{tabular}
      \begin{tablenotes}[para,flushleft]
       \footnotesize      
       Measurements were made on a Ryzen 9 5950x CPU and RTX3090 GPU.
     \end{tablenotes}
   \end{threeparttable}}
   \label{tab:dino_speed}
\vspace{-0.3cm}
\end{table}

\para{Tasks:}
We evaluate \ourmethod{} on a real Franka Emika robot equipped with a Realsense D435 wrist-mounted camera, and utilize a comprehensive suite of diverse tasks illustrated in \figref{fig:real_tasks}.
Our evaluation comprises three main task categories:
First, we adopt four mildly constrained tasks from TAPAS-GMM~\cite{vonhartz2024art}.
These include two long-horizon tasks (\texttt{MakeCoffee} and \texttt{BakeBread}), as well as two tasks focused on visual generalization (\texttt{PickAndPlace} and \texttt{PourDrink}).
Second, we introduce four highly constrained tasks:  \texttt{OpenCabinet}, \texttt{WipeDesk}, \texttt{ScoopWithSpatula}, and \texttt{PourSpiral}, each demanding precise motion prediction due to narrow operational tolerances.
Third, we establish three multimodal tasks: \texttt{PickAndPlace}, \texttt{OpenDrawer}, and \texttt{WipeDesk}.
Each admits multiple valid execution modes, e.g., placing objects at different locations, choosing among drawers, or wiping along different patterns.
Qualitative examples of task executions are provided in the supplementary video.
We train all policies on five task demonstrations for the unimodal tasks and 10 demonstrations for the multimodal tasks, and evaluate them for 25 episodes each.
For ARP, we plan the trajectories using MPLib~\cite{mplib}.

\para{Task Parameters:}
Following TAPAS-GMM~\cite{vonhartz2024art}, we utilize DINO features to automatically extract relevant task parameters from RGB-D observations.
This approach enables robust visual generalization, allowing policies to ignore distractor objects and generalize across object instances.
We show that \ourmethod{} can leverage these features just as effectively. 
Specifically, we use DINO-derived task parameters on all mildly constrained tasks except \texttt{PickAndPlace}.
For the remaining tasks, we employ FoundationPose~\cite{wen2024foundationpose} to estimate the task parameters, as the DINO-based approach is not well suited for estimating object \emph{orientations}.
Importantly, all baseline models are also provided with the same estimated task parameters, whether extracted via DINO or FoundationPose, to ensure a fair comparison.
\rebuttal{\tabref{tab:dino_speed} lists the the DINO encoder's inference costs.}

\para{Visual Generalization:}
Additionally, we conduct a generalization study using \texttt{PickAndPlace} (FoundationPose) and \texttt{PourDrink} (DINO).
In both tasks, we alter the visual environment using different tablecloths and tablets and introduce distractor objects.
For the DINO-based experiment, we further test instance-level generalization by replacing the can and cup with various cans, bottles, cups, and mugs.
FoundationPose does not generalize across object instances in the same way.
Representative example scenes are shown in \figref{fig:viz_gen}, with the full range of generalization capabilities best appreciated in the supplementary video.

\para{Obstacle Avoidance:}
Finally, to evaluate the effectiveness of our constrained Gaussian updating, we augment the multimodal task scenes with additional clutter objects, strategically placed to obstruct specific execution modes.
Example setups are shown in \figref{fig:viz_gen}.
This setup enables us to assess how well \ourmethod{} can adapt its behavior at inference time to avoid obstacles and select feasible modes.

\para{Results:}
The policy success rates on the unimodal tasks, as reported in \tabref{tab:success_rates_real_uni}, corroborate our findings from the simulation experiments.
First, all deep learning baselines fail in the few-shot regime.
This failure extends to both dense trajectory prediction models, such as Diffusion Policy, and key-pose-based models like ARP.
Second, TAPAS-GMM performs strongly on mildly constrained tasks but struggles on highly constrained tasks due to its limited expressivity.
It also fails on the dynamic task \texttt{ScoopWithSpatula}.
In contrast, \ourmethod{} excels across the board, achieving strong policy success even on the hardest tasks.
Compared to TAPAS-GMM, \ourmethod{} improves policy success on average by five percentage points on the mildly constrained tasks and by 91 percentage points on the highly constrained tasks.

Third, \ourmethod{} also effectively leverages the DINO keypoints and FoundationPose prediction for robust visual generalization, including instance-level generalization, added distractor objects, and changing task environments.
Success rates (\tabref{tab:success_rates_real_uni}) remain competitive with the original task setup, even slightly outperforming TAPAS-GMM due to more robust trajectory modeling.
The generalization capabilities can be appreciated qualitatively in the supplementary video.

Fourth, as in \secref{sec:exp_uni_rlbench}, \ourmethod{} generates trajectories that are  much smoother than those predicted by TAPAS-GMM.
On the real robot, smoothness of the trajectories is much more critical than in simulation, as the simulation simplifies physical aspects such as inertia and imperfect controllers.
Consequently, for safe execution, TAPAS-GMM requires post-processing with TOPP-RA~\cite{pham2018new} to smooth out the trajectories.
And even after postprocessing, \ourmethod{}'s trajectories remain substantially smoother than those of TAPAS-GMM, as can be appreciated in the supplementary video.

Likewise, the results for the multimodal tasks, reported in \tabref{tab:success_rates_real_multi}, confirm our results from \secref{sec:exp_multi_rlbench}.
None of the baseline methods can reliably solve any of the tasks given only 10 demonstrations.
In contrast, \ourmethod{} masters all tasks.
Both DBSCAN and \(k\)-Means partition the modes of each trajectory distribution in less than a second.
Although again, \(k\)-Means is a bit slower, it works more robustly and requires less hyperparameter tuning.
\ourmethod{} then effectively fits the multimodal distributions and generates safe and smooth trajectories for all modes.

Finally, \tabref{tab:success_rates_real_update} shows the results for the collision avoidance experiments using inference-time policy updating.
As in the simulated experiments, Diffusion Policy and ITPS are ineffective in avoiding obstacles at inference time as they tend to go out-.
Diffusion Policy is already brittle in the few-shot regime, and ITPS tends to push it out of distribution.
We provide a deeper analysis in \secref{sec:exp_constraints}.
In contrast, \ourmethod{} solves the task reliably.
When equipped with constrained Gaussian updating, it successfully avoids the obstacles while still fulfilling the task objectives.
This can also be appreciated in the supplementary video.

\section{Limitations}
While \ourmethod{} can in principle learn robot manipulation policies with sub-millimeter accuracy, in practice it is limited by the quality of the available perception systems.
Yet, this modularity is also a great advantage of our method, as it allows us to further improve policy learning in lockstep with each advance in computer vision.
For example, by harnessing both DINO and FoundationPose in our experiments, we have demonstrated how their respective advantages (instance-level generalization versus 6-DoF pose estimation) can be leveraged for effective policy learning. 
Furthermore, the modularity of our approach improves its interpretability, thereby fostering trust and safety.
Next, scaling up deep policy learning may lead to emergent capabilities, such as \rebuttal{retrying}.
\ourmethod{} is unlikely to exhibit such emergent capabilities.
However, \rebuttal{retrying} behavior might be added in other ways, for example, by monitoring policy execution and triggering re-execution of the skill policy.

Furthermore, in this work, we have focused on task-parameterized learning of end-effector motions.
We did not study movements that require learning in joint space, such as walking.
While \ourmethod{}'s expressivity would allow for imitation of joint trajectories, joint space has a fundamentally different geometry, preventing the kind of task-space generalization we have exploited here.
Instead, one might need to learn high-level navigation commands or focus on wheeled locomotion.
Finally, we have relied on TAPAS~\cite{vonhartz2024art} for skill segmentation and alignment.
Consequently, our results might not translate to tasks where individual skills cannot be readily segmented.
\section{Conclusion}
In this work, we presented Mixture of Discrete-time Gaussian Processes (\ourmethod{}), a novel policy representation and policy learning approach.
Through extensive experiments, we showed that \ourmethod{} excels at long-horizon tasks, highly constrained movements, dynamic movements, and multimodal tasks, while requiring as few as five demonstrations.
\ourmethod{} exhibits excellent extrapolation capabilities, while generating smooth trajectories.
It is not only more expressive than the previous few-shot methods, but also more computationally efficient.
We further showed that our probabilistic approach to policy learning enables effective inference-time updating, unlocking novel axes of generalization, such as object avoidance, while still ensuring task success.
Additionally, we proposed variance-aware path optimization (\vapor) and showed that together with \ourmethod{} it enables potent cross-embodiment transfer.

\bibliographystyle{IEEEtran}
\bibliography{IEEEabrv,root}

\begin{thebibliography}{10}
\providecommand{\url}[1]{#1}
\csname url@rmstyle\endcsname
\providecommand{\newblock}{\relax}
\providecommand{\bibinfo}[2]{#2}
\providecommand\BIBentrySTDinterwordspacing{\spaceskip=0pt\relax}
\providecommand\BIBentryALTinterwordstretchfactor{4}
\providecommand\BIBentryALTinterwordspacing{\spaceskip=\fontdimen2\font plus
\BIBentryALTinterwordstretchfactor\fontdimen3\font minus \fontdimen4\font\relax}
\providecommand\BIBforeignlanguage[2]{{%
\expandafter\ifx\csname l@#1\endcsname\relax
\typeout{** WARNING: IEEEtran.bst: No hyphenation pattern has been}%
\typeout{** loaded for the language `#1'. Using the pattern for}%
\typeout{** the default language instead.}%
\else
\language=\csname l@#1\endcsname
\fi
#2}}

\bibitem{janner2022planning}
M.~Janner, Y.~Du, J.~B. Tenenbaum, and S.~Levine, ``Planning with diffusion for flexible behavior synthesis,'' \emph{arXiv preprint arXiv:2205.09991}, 2022.

\bibitem{chi2023diffusionpolicy}
C.~Chi, S.~Feng, Y.~Du, Z.~Xu, E.~Cousineau, B.~Burchfiel, and S.~Song, ``Diffusion policy: Visuomotor policy learning via action diffusion,'' in \emph{Proc. of Robotics: Science and Systems (RSS)}, 2023.

\bibitem{chisari2024learningroboticmanipulationpolicies}
E.~Chisari, N.~Heppert, M.~Argus, T.~Welschehold, T.~Brox, and A.~Valada, ``Learning robotic manipulation policies from point clouds with conditional flow matching,'' \emph{arXiv preprint arXiv:2409.07343}, 2024.

\bibitem{lipman2022flow}
Y.~Lipman, R.~T. Chen, H.~Ben-Hamu, M.~Nickel, and M.~Le, ``Flow matching for generative modeling,'' \emph{arXiv preprint arXiv:2210.02747}, 2022.

\bibitem{braun2024riemannian}
M.~Braun, N.~Jaquier, L.~Rozo, and T.~Asfour, ``Riemannian flow matching policy for robot motion learning,'' \emph{arXiv preprint arXiv:2403.10672}, 2024.

\bibitem{vonhartz2023treachery}
J.~O. von Hartz, E.~Chisari, T.~Welschehold, W.~Burgard, J.~Boedecker, and A.~Valada, ``The treachery of images: Bayesian scene keypoints for deep policy learning in robotic manipulation,'' \emph{IEEE Robotics and Automation Letters}, vol.~8, no.~11, pp. 6931--6938, 2023.

\bibitem{zhang2024arp}
X.~Zhang, Y.~Liu, H.~Chang, L.~Schramm, and A.~Boularias, ``Autoregressive action sequence learning for robotic manipulation,'' \emph{arXiv preprint arXiv:2410.03132}, 2024.

\bibitem{calinon2006learning}
S.~Calinon, F.~Guenter, and A.~Billard, ``On learning the statistical representation of a task and generalizing it to various contexts,'' in \emph{Int.~Conf.~on Robotics and Automation}, 2006, pp. 2978--2983.

\bibitem{zeestraten2018programming}
M.~J. Zeestraten, ``Programming by demonstration on riemannian manifolds.'' Ph.D. dissertation, University of Genoa, Italy, 2018.

\bibitem{vonhartz2024art}
J.~O. von Hartz, T.~Welschehold, A.~Valada, and J.~Boedecker, ``The art of imitation: Learning long-horizon manipulation tasks from few demonstrations,'' \emph{arXiv preprint arXiv:2407.13432}, 2024.

\bibitem{deisenroth2013gaussian}
M.~P. Deisenroth, D.~Fox, and C.~E. Rasmussen, ``Gaussian processes for data-efficient learning in robotics and control,'' \emph{IEEE transactions on pattern analysis and machine intelligence}, vol.~37, no.~2, pp. 408--423, 2013.

\bibitem{titsias2009variational}
M.~Titsias, ``Variational learning of inducing variables in sparse gaussian processes,'' in \emph{Artificial intelligence and statistics}, 2009, pp. 567--574.

\bibitem{franzese2024generalization}
G.~Franzese, R.~Prakash, and J.~Kober, ``Generalization of task parameterized dynamical systems using gaussian process transportation,'' \emph{arXiv preprint arXiv:2404.13458}, 2024.

\bibitem{sun2023damm}
S.~Sun, H.~Gao, T.~Li, and N.~Figueroa, ``Damm: Directionality-aware mixture model parallel sampling for efficient dynamical system learning,'' \emph{arXiv preprint arXiv:2309.02609}, 2023.

\bibitem{bouveyron2007high}
C.~Bouveyron, S.~Girard, and C.~Schmid, ``High-dimensional data clustering,'' \emph{Computational statistics \& data analysis}, vol.~52, no.~1, pp. 502--519, 2007.

\bibitem{chen2024mirage}
L.~Y. Chen, K.~Hari, K.~Dharmarajan, C.~Xu, Q.~Vuong, and K.~Goldberg, ``Mirage: Cross-embodiment zero-shot policy transfer with cross-painting,'' \emph{arXiv preprint arXiv:2402.19249}, 2024.

\bibitem{calinon2007learning}
S.~Calinon, F.~Guenter, and A.~Billard, ``On learning, representing, and generalizing a task in a humanoid robot,'' \emph{IEEE Trans. on Systems, Man, and Cybernetics}, vol.~37, no.~2, pp. 286--298, 2007.

\bibitem{goyal2023rvt}
A.~Goyal, J.~Xu, Y.~Guo, V.~Blukis, Y.-W. Chao, and D.~Fox, ``Rvt: Robotic view transformer for 3d object manipulation,'' in \emph{Conf. on Robot Learning}, 2023, pp. 694--710.

\bibitem{hochreiter1997long}
S.~Hochreiter and J.~Schmidhuber, ``Long short-term memory,'' \emph{Neural computation}, vol.~9, no.~8, pp. 1735--1780, 1997.

\bibitem{chisari2022correct}
E.~Chisari, T.~Welschehold, J.~Boedecker, W.~Burgard, and A.~Valada, ``Correct me if i am wrong: Interactive learning for robotic manipulation,'' \emph{IEEE Robotics and Automation Letters}, 2022.

\bibitem{florence2019self}
P.~Florence, L.~Manuelli, and R.~Tedrake, ``Self-supervised correspondence in visuomotor policy learning,'' \emph{IEEE Robotics and Automation Letters}, vol.~5, no.~2, pp. 492--499, 2019.

\bibitem{rana2024affordancecentricpolicylearningsample}
K.~Rana, J.~Abou-Chakra, S.~Garg, R.~Lee, I.~Reid, and N.~Suenderhauf, ``Affordance-centric policy learning: Sample efficient and generalisable robot policy learning using affordance-centric task frames,'' \emph{arXiv preprint arXiv:2410.12124}, 2024.

\bibitem{joukov2017gaussian}
V.~Joukov and D.~Kulic, ``Gaussian process based model predictive controller for imitation learning,'' in \emph{IEEE-RAS 17th International Conference on Humanoid Robotics (Humanoids)}, 2017, pp. 850--855.

\bibitem{lang2017computationally}
M.~Lang and S.~Hirche, ``Computationally efficient rigid-body gaussian process for motion dynamics,'' \emph{IEEE Robotics and Automation Letters}, vol.~2, no.~3, pp. 1601--1608, 2017.

\bibitem{jaquier2020learning}
N.~Jaquier, D.~Ginsbourger, and S.~Calinon, ``Learning from demonstration with model-based gaussian process,'' in \emph{Conference on Robot Learning}.\hskip 1em plus 0.5em minus 0.4em\relax PMLR, 2020, pp. 247--257.

\bibitem{paraschos2013probabilistic}
A.~Paraschos, C.~Daniel, J.~R. Peters, and G.~Neumann, ``Probabilistic movement primitives,'' \emph{Advances in neural information processing systems}, vol.~26, 2013.

\bibitem{paraschos2018using}
A.~Paraschos, C.~Daniel, J.~Peters, and G.~Neumann, ``Using probabilistic movement primitives in robotics,'' \emph{Autonomous Robots}, vol.~42, pp. 529--551, 2018.

\bibitem{rozo2022orientation}
L.~Rozo and V.~Dave, ``Orientation probabilistic movement primitives on riemannian manifolds,'' in \emph{Conference on Robot Learning}, 2022, pp. 373--383.

\bibitem{rueckert2015lowdim}
E.~Rueckert, J.~Mundo, A.~Paraschos, J.~Peters, and G.~Neumann, ``Extracting low-dimensional control variables for movement primitives,'' in \emph{2015 IEEE International Conference on Robotics and Automation (ICRA)}, 2015, pp. 1511--1518.

\bibitem{brandi2014generalizing}
S.~Brandi, O.~Kroemer, and J.~Peters, ``Generalizing pouring actions between objects using warped parameters,'' in \emph{IEEE-RAS Int. Conf. on Humanoid Robots}, 2014, pp. 616--621.

\bibitem{huang2019kernelized}
Y.~Huang, L.~Rozo, J.~Silv{\'e}rio, and D.~G. Caldwell, ``Kernelized movement primitives,'' \emph{The Int. Journal of Robotics Research}, vol.~38, no.~7, pp. 833--852, 2019.

\bibitem{kramberger2016generalization}
A.~Kramberger, A.~Gams, B.~Nemec, and A.~Ude, ``Generalization of orientational motion in unit quaternion space,'' in \emph{IEEE-RAS Int. Conf. on Humanoid Robots (Humanoids)}, 2016, pp. 808--813.

\bibitem{calinon2016tutorial}
S.~Calinon, ``A tutorial on task-parameterized movement learning and retrieval,'' \emph{Intelligent service robotics}, vol.~9, pp. 1--29, 2016.

\bibitem{li2023task}
T.~Li and N.~Figueroa, ``Task generalization with stability guarantees via elastic dynamical system motion policies,'' in \emph{Conf. on Robot Learning}, 2023, pp. 3485--3517.

\bibitem{heppert2024ditto}
N.~Heppert, M.~Argus, T.~Welschehold, T.~Brox, and A.~Valada, ``Ditto: Demonstration imitation by trajectory transformation,'' \emph{arXiv preprint arXiv:2403.15203}, 2024.

\bibitem{Ze2024DP3}
Y.~Ze, G.~Zhang, K.~Zhang, C.~Hu, M.~Wang, and H.~Xu, ``3d diffusion policy: Generalizable visuomotor policy learning via simple 3d representations,'' in \emph{Proceedings of Robotics: Science and Systems (RSS)}, 2024.

\bibitem{3d_diffuser_actor}
T.-W. Ke, N.~Gkanatsios, and K.~Fragkiadaki, ``3d diffuser actor: Policy diffusion with 3d scene representations,'' \emph{Arxiv}, 2024.

\bibitem{james2022coarse}
S.~James, K.~Wada, T.~Laidlow, and A.~J. Davison, ``Coarse-to-fine q-attention: Efficient learning for visual robotic manipulation via discretisation,'' in \emph{Proceedings of the IEEE/CVF Conference on Computer Vision and Pattern Recognition}, 2022, pp. 13\,739--13\,748.

\bibitem{chisari2023centergrasp}
E.~Chisari, N.~Heppert, T.~Welschehold, W.~Burgard, and A.~Valada, ``Centergrasp: Object-aware implicit representation learning for simultaneous shape reconstruction and 6-dof grasp estimation,'' \emph{arXiv preprint arXiv:2312.08240}, 2023.

\bibitem{alizadeh2016identifying}
T.~Alizadeh and M.~Malekzadeh, ``Identifying the relevant frames of reference in programming by demonstration using task-parameterized gaussian mixture regression,'' in \emph{IEEE/SICE Int. Symposium on System Integration}, 2016, pp. 453--458.

\bibitem{rozo2020learning}
L.~Rozo, M.~Guo, A.~G. Kupcsik, M.~Todescato, P.~Schillinger, M.~Giftthaler, M.~Ochs, M.~Spies, N.~Waniek, P.~Kesper, \emph{et~al.}, ``Learning and sequencing of object-centric manipulation skills for industrial tasks,'' in \emph{Int.~Conf.~on Intelligent Robots and Systems}, 2020, pp. 9072--9079.

\bibitem{arduengo2023gaussian}
M.~Arduengo, A.~Colom{\'e}, J.~Lobo-Prat, L.~Sentis, and C.~Torras, ``Gaussian-process-based robot learning from demonstration,'' \emph{Journal of Ambient Intelligence and Humanized Computing}, pp. 1--14, 2023.

\bibitem{monterolearning}
M.~R. Montero, G.~Franzese, J.~Kober, and C.~Della~Santina, ``Learning multi-reference frame skills from demonstration with task-parameterized gaussian processes,'' pp. 2832--2839, 2024.

\bibitem{ho2022classifier}
J.~Ho and T.~Salimans, ``Classifier-free diffusion guidance,'' \emph{arXiv preprint arXiv:2207.12598}, 2022.

\bibitem{ajay2022conditional}
A.~Ajay, Y.~Du, A.~Gupta, J.~Tenenbaum, T.~Jaakkola, and P.~Agrawal, ``Is conditional generative modeling all you need for decision-making?'' \emph{arXiv preprint arXiv:2211.15657}, 2022.

\bibitem{wang2024inference}
Y.~Wang, L.~Wang, Y.~Du, B.~Sundaralingam, X.~Yang, Y.-W. Chao, C.~Perez-D'Arpino, D.~Fox, and J.~Shah, ``Inference-time policy steering through human interactions,'' \emph{arXiv preprint arXiv:2411.16627}, 2024.

\bibitem{wang2024poco}
L.~Wang, J.~Zhao, Y.~Du, E.~H. Adelson, and R.~Tedrake, ``Poco: Policy composition from and for heterogeneous robot learning,'' \emph{arXiv preprint arXiv:2402.02511}, 2024.

\bibitem{xian2023chaineddiffuser}
Z.~Xian, N.~Gkanatsios, T.~Gervet, T.-W. Ke, and K.~Fragkiadaki, ``Chaineddiffuser: Unifying trajectory diffusion and keypose prediction for robotic manipulation,'' in \emph{Conf. on Robot Learning}, 2023.

\bibitem{bishop1994mixture}
C.~M. Bishop, ``Mixture density networks,'' 1994.

\bibitem{cohn1996active}
D.~A. Cohn, Z.~Ghahramani, and M.~I. Jordan, ``Active learning with statistical models,'' \emph{Journal of artificial intelligence research}, vol.~4, pp. 129--145, 1996.

\bibitem{ester1996density}
M.~Ester, H.-P. Kriegel, J.~Sander, X.~Xu, \emph{et~al.}, ``A density-based algorithm for discovering clusters in large spatial databases with noise,'' in \emph{kdd}, vol.~96, 1996, pp. 226--231.

\bibitem{rofer2022kineverse}
A.~Rofer, G.~Bartels, W.~Burgard, A.~Valada, and M.~Beetz, ``Kineverse: A symbolic articulation model framework for model-agnostic mobile manipulation,'' \emph{IEEE Robotics and Automation Letters}, 2022.

\bibitem{toussaint2017tutorial}
M.~Toussaint, ``A tutorial on newton methods for constrained trajectory optimization and relations to slam, gaussian process smoothing, optimal control, and probabilistic inference,'' \emph{Geometric and numerical foundations of movements}, pp. 361--392, 2017.

\bibitem{said2017riemannian}
S.~Said, L.~Bombrun, Y.~Berthoumieu, and J.~H. Manton, ``Riemannian gaussian distributions on the space of symmetric positive definite matrices,'' \emph{IEEE Transactions on Information Theory}, vol.~63, no.~4, pp. 2153--2170, 2017.

\bibitem{kuffner2000rrtc}
J.~Kuffner and S.~LaValle, ``Rrt-connect: An efficient approach to single-query path planning,'' in \emph{Proceedings 2000 ICRA. Millennium Conference. IEEE International Conference on Robotics and Automation. Symposia Proceedings (Cat. No.00CH37065)}, vol.~2, 2000, pp. 995--1001.

\bibitem{steinhaus1957}
H.~Steinhaus, ``Sur la division des corps matériels en parties,'' in \emph{Bulletin L'Académie Polonaise des Science}, vol.~4, no.~12, 1957, pp. 801--804.

\bibitem{chisari2024flowmatch}
E.~Chisari, N.~Heppert, M.~Argus, T.~Welschehold, T.~Brox, and A.~Valada, ``Learning robotic manipulation policies from point clouds with conditional flow matching,'' \emph{arXiv preprint arXiv:2409.07343}, 2024.

\bibitem{mplib}
\BIBentryALTinterwordspacing
R.~K. Guo, X.~Lin, M.~Liu, J.~Gu, and H.~Su, ``{MPlib: a Lightweight Motion Planning Library}.'' [Online]. Available: \url{https://github.com/haosulab/MPlib}
\BIBentrySTDinterwordspacing

\bibitem{wen2024foundationpose}
B.~Wen, W.~Yang, J.~Kautz, and S.~Birchfield, ``Foundationpose: Unified 6d pose estimation and tracking of novel objects,'' in \emph{Proceedings of the IEEE/CVF Conference on Computer Vision and Pattern Recognition}, 2024, pp. 17\,868--17\,879.

\bibitem{pham2018new}
H.~Pham and Q.-C. Pham, ``A new approach to time-optimal path parameterization based on reachability analysis,'' \emph{IEEE Trans. on Robotics}, vol.~34, no.~3, pp. 645--659, 2018.

\end{thebibliography}

\clearpage
\renewcommand{\baselinestretch}{1}
\setlength{\belowcaptionskip}{0pt}

\begin{strip}
\begin{center}
\vspace{-5ex}
\textbf{\LARGE \bf
The Unreasonable Effectiveness of Discrete-Time Gaussian Process\\
\vspace{0.5ex}Mixtures for Robot Policy Learning} \\
\vspace{3ex}

\Large{\bf- Supplementary Material -}\\
\vspace{0.4cm}
\normalsize{Jan Ole von Hartz, Adrian Röfer, Joschka Boedecker, and Abhinav Valada%
}
\end{center}
\end{strip}

\setcounter{section}{0}
\setcounter{equation}{0}
\setcounter{figure}{0}
\setcounter{table}{0}
\setcounter{page}{1}
\makeatletter

\renewcommand{\thesection}{S.\arabic{section}}
\renewcommand{\thesubsection}{S.\arabic{subsection}}
\renewcommand{\thetable}{S.\arabic{table}}
\renewcommand{\thefigure}{S.\arabic{figure}}

\section{Policy Success vs.\ Task Instance Dispersion}\label{sec:taskvar}
\begin{figure*}
    \centering
    \includegraphics[width=\linewidth]{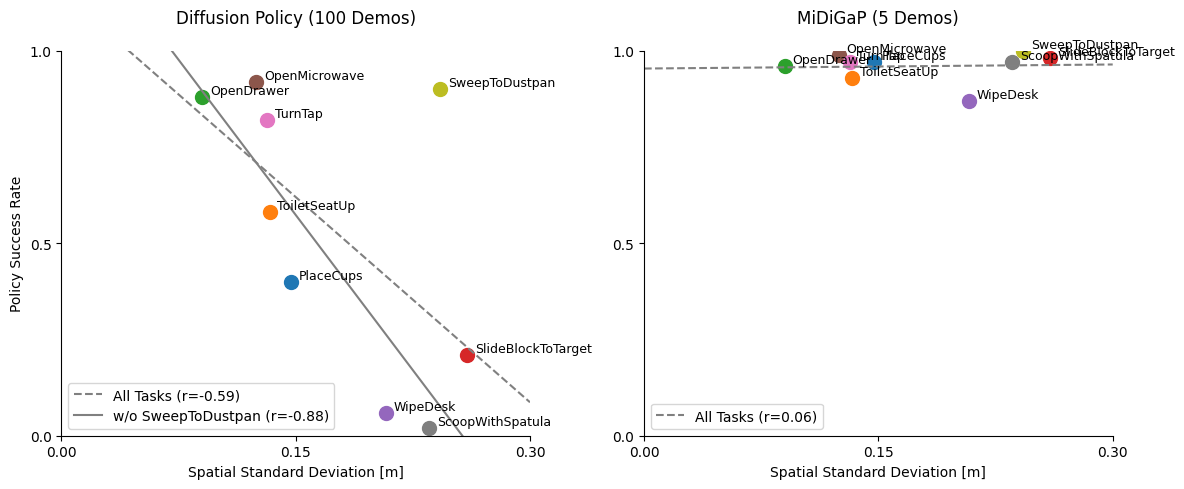}
    \caption{Policy success rates versus positional standard deviation of task objects.
    Diffusion Policy success rate are for 100 demonstrations.
    Policy success rates are taken from Table III (p.\ 12).
    The positional standard deviation of the task objects is computed empirically from the set of demonstrations.
    For multi-object tasks, only the first object is used for simplicity.
    Even using such a simple metric for the task dispersion shows how strongly Diffusion Policy's performance depends on the task dispersion (\(r=-0.59\)).
    SweepToDustpan is a clear outlier because the task distribution is bimodal (sweep left vs.\ sweep right) with little variance within the modes.
    Diffusion Policy has no problems modeling multimodal distributions, but struggles with large variance in the task space.
    Therefore removing SweepToDustpan further strengthens the correlation (\(r=-0.88\)).
    In contrast, MiDiGaP's performance does not strongly depend on the task dispersion.
    }
    \label{fig:diff_vs_var}
\end{figure*}

Our results yield valuable insight into which tasks current policy learning methods can learn effectively.
Diffusion Policy seem to struggle primarily when task instance dispersion (i.e.\ the dispersion of the object poses across task instances) is large.
\figref{fig:diff_vs_var} shows how strong this effect is.
Even a very crude measure of task dispersion (the positional standard deviation of the first task object) correlates negatively and strongly with Diffusion Policy's task performance (\(r=-0.88\)).
This measure omits many dimensions of hardnesss, like the fact that ScoopWithSpatula is a dynamic task.
In contrast, \ourmethod{} is not negatively affected by task dispersion.

\section{Scaling Behavior}
\begin{figure*}
    \centering
    \includegraphics[width=0.4\linewidth]{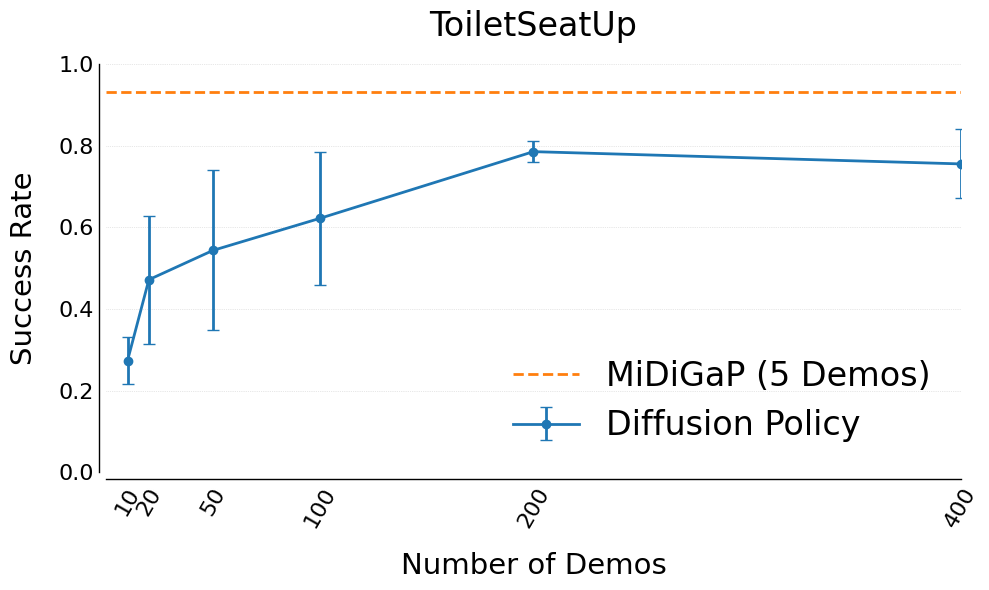}\hfil
    \includegraphics[width=0.4\linewidth]{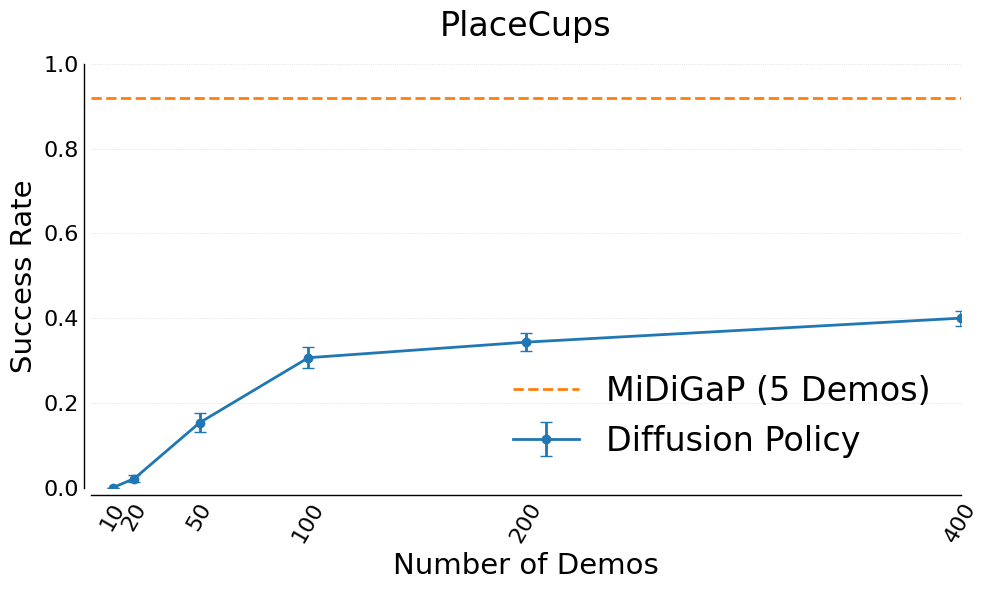}
    \caption{Scaling of Diffusion Policy with number of demonstrations on two RLBench tasks.
    Results are aggregated over three training seeds.}
    \label{fig:demo_scale}
\end{figure*}

We perform additional scaling experiments to see whether additional demonstrations help Diffusion Policy catch up with \ourmethod{}.
\figref{fig:demo_scale} shows Diffusion Policy's scaling behavior on two RLBench tasks: ToiletSeatUp and PlaceCups.
On the easier ToiletSeatUp task, policy success seems to saturate around 80\%, while on the more challenging PlaceCups tasks policy success continues to rise, albeit only very slowly.
Even with 400 demonstrations, Diffusion Policy only achieves 40\% policy success.
In contrast, MiDiGaP achieves 93\% on ToiletSeatUp and 97\% on PlaceCups from 5 demonstrations.
It seems reasonable to us that further scaling and architectural improvements could boost Diffusion Policy's performance.
However, at present, even with 400 demonstrations it seems not to catch up with MiDiGaP.

Note that when training on a subset of 100 demonstrations from the newly collected 400 demonstrations for PlaceCups, Diffusion Policy seems to perform slightly worse than on the original 100 demonstrations.
We continue to report the original, higher performance in the paper as it is the stronger baseline and for consistency with the other results.

\begin{table*}[tb]
    \centering
    \caption{Policy success rates on RoboSuite tasks.} 
    \begin{threeparttable}
    \begin{tabular}{ll cccc}
        \toprule
        \multirow{2}{*}{\textbf{Demos}} & \multirow{2}{*}{\textbf{Method}} & \multicolumn{4}{c}{\textbf{Task}}\\
        \cmidrule(lr){3-6}
        & & Lift & Can & ToolHang & Transport \\
        \midrule
        100 & Diffusion Policy~\cite{chi2023diffusionpolicy} & 1.00/1.00 & 1.00/1.00 & 1.00/0.87 & 1.00/0.84\\
        5 & MiDiGaP (Ours) & 1.00 & 1.00 & 1.00 & 1.00\\
    \bottomrule
    \end{tabular}
      \begin{tablenotes}[para,flushleft]
       \footnotesize      
       Success rates for Diffusion Policy are taken from Chi \etal{}\cite{chi2023diffusionpolicy}. Only the best performing Diffusion Policy variant (state-based, U-Net) is presented.
       For Diffusion Policy, success rates are reported as the maximum performance and the average performance over the last 10 checkpoints.
       MiDiGaP's results are deterministic.
     \end{tablenotes}
   \end{threeparttable}
   \label{tab:robosuite}
\end{table*}

\section{Robotsuit Experiments}
To demonstrate that our results are independent of the choice of simulator, we have conducted additional experiments on four RoboSuite tasks: \texttt{Lift}, \texttt{Can}, \texttt{ToolHang}, and \texttt{Transport}.
We compare against the results reported by Chi \etal{} for the Diffusion Policy trained on 100 proficient-human demonstrations.
MiDiGaP is again trained on only five demonstrations per tasks.
The policy success rates are shown in \tabref{tab:robosuite}.
As in RLBench, MiDiGaP performs on par with Diffusion Policy while requiring significantly fewer demonstrations.
Additional insights are harder to elucidate because Diffusion Policy already reaches the performance ceiling on these tasks.

Note that \texttt{Transport} is a bimanual tasks.
The critical challenge here is the (spatial) coordination of the arms.
Because the handover happens in free space and might happen at different positions in different demonstrations, the arms need a mechanism to agree on a common handover point.
We tried two strategies to address this issue, both of which proved to be effective.
First, the arms can be controlled in a leader-follower paradigm, where the second arm observes the pose of the first arm (as an additional task-parameter).
This strategy is easy to set up, as it does not require dynamic object tracking.
However, it is only useful in scenarios where it is apriori know which arm will be the leading arm.
The second, more general strategy is to coordinate the arms via the task objects themselves.
Dynamically tracking the payload through the task execution enables using its dynamic frame as a task-parameter.
When the payload is moved by one arm, the dynamic payload frame signals the handover location to the second arm.
While these are interesting initial results, we leave a more thorough investigation to future work as it is out-of-scope in this work.

\end{document}